%% file: main.tex
\documentclass{article}

\PassOptionsToPackage{numbers, compress}{natbib}

\usepackage[preprint]{neurips_2026}

\usepackage[utf8]{inputenc} 
\usepackage[T1]{fontenc}    
\usepackage{hyperref}       
\usepackage{url}            
\usepackage{booktabs}       
\usepackage{multirow}       
\usepackage{amsfonts}       
\usepackage{nicefrac}       
\usepackage{microtype}      
\usepackage{xcolor}         
\usepackage{amsmath}        
\usepackage{amssymb}        
\usepackage{mathtools}      
\usepackage{graphicx}       
\usepackage{subcaption}     
\usepackage{array}
\usepackage{pdflscape}
\newcommand{\eg}{e\/.\/g\/.,\/~}

\DeclareMathOperator*{\argmin}{arg\,min}

\title{CENDRe: Concept Extraction with Natural Domain Representations}

\author{%
  Antonia Holzapfel\thanks{Corresponding author: \texttt{holzapfel@dsme.rwth-aachen.de}} \quad
  Andres Felipe Posada Moreno \quad
  Sebastian Trimpe \\
  Institute for Data Science in Mechanical Engineering (DSME)\\
  RWTH Aachen University\\
  Theaterstraße 35-39, 52062 Aachen, Germany\\
}

\begin{document}

\maketitle

\begin{abstract}
Convolutional neural networks (CNNs) are widely used for time-series classification, but their deployment in critical contexts requires understanding the patterns  across signal domains  
that drive their predictions (\eg temporal or spectral).
Concept extraction (CE) methods identify such patterns by analyzing representations within the models' latent space.
However, existing time-series CE methods have three limitations: 
they operate only in the time domain and cannot locate features in other domains, they depend on a predefined number of concepts that biases them, and they produce localizations misaligned with the regions the model uses.
We address these limitations with CENDRe, a concept extraction method for CNNs.
For concept discovery, we propose a procedure that clusters per-timestep latent representations in two stages, with a silhouette-guided aggregation that selects the number of concepts automatically.
For localization, we define a presence score that contrasts the latent representations with their prototypes, and use its gradients to produce
masks that concentrate on the regions driving the concept.
These gradients, propagated through a differentiable invertible mapping of the input such as a Fourier transform, yield localizations for the same concepts in different domains.
Finally, each concept receives a relevance score that quantifies its contribution to each class.
On synthetic benchmarks, CENDRe achieves representation correctness comparable to state-of-the-art CE methods and higher importance correctness, while producing visualizations in different domains.
On real bearing-fault data, CENDRe extracts the frequency bands driving the model's predictions, located in regions commonly inspected for fault diagnosis, producing evidence to assess the model that time-domain CE methods cannot.
\end{abstract}

\input{text/10_introduction.tex}
\input{text/20_related_work}

\input{text/40_methods}

\input{text/45_experiment_design}

\input{text/50_results}
\input{text/60_conclusion}

\begin{ack}
We thank Franziska Hoffmann for her valuable help in designing the synthetic datasets.

In addition, model training and experiments were partially performed with computing resources granted by RWTH Aachen
University, under projects p0021919 and p0027430.

This work is funded by the Deutsche Forschungsgemeinschaft (DFG, German Research Foundation) – 520460745.

Funded by the European Union. This work has received funding from the  European High Performance Computing Joint Undertaking (JU) and from the  German Federal Ministry of Research, Technology and Space (BMFTR), the  Ministry of Culture and Science of North Rhine-Westphalia (MKW NRW) and  the Hessian Ministry of Science and Research, Arts and Culture (HMWK)  under grant agreement No 101250682.
\end{ack}

\bibliographystyle{plainnat}
\bibliography{literature}


\appendix

\input{text/97_appendix_spectral_derivation}
\input{text/90_appendix_datasets}
\input{text/92_appendix_models}
\input{text/95_appendix_additional_results}
\input{text/94_appendix_compute}


\newpage

\end{document}

%% file: text/10_introduction.tex
\section{Introduction}

Deep learning, particularly convolutional neural networks (CNNs),
has become a widespread choice for time-series classification and regression.
Applications range from condition monitoring and predictive maintenance on industrial equipment~\cite{kim2023time,stathatos2024convolutional}
to early-warning and diagnostic systems in healthcare~\cite{kraft2025atrial,ravi2024hybrid,fratti2024multi},
settings where a wrong prediction can translate into costly misbehavior.
In such domains, stakeholders cannot simply take a prediction at face value:
they need explanations to verify that it is based on plausible signal features and not on spurious correlations.
For such inspection, explanations must be both faithful to the model's behavior and expressed in an interpretable domain.
Time-domain explanations cover at most half the story
and other domains are often better suited to interpreting signals.
Since models trained on time signals, especially CNNs, can learn features in the time, frequency, or other domains, explanation methods should also operate on these domains to make the corresponding features intelligible.

Among XAI methods, concept methods~\cite{poeta2023concept} are post-hoc and grounded in the model's own latent space, capturing its global behavior.
Specifically, Concept Extraction (CE) discovers recurring latent patterns that the model learned to distinguish across a dataset~\cite{kim2018interpretability}.
While original work in CE has focused on the image domain~\cite{ghorbani2019automatic,kamakshi2021pace,posada-moreno2023scalepreserving,posada-moreno2023eclad},
recent works have transferred CE to time series~\cite{holzapfel2025concept,younis2022multivariate}.
However, these existing methods have three main limitations:
First, their explanations live exclusively in the time domain, leaving unexplained the features that the model uses in any other relevant domain (\eg frequency).
Second, the number of concepts $K$ must be fixed before extraction, forcing a manual sweep that biases the resulting explanations.
Yet $K$ reflects how the model decides and should be determined by the data, not the user.
Third, their localization methods produce masks that may either miss regions the model uses~\cite{posada-moreno2023eclad,holzapfel2025concept} or cover regions it does not~\cite{younis2022multivariate,yeh2020completenessaware}. 
These limitations restrict the practical usefulness of existing CE time-series methods, as explanations in other domains are not covered, and strong priors on concept number and localization are included.

We address these three limitations with \textbf{CENDRe} (\emph{Concept Extraction with Natural Domain Representations}),
a CE method for time-series CNNs that extracts concepts in any natural domain, that is, any invertible and differentiable transform of the input in which signal content is directly interpretable, such as time or frequency.
To this end, we propose a concept discovery algorithm that clusters timestep-wise activations in two stages, with silhouette-guided aggregation selecting the number of concepts automatically, rather than a user. 
Building on the discovered concepts, we introduce a localization mechanism based on the gradients of a contrastive presence score that compares the latent representations and concept prototypes, with respect to the input.
These gradients produce masks that concentrate attribution on the regions influencing the concept.
The same gradients, propagated through virtual inspection layers~\cite{vielhaben2024explainable}, yield concept localizations in other natural domains (for example, the frequency domain).
We evaluate CENDRe on synthetic time-series with ground-truth concepts and on two natural bearing-fault datasets~\cite{lessmeier2016condition,smith2015rolling}.
Across all settings, CENDRe extracts concepts in both the time and frequency domains, while the baselines are inherently limited to the time domain.
On time-domain concepts, CENDRe outperforms the baselines on importance correctness while recovering the same ground-truth concepts.
On frequency-domain concepts, CENDRe recovers the ground-truth bands on synthetic data and extracts bands that include the characteristic fault frequencies experts use to identify each fault type on real bearing-fault data~\cite{lessmeier2016condition,smith2015rolling}.

To summarize, the main contributions in this work are:
\begin{enumerate}
  \item \textbf{CENDRe.}
  We introduce a CE method for time-series CNNs that, to our knowledge, is the first to extract and visualize concepts in both the time and frequency domains.

  \item \textbf{Concept localization mechanism that transfers across domains.}
  We backpropagate gradients of a novel contrastive presence score through differentiable invertible transforms of the input, yielding concept masks in any natural domain.

  \item \textbf{Automatic concept number selection.}
  We develop a procedure that automatically estimates the number of relevant concepts $K$ from the latent structure of the data, by overestimating and then aggregating clusters. This replaces the manual choice of $K$ with the micro-cluster count $J\gg K$, to which the extraction is substantially less sensitive.

  \item \textbf{Evaluation on synthetic and real datasets.}
  On synthetic time-series, CENDRe substantially outperforms prior CE baselines on importance correctness and uniquely recovers ground-truth concepts in the frequency domain alongside the time domain. 
  On real bearing-fault datasets, it identifies the characteristic fault frequencies experts use for diagnosis.
\end{enumerate}

%% file: text/20_related_work.tex
\section{Related Work}

Concept-based explanations describe a model through human-interpretable patterns that correspond to directions or regions in its latent space~\cite{poeta2023concept}.
Concept Extraction (CE) discovers such patterns automatically from data, without requiring a predefined example set.
In the image domain, established CE methods include ACE~\cite{ghorbani2019automatic},
 ECLAD~\cite{posada-moreno2023eclad}, ConceptSHAP~\cite{yeh2020completenessaware}, and PACE~\cite{kamakshi2021pace}.
In time-series, prior concept-based work has applied concept \emph{testing} with TCAV~\cite{kim2018interpretability} and manually curated concept sets~\cite{madsen2023conceptbased,decker2023explaining}. 
Only two recent works perform concept \emph{extraction}: ECLAD-ts~\cite{holzapfel2025concept} clusters per-timestep local aggregated descriptors at a probe layer,
and MultiVISION~\cite{younis2022multivariate} clusters segments defined by the effective receptive field (ERF) of highly activated neurons.
However, both methods are contrained to the time domain for explanations. 
CENDRe is a CE method for time-series CNNs that extracts concepts across natural domains, including the frequency domain.

Despite the central role of frequency domain analysis in signal processing and related areas~\cite{oppenheim1999discrete,aastrom2021feedback}, no CE method operates in this domain.
Spectral attribution methods, on the other hand, produce local explanations directly in the frequency domain either via perturbation or gradients. 
Perturbation methods in the frequency or time-frequency domain~\cite{chung2024time,brusch2024explaining} probe the model through targeted alterations of the spectrum, 
and virtual inspection layers~\cite{vielhaben2024explainable} insert a differentiable, invertible transform into the forward graph 
so that attribution can be propagated to any chosen signal representation. 
These methods are local, explaining individual predictions rather than global model behavior. 
CENDRe reformulates the virtual inspection layer mechanism to bring global concept explanations into the frequency domain, which, to our knowledge, has not been done before.

In CE the number of concepts $K$ is typically fixed manually through sweeps or visual inspection~\cite{ghorbani2019automatic,kamakshi2021pace,younis2022multivariate,posada-moreno2023eclad,holzapfel2025concept}. 
In general clustering, by contrast, silhouette analysis has long been proposed as a criterion for selecting $K$~\cite{rousseeuw1987silhouettes},
alongside alternatives such as density-based methods~\cite{campello2013density}, which we include as variants. 
To our knowledge, CENDRe is the first CE method to determine the number of concepts automatically, 
 guided by silhouette scores, and thus reducing the inductive bias introduced by manual concept number selection. 

To localize concepts in the input, existing time-series CE methods typically rely on two heuristics:
either upsampling low-resolution activation maps to the input size~\cite{posada-moreno2023eclad,holzapfel2025concept}, 
or taking the full ERF of the probe layer as a concept patch~\cite{younis2022multivariate}. 
Upsampling attributes importance to the aggregation position even though its activation is driven by the full ERF, while the full-ERF patch can include regions the model did not meaningfully rely on.
CENDRe avoids both caveats by introducing visualizations based on gradients with respect to the latent representation of the concept, propagated through virtual inspection layers,
so that the attribution is concentrated on the regions that influence the concept, yielding compact localizations across domains.

%% file: text/40_methods.tex
\section{Concept Extraction with Natural Domain Representations (CENDRe)} \label{sec:methods}

CENDRe is a CE method for time-series CNNs that produces explanations across natural domains with a data-driven number of concepts.
Realizing this rests on three technical contributions: 
(i) discovering the number of concepts $K$ from the data rather than imposing it, 
(ii) localizing each concept on the regions the model actually uses, and 
(iii) extending these localizations to natural domains beyond the input.
The paragraphs below recall the per-timestep latent representation we build on, address each of these challenges, and conclude with a class-wise concept-importance score.

\paragraph{Latent representation via LADs.}
For concept discovery, we adapt LADs~\cite{posada-moreno2023eclad,holzapfel2025concept},
 timestep-wise aggregations of activations across layers, as they capture patterns at multiple levels of abstraction and are approximately position-invariant thanks to the translation equivariance of CNNs.
%
For building such a per-timestep descriptor, we first consider a trained 1D CNN $f:\mathbb{R}^{T\times D}\to\mathbb{R}^{n_\mathrm{out}}$ for supervised classification or regression on multivariate time-series of length $T$ with $D$ channels and $n_\mathrm{out}$ targets.
We select a subset $\mathcal{L}$ of its internal layers as probe layers.
Concept discovery operates on a dataset $\mathcal{D}\subset\mathbb{R}^{T\times D}$.
For each input sample $x\in\mathcal{D}$, $a_\ell(x)\in\mathbb{R}^{T_\ell\times D_\ell}$ denotes the activation map at probe layer $\ell\in\mathcal{L}$.
We linearly upsample each activation map to the longest probe length $T^*$ via $f_U$. 
For each $x\in\mathcal{D}$ and timestep $t\in\{0,\dots,T^*-1\}$, the LAD $z_t(x)$ is the per-timestep slice of the concatenated upsampled maps,
\begin{equation}
z_t(x) \;=\; \big[\,f_U(a_{\ell_1}(x))[t]\;\dots\;f_U(a_{\ell_{|\mathcal{L}|}}(x))[t]\,\big]\in\mathbb{R}^{D^*},\qquad D^*\coloneqq\textstyle\sum_{\ell\in\mathcal{L}}D_\ell.
\end{equation}
Thus, similar LADs correspond to similar local evidence across the selected layers, and the collection of LADs over the dataset is the input to the first clustering stage.

\paragraph{Concept discovery via clustering.}
CENDRe discovers latent concepts by clustering LADs $z_t(x)$ with agglomerative or density-based methods, 
which select the number of concepts $K$ automatically and can capture non-spherical clusters expected in high-dimensional latent spaces.
Applying such methods directly to all LADs is infeasible at scale, so we first compress LADs to micro-centroids, and then cluster those micro-centroids into concepts.

In the first stage, mini-batch $k$-means on LADs produces a set of $J\gg K$ micro-centroids $\Gamma=\{\mu_j\}_{j=0}^{J-1}$ with $\mu_j\in\mathbb{R}^{D^*}$, and we assign each LAD to its nearest micro-centroid,
\begin{equation}
\pi(x,t) \;\coloneqq\; \argmin_{j\in\{0,\dots,J-1\}} \lVert z_t(x)-\mu_j\rVert_2^2 .
\end{equation}
In the second stage, we cluster the $J$ micro-centroids into $K$ concepts via a mapping $\kappa:\{0,\dots,J-1\}\to\{0,\dots,K-1\}$,
so each concept $C_k=\{\mu_j\mid\kappa(j)=k\}$ is a group of micro-centroids capturing similarly encoded patterns.

By default, we use silhouette-guided hierarchical aggregation under cosine distance,
with $K$ the value maximizing the average silhouette score across micro-centroids (CENDRe$_{\text{silhouette}}$)~\cite{mullner2011modern,rousseeuw1987silhouettes}. 
We additionally evaluate two variants: $k$-means~\cite{sculley2010web} directly on LADs with predefined $K$ (CENDRe$_{k\text{Means}}$),
and HDBSCAN~\cite{campello2013density} on the micro-centroids (CENDRe$_{\text{HDBSCAN}}$).
See App.~\ref{app:cendre-clustering} for hyperparameters.


\paragraph{Concept masks.}
We localize each concept on each input through gradients with respect to an interpretable domain.
Gradients concentrate the attribution on regions that relate to the concept, rather than covering the full receptive field or shrinking past the relevant evidence within it. 
Pulling the same gradient through any invertible differentiable transform allows the visualization of the concept in the corresponding domain (e.g.\ the input domain or the frequency domain).

We propose a presence score, defined below, which captures what makes a LAD belong to concept $C_k$ by contrasting its similarity to the assigned micro-centroid with the mean similarity to micro-centroids of the other concepts.
To express this contrast, we partition the micro-centroid indices into the in-concept set $\mathcal{J}_k\coloneqq\kappa^{-1}(k)\subseteq\{0,\dots,J-1\}$ and its complement $\bar{\mathcal{J}}_k\coloneqq\{0,\dots,J-1\}\setminus\mathcal{J}_k$, and we collect the timesteps where the LAD of $x$ is in $C_k$ in $T_k(x)\coloneqq\{t\in\{0,\dots,T^*-1\}\mid\pi(x,t)\in\mathcal{J}_k\}$.
To assess how similar a LAD is to a micro-centroid, we use the inverse-decay similarity $d(z,\mu)\coloneqq(1+\lVert z-\mu\rVert_2)^{-1}\in(0,1]$ between a LAD $z$ and a micro-centroid $\mu$.
For every $t\in T_k(x)$, the presence score is then
  \begin{equation}
\rho_k(x,t) \;\coloneqq\; d\!\bigl(z_t(x),\,\mu_{\pi(x,t)}\bigr) \;-\; \frac{1}{|\bar{\mathcal{J}}_k|}\sum_{j\in\bar{\mathcal{J}}_k} d\!\bigl(z_t(x),\,\mu_j\bigr).
  \end{equation}
A high presence score indicates strong concept membership: the LAD is close to its assigned micro-centroid and far from those of the other concepts.
We aggregate it over $T_k(x)$ to obtain a concept presence per sample
\(\bar\rho_k(x)\coloneqq |T_k(x)|^{-1}\sum_{t\in T_k(x)}\rho_k(x,t)\),
with $\bar\rho_k(x)=0$ when $T_k(x)=\emptyset$.

To obtain a mask in an arbitrary domain we follow \emph{virtual inspection layers}~\cite{vielhaben2024explainable}: 
a fixed, differentiable, invertible transform $\phi$ is inserted into the computational graph as $\phi^{-1}\circ\phi$ for attribution only, leaving the model's forward mapping unchanged.
We instantiate $\phi$ in two ways: the frequency domain ($\phi=\mathcal{F}$),
where $\mathcal{F}:\mathbb{R}^{T\times D}\to\mathbb{C}^{F\times D}$ is the real-input fast Fourier transform (rFFT) with $F\coloneqq\lfloor T/2\rfloor+1$ and $x_f\coloneqq\mathcal{F}(x)$,
and the time domain, where $\phi$ is the identity. 
App.~\ref{sec:app-domains} instantiates two further choices in time-frequency domains. 
Through the inserted inverse transform, $\bar\rho_k$ becomes a function of $x_f$, and $\nabla_{x_f}\bar\rho_k$ denotes its gradient at $x_f=\phi(x)$.
The masks become the sensitivities of $\bar\rho_k$ to the input and to the log-power-spectrum, respectively (derivation in App.~\ref{sec:app-spectral-derivation}),
\begin{equation}
m_k^{\mathrm{time}}(x) \coloneqq \mathrm{ReLU}\!\bigl(\nabla_x \bar\rho_k(x)\bigr),
\qquad
m_k^{\mathrm{freq}}(x) \coloneqq \mathrm{ReLU}\!\bigl(\Re\!\bigl(\overline{x_f}\odot\nabla_{x_f}\bar\rho_k(x)\bigr)\bigr).
\label{eq:masks}
\end{equation}
The ReLU keeps the components at which amplifying the signal, or its spectral power, increases the concept's presence.
Negative components mark evidence against the concept and are discarded for localization.
Although LADs are indexed on the upsampled grid $t\in\{0,\dots,T^*-1\}$, masks are obtained by backpropagation to the original input $x\in\mathbb{R}^{T\times D}$ or its rFFT image $x_f\in\mathbb{C}^{F\times D}$.
For visualization, a noise tunnel is applied on the masks~\cite{smilkov2017smoothgrad}
and they are smoothed by a moving average along the domain axis.


\paragraph{Concept importance scores.}
For each class $c$, concept $C_k$, and input channel $\mathrm{ch}$, we ask whether the presence of $C_k$ on channel $\mathrm{ch}$ pushes the model toward or away from predicting $c$, and by how much.
The importance score $I_{c,k,\mathrm{ch}}\in[-1,1]$ encodes this: magnitude reports the strength of the effect.
The sign on its own carries no meaning, but opposite signs across classes indicate that the concept pushes those classes apart.
We compute $I_{c,k,\mathrm{ch}}$ from the gradient of the per-class score $y_c$ with respect to the input, weighted by the concept mask of $C_k$ on channel $\mathrm{ch}$.
For classification, $y_c(x)$ is the logit $f(x)_c$.
For regression, $y_{c}(x)$ is the $c$-th regression output.

We measure the instance-level relevance as the overlap between the regions where the concept lives and the regions to which $y_c$ is sensitive.
Let $m^{\mathrm{dom}}_{k,\mathrm{ch}}(x)$ and $s^{\mathrm{dom}}_{c,\mathrm{ch}}(x)$ denote the channel slices of the mask and the sensitivity of $y_c$ in domain $\mathrm{dom}\in\{\mathrm{time},\mathrm{freq}\}$.
We write $x_{\mathrm{ch}}\in\mathbb{R}^T$ and $x_{f,\mathrm{ch}}\in\mathbb{C}^F$ for the time- and frequency-domain channels of $x$.
The relevance is then their inner product over the corresponding domain axis, $r_{c,k,\mathrm{ch}}(x) \;\coloneqq\; \big\langle m^{\mathrm{dom}}_{k,\mathrm{ch}}(x),\; s^{\mathrm{dom}}_{c,\mathrm{ch}}(x) \big\rangle .$

The sensitivity $s^{\mathrm{dom}}_{c,\mathrm{ch}}(x)$ is obtained by backpropagation in the chosen domain: $s^{\mathrm{time}}_{c,\mathrm{ch}}(x)=\partial y_c(x)/\partial x_{\mathrm{ch}}$, and $s^{\mathrm{freq}}_{c,\mathrm{ch}}(x)=\Re\!\bigl(\overline{x_{f,\mathrm{ch}}}\odot \nabla_{x_{f,\mathrm{ch}}} y_c(x)\bigr)$ (App.~\ref{sec:app-spectral-derivation}).
We smooth $s^{\mathrm{dom}}_{c,\mathrm{ch}}(x)$ with a noise tunnel to damp single-point gradient noise.
Then, we average $r_{c,k,\mathrm{ch}}(x)$ over $\mathcal{D}_{k,\mathrm{ch}}\coloneqq\{\,x\in\mathcal{D}\mid m^{\mathrm{dom}}_{k,\mathrm{ch}}(x) \neq 0\,\}$, the samples on which $C_k$ leaves a non-zero mask on channel $\mathrm{ch}$, to obtain a dataset-level estimate of the concept's effect.
If $\mathcal{D}_{k,\mathrm{ch}}=\emptyset$, we set the expectation to $0$.
We divide the result by the largest such average across all $(c',k',\mathrm{ch}')$ to make entries comparable on $[-1,1]$:
\[
I_{c,k,\mathrm{ch}} \;=\; \frac{\mathbb{E}_{x\sim \mathcal{D}_{k,\mathrm{ch}}}[\,r_{c,k,\mathrm{ch}}(x)\,]}{\max_{c', k',\mathrm{ch}'}\bigl|\mathbb{E}_{x\sim \mathcal{D}_{k',\mathrm{ch}'}}[\,r_{c',k',\mathrm{ch}'}(x)\,]\bigr|+\varepsilon},
\]
for a small $\varepsilon>0$.


To obtain a single, class-agnostic importance value per concept--channel pair (as required by downstream metrics), we contrast the most-impacted class against the mean of the rest,
\begin{equation}
I_{k,\mathrm{ch}} \;=\; 0.5 \left|I_{c^\star(k,\mathrm{ch}),\,k,\mathrm{ch}} \;-\; \frac{1}{n_\mathrm{out}-1}\sum_{c\neq c^\star(k,\mathrm{ch})} I_{c,k,\mathrm{ch}} \right| \;\in\; [0,1],
\label{eq:Ikch}
\end{equation}
with $c^\star(k,\mathrm{ch})\coloneqq\arg\max_{c}|I_{c,k,\mathrm{ch}}|$. 
For $n_\mathrm{out}=1$, we set $I_{k,\mathrm{ch}}\coloneqq |I_{0,k,\mathrm{ch}}|$.
Values close to $1$ identify concepts that selectively drive the prediction for at least one class, 
whereas values close to $0$ mark concepts whose contribution is either negligible or uniform across classes. 

%% file: text/45_experiment_design.tex
\section{Experimental design}
\label{sec:experiment-design}


A CE method should identify the recurring patterns a model learns, how they relate to the input, and how they influence the predictions.
Yet, there is no direct ground truth on what a CNN relies on.
Thus, we evaluate CE methods on synthetic datasets where each input is annotated with the masks of its class-discriminative primitives, assuming a model trained on these datasets relies on them.
A faithful CE method should produce concepts tied to those primitives (\emph{representation}) and assign higher importance to concepts on discriminative primitives than to unrelated ones (\emph{importance}).

We compare CE methods by running each through multiple trained models and measuring representation and importance on the resulting concepts.
In an experimental run, we train a CNN on a synthetic dataset, run the CE method on the same data to obtain concepts with time- and frequency-domain masks and per-class importance scores, and score these concepts on the validation data against the primitive masks.
We repeat across datasets and seeds, then compare CE methods on representation and importance correctness on the synthetic data, and qualitatively on the real bearing-fault datasets CWRU and BearingPD by inspecting how concepts relate to characteristic fault frequencies.

\subsection{Evaluation metrics}
\label{sec:metrics}


To evaluate a CE method, we assess how much the resulting concepts align with the primitives through two correctness scores: 
\emph{representation}, measuring how well concept masks align with primitive masks, 
and \emph{importance}, measuring whether each concept's importance score reflects its primitive's function in the dataset.
To this end, we propose two modifications over the evaluation procedure of~\cite{posada-moreno2023eclad,holzapfel2025concept}: 
We score concept alignment with the \emph{Relevance Mass Accuracy} rather than the EDT score, and we compute the correctness metrics using \emph{soft} variants (sRC, sIC) that avoid the \textit{sensitivity to an alignment threshold} of the original Representation and Importance Correctness scores.



\paragraph{Alignment.}
For the alignment comparison between concepts and primitives to make sense, all masks must be on the same scale.
CENDRe masks $m_k^{\mathrm{dom}}(x)$ can take any real value, so we normalize them per instance to $\tilde m_k^{\mathrm{dom}}(x)\in[0,1]$ by clamping negatives to zero and dividing by the maximum, while baseline masks are already bounded or binary.
On the ground-truth side, $g_p^{\mathrm{dom}}(x)\in\{0,1\}$ is the binary mask of primitive $p$, with $n_p$ primitives and $K$ extracted concepts in total.
We score the alignment with the \emph{Relevance Mass Accuracy} (RMA)~\cite{arras2022clevr}, which for a continuous mask $u\in\mathbb{R}^{|\mathrm{dom}|\times D}$ and binary reference $v\in\{0,1\}^{|\mathrm{dom}|\times D}$ (with $|\mathrm{time}|=T$ and $|\mathrm{freq}|=F$) measures the fraction of $u$'s positive energy that falls inside $v$,
\begin{equation}
\mathrm{RMA}(u,v) = \frac{\sum_{t,\mathrm{ch}} \mathrm{ReLU}(u_{t,\mathrm{ch}})\cdot \mathbf{1}[v_{t,\mathrm{ch}}>0]}{\sum_{t,\mathrm{ch}} \mathrm{ReLU}(u_{t,\mathrm{ch}}) + \varepsilon}\;\in\;[0,1],
\label{eq:rma}
\end{equation}
where $\varepsilon>0$ penalizes empty masks. 
We choose RMA over the EDT score because it accepts continuous masks directly and tolerates models that use only part of a primitive's support.



\paragraph{Representation and Importance metrics.}
With the concept and primitive alignments computed per instance, we aggregate them into one score per property.
Following~\cite{holzapfel2025concept}, we treat each pair $(k,\mathrm{ch})$ of concept and channel as its own evaluation unit, and define its alignment to a pair $(p,\mathrm{ch})$ of primitive and channel as the average RMA over the dataset, $A_{p,k,\mathrm{ch}}\coloneqq\mathbb{E}_{x\sim\mathcal{D}}[\mathrm{RMA}(\tilde m_{k,\mathrm{ch}}^{\mathrm{dom}}(x),\,g_{p,\mathrm{ch}}^{\mathrm{dom}}(x))]\in[0,1]$.
For each primitive, \emph{Soft Representation Correctness} (sRC) picks the concept whose alignment summed across channels is highest, and averages this over primitives and channels.
\emph{Soft Importance Correctness} (sIC) instead weights each concept's class-agnostic importance $I_{k,\mathrm{ch}}$ from Eq.~\eqref{eq:Ikch} by its best alignment to any primitive,
\begin{equation}
\mathrm{sRC} \coloneqq \frac{1}{n_p D}\sum_{p} \max_{k} \sum_\mathrm{ch} A_{p,k,\mathrm{ch}},
\qquad
\mathrm{sIC} \coloneqq \frac{\sum_{k,\mathrm{ch}} \bigl(\max_{p} A_{p,k,\mathrm{ch}}\bigr)\, I_{k,\mathrm{ch}}}{\sum_{k,\mathrm{ch}} |I_{k,\mathrm{ch}}|}\;\in\;[0,1].
\label{eq:src_sic}
\end{equation}

\subsection{Experimental setup}
\label{sec:setup}

\paragraph{Models and training.}
\label{sec:models}
We train three 1D CNN architectures: InceptionTime10~\cite{ismail2020inceptiontime}, ResNet1D-18~\cite{he2016deep}, 
and DenseNet1D-121~\cite{huang2017densely}.
Each model takes the standard hyperparameters of its reference implementation, with full configurations listed in App.~\ref{app:models-arch}.
We optimize the negative log-likelihood using AdamW~\cite{loshchilov2017decoupled} (learning rate $10^{-4}$, weight decay $10^{-7}$) with early stopping on the validation loss (patience~$20$, up to $500$~epochs).
Each dataset is split 80/20 with balanced sampling, and every training configuration is repeated over $11$~seeds ($0$--$10$).

\paragraph{Datasets.}
\label{sec:datasets}
We use two synthetic binary-classification families with known ground-truth primitive masks: \textsc{syntheticLocal}, 
which injects localized temporal shapes (square, circle, triangle) into a noisy sinusoidal baseline, 
and \textsc{syntheticFrequency}, which places class-discriminative peaks in specific frequency bands (mid, mid-high, high).
For real-world signals we use \emph{CWRU Bearing}~\cite{smith2015rolling},
a rolling-element bearing fault dataset with four classes (healthy plus three fault locations) and two accelerometer channels (fan end and drive end).
Both channels are sampled at $f_s = 12$\,kHz and segmented into windows of $T = 2048$ samples with stride $1024$ ($50\%$ overlap).
The full dataset list is in Table~\ref{tab:datasets} of App.~\ref{app:datasets}, 
with additional synthetic variants and a second bearing-fault benchmark, which are evaluated in App.~\ref{sec:app-additional}.

\paragraph{Concept extraction hyperparameters.}
\label{sec:ce-hparams}
All CE methods are implemented in PyTorch~\cite{pytorch} and require probe layers per model, listed in Table~\ref{tab:layers} of App.~\ref{app:models}.
Each method runs on $256$ samples with the same random seed as the corresponding model training, is fit on the model's training partition and evaluated on its validation partition.
The methods differ in how the number of concepts $K$ is set.
For fixed-$K$ methods (ECLAD-ts, MultiVISION, CENDRe$_{k\text{Means}}$) we sweep over $K\in\{2,4,6,8,10\}$ and report the configuration with the highest sRC.
For methods that determine $K$ automatically (CENDRe$_{\text{silhouette}}$, CENDRe$_{\text{HDBSCAN}}$) we take $50$ micro-clusters, a choice whose effect the sensitivity analysis in App.~\ref{sec:app-sensitivity} quantifies.
CENDRe's noise tunnel uses $5$ samples with Gaussian perturbations of standard deviation $0.01$ for instance masks, $0.1$ for logit sensitivities,
and a moving average of $5$ for visualizations. \footnote{The code is freely available at https://github.com/antoHolz/CENDRe/}

%% file: text/50_results.tex
\section{Empirical results} \label{sec:results}
We report results on the \textsc{syntheticLocal} and \textsc{syntheticFrequency} dataset families, and on the natural bearing-fault dataset CWRU.
The experimental setup follows Section~\ref{sec:experiment-design}, with further evaluations in App.~\ref{sec:app-additional}.
The main findings of our experiments are: 
(i) On \textsc{syntheticLocal}, CENDRe extracts concept regions that match the ground truth qualitatively, with representation correctness similar to ECLAD-ts but higher importance correctness.
(ii) Frequency-domain concepts on \textsc{syntheticFrequency} localize in the ground-truth bands, with high representation and importance correctness.
(iii) Concepts extracted from CWRU separate cleanly by class and align with characteristic fault frequencies, evidencing that CENDRe transfers from controlled synthetic settings to real signals.

\begin{figure*}[!ht]
    \begin{minipage}{0.1\linewidth}
         \rotatebox{90}{\scriptsize CENDRe$_\textrm{silhouette}$}
    \end{minipage}
    \begin{minipage}{0.8\linewidth}
        \begin{subfigure}[t]{\linewidth}
            \includegraphics[width=\linewidth]{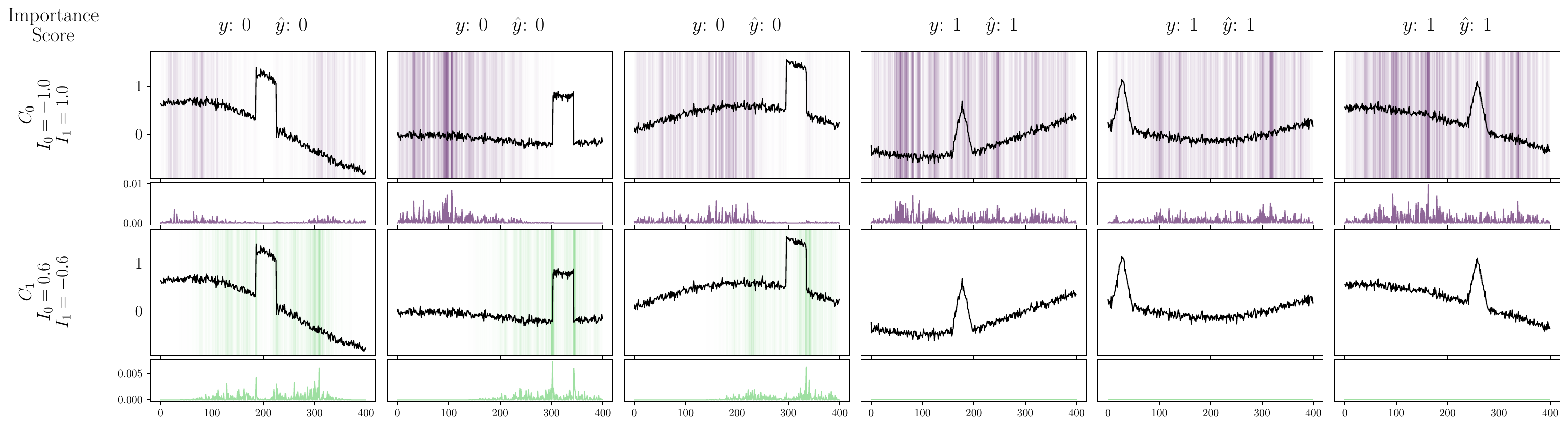}
        \end{subfigure}
    \end{minipage}

    \begin{minipage}{0.1\linewidth}
         \rotatebox{90}{\scriptsize ECLAD-ts}
    \end{minipage}
    \begin{minipage}{0.8\linewidth}
        \begin{subfigure}[t]{\linewidth}
            \includegraphics[width=\linewidth]{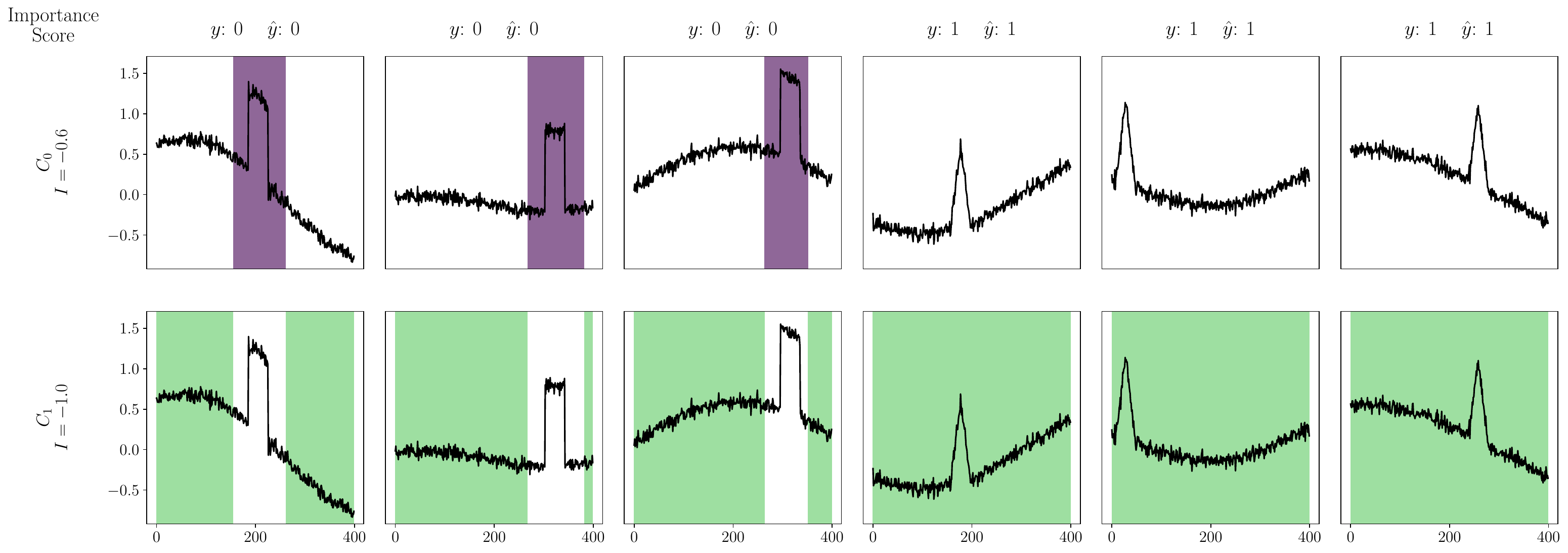}
        \end{subfigure}
    \end{minipage}

    \begin{minipage}{0.1\linewidth}
         \rotatebox{90}{\scriptsize MultiVISION}
    \end{minipage}
    \begin{minipage}{0.8\linewidth}
        \begin{subfigure}[t]{\linewidth}
            \includegraphics[width=\linewidth]{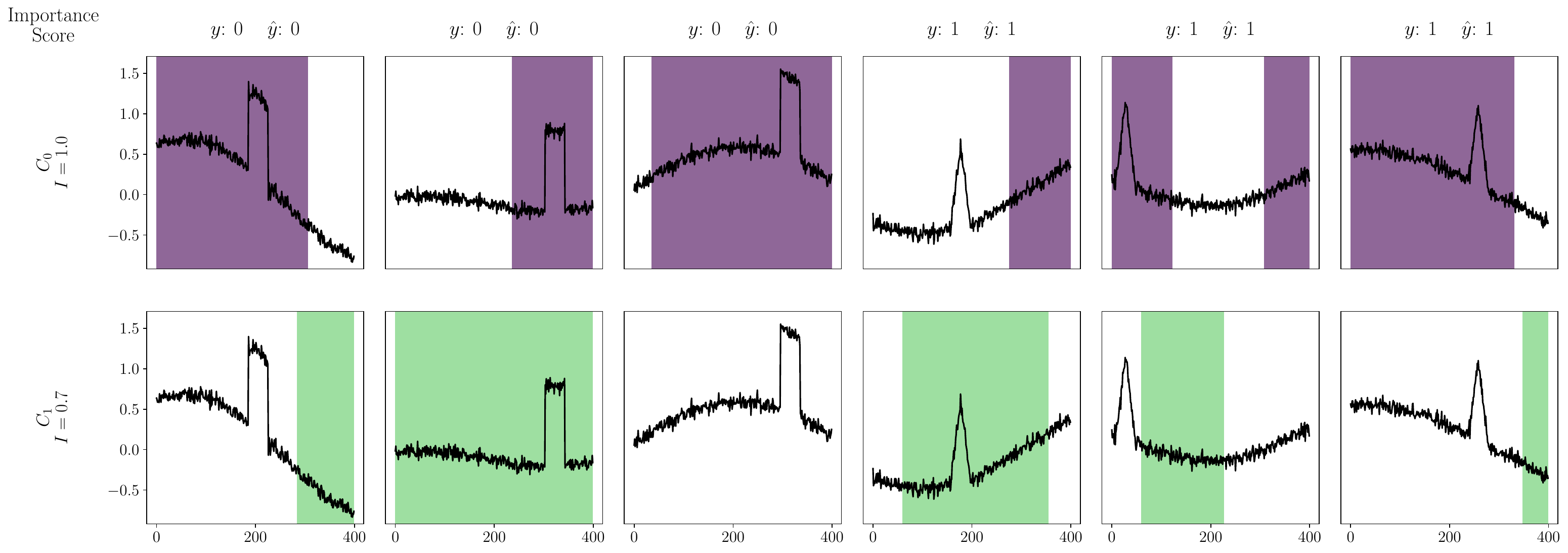}
        \end{subfigure}
    \end{minipage}

    \begin{minipage}{0.1\linewidth}
         \rotatebox{90}{\scriptsize Primitives}
    \end{minipage}
    \begin{minipage}{0.055\linewidth}
         \phantom{asdsa}
    \end{minipage}
    \begin{minipage}{0.74\linewidth}
        \begin{subfigure}[t]{\linewidth}
            \includegraphics[width=\linewidth]{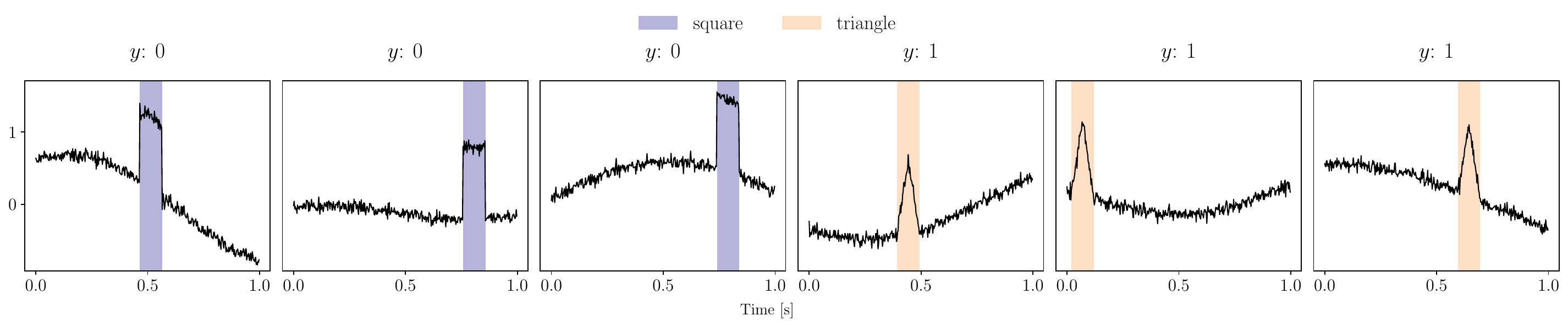}
        \end{subfigure}
    \end{minipage}
    \caption{Concepts extracted from a ResNet1D-18 trained on a \textsc{syntheticLocal} square-triangle dataset. Within each grid, rows are concepts, columns are samples, and row labels show importance scores. The bottom row shows the ground-truth primitive locations for the same samples. CENDRe and ECLAD-ts extract the same input features, while MultiVISION shows no coherent pattern.}
    \label{fig:syntheticLocal}
\end{figure*}%
\paragraph{Synthetic datasets.}
The \textsc{syntheticLocal} family tests whether the compared CE methods extract the local cues time-series CNNs rely on for classification.
Figure~\ref{fig:syntheticLocal} shows CE on a representative square-triangle variant from this family.
ECLAD-ts highlights the complete square defined by the ground-truth primitive. 
CENDRe picks up the same regions with importance scores of similar magnitude, but weights the corners more strongly, where the signal differs most recognizably from the baseline.
MultiVISION, by contrast, returns no coherent pattern.
Overall, CENDRe extracts the local cues without tuning the number of concepts, and additionally highlights the most important structures within each concept.
Figure~\ref{fig:syntheticLocalVariants} of App.~\ref{sec:app-variants} repeats this comparison on two further architectures and the two remaining CENDRe variants, with the same outcome.

\begin{figure*}[hbt]
    \begin{minipage}{.49\linewidth}
        \begin{subfigure}[t]{\linewidth}
            \includegraphics[width=\linewidth]{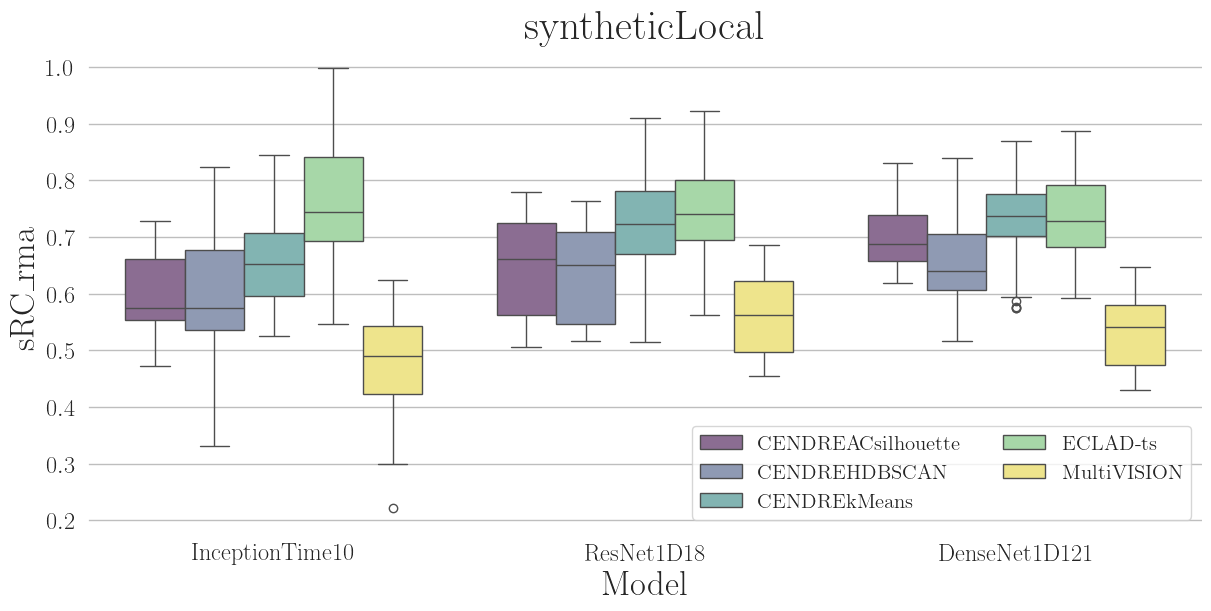}
        \end{subfigure}
    \end{minipage}%
    \begin{minipage}{.49\linewidth}
        \begin{subfigure}[t]{\linewidth}
            \includegraphics[width=\linewidth]{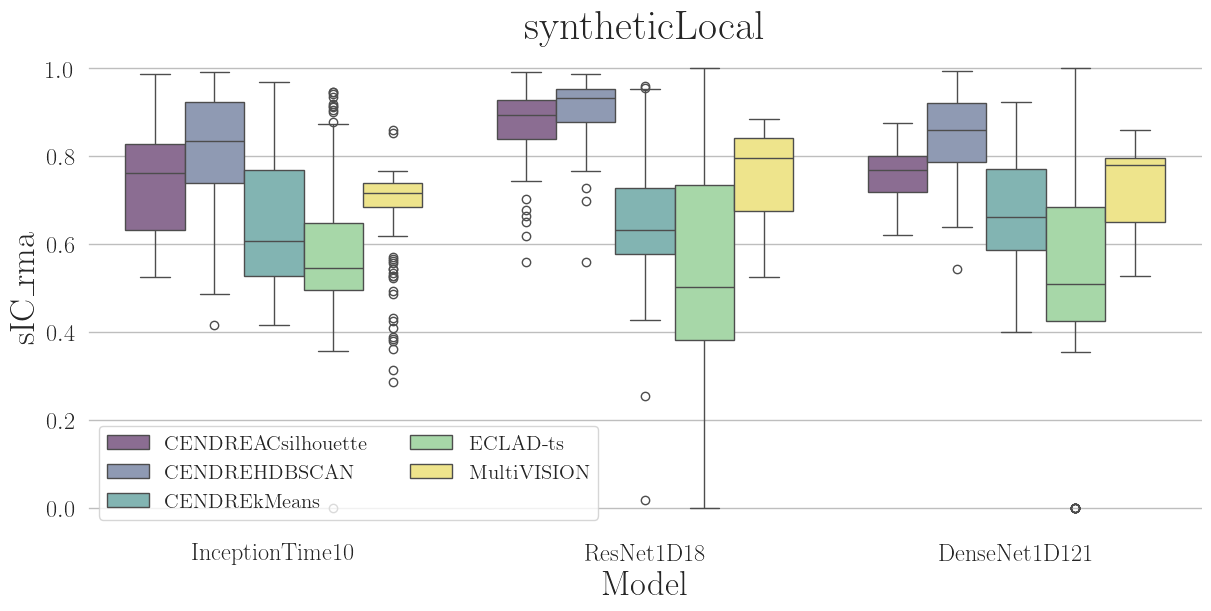}
        \end{subfigure}
    \end{minipage}
    \begin{minipage}{.49\linewidth}
        \begin{subfigure}[t]{\linewidth}
            \includegraphics[width=\linewidth]{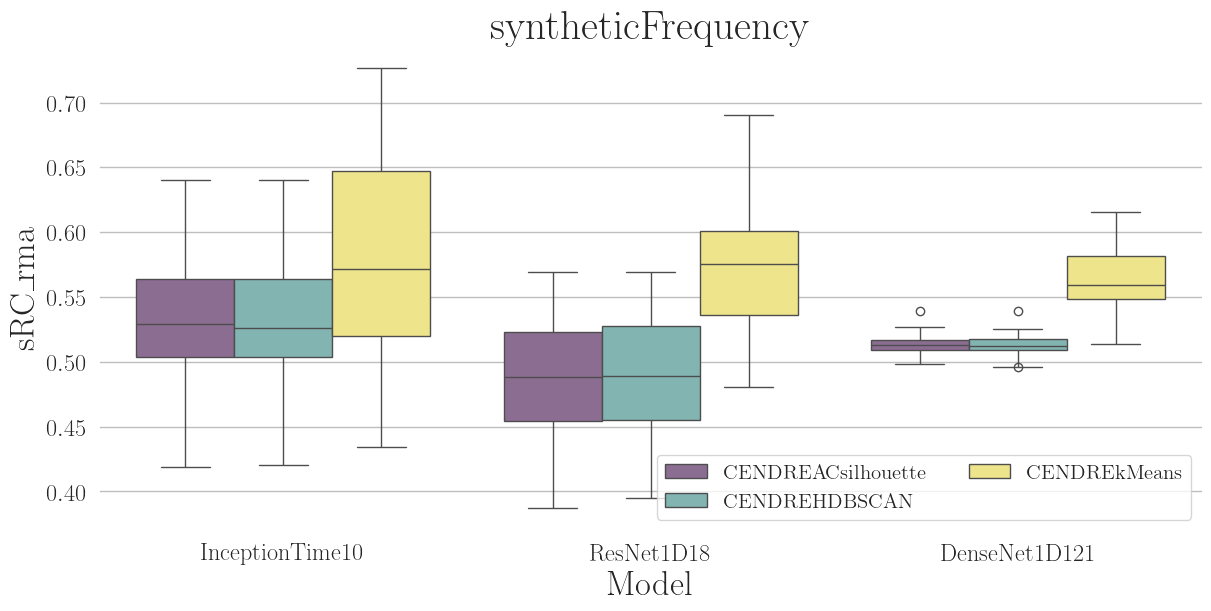}
        \end{subfigure}
    \end{minipage}%
    \begin{minipage}{.49\linewidth}
        \begin{subfigure}[t]{\linewidth}
            \includegraphics[width=\linewidth]{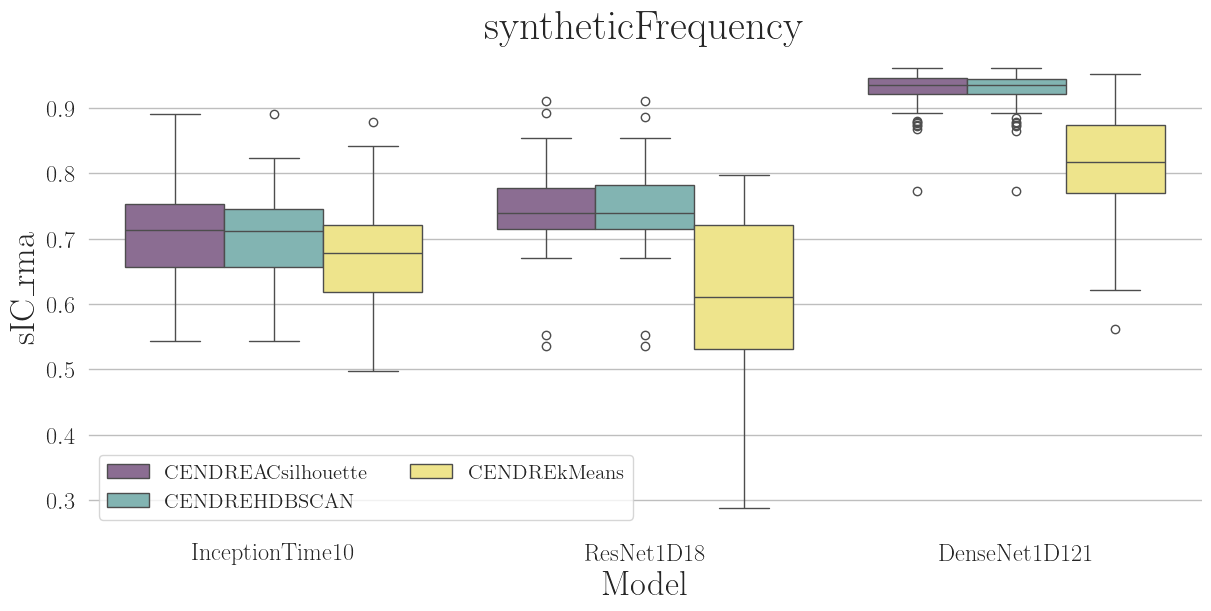}
        \end{subfigure}
    \end{minipage}
    \caption{Boxplots of sRC (left) and sIC (right) on \textsc{syntheticLocal} (top) and \textsc{syntheticFrequency} (bottom); higher is better. On \textsc{syntheticLocal}, CENDRe achieves sRC close to ECLAD-ts and higher sIC, while MultiVISION is weak on both metrics. On \textsc{syntheticFrequency}, CENDRe variants reach sRC and sIC values close to those on \textsc{syntheticLocal}.}
    \label{fig:quant}
\end{figure*}

The quantitative metrics in Figure~\ref{fig:quant} (top row) are consistent with this qualitative picture.
For the CENDRe$_{\text{silhouette/HDBSCAN}}$ variants, sRC is slightly lower than ECLAD-ts's on \textsc{syntheticLocal} ($k$-means is essentially on par).
Two factors plausibly contribute.
First, the model exploits only a portion of the ground-truth primitive (a shortcut-learning pattern), and CENDRe$_{\text{silhouette/HDBSCAN}}$ stays faithful to that partial use, while ECLAD-ts's broader masks cover the full primitive.
Second, $k$-means and ECLAD-ts are reported at the $K$ with best sRC, which biases them toward human-recognizable concepts, whereas silhouette and HDBSCAN choose $K$ from the latent structure itself.
The latter is also visible in sIC, which is higher for the CENDRe$_{\text{silhouette/HDBSCAN}}$ variants, meaning their masks better match the model's per-timestep importance signal.
MultiVISION underperforms on both metrics, in line with the lack of coherent concept structure observed qualitatively.
The same ranking holds on the multichannel and confounded variants of the family (Figure~\ref{fig:quant_mc_conf} of App.~\ref{sec:app-mc-conf}). 
Per-dataset values and pairwise significance tests are in App.~\ref{sec:app-distributional} and App.~\ref{sec:app-significance}. 
Across CENDRe variants the differences are small qualitatively (the three variants point to the same regions).
Quantitatively CENDRe$_{\text{silhouette}}$ and CENDRe$_{\text{HDBSCAN}}$ perform similarly, while CENDRe$_{k\text{Means}}$ trades slightly higher sRC for slightly lower sIC, consistent with the $K$-selection effect noted above.
We favor the silhouette variant because it reaches comparable quality in a single run without a $K$ sweep, and picks a usable $K$ across datasets out of the box.
It is also the more robust of the two to its remaining hyperparameter, the number of micro-clusters $J$.
Over three orders of magnitude, $J$ explains $0.02$ of the sRC variance under CENDRe$_{\text{silhouette}}$ and $0.14$ under CENDRe$_{\text{HDBSCAN}}$, against $0.70$ for the random seed (App.~\ref{sec:app-sensitivity}), supporting our claim that silhouette-guided aggregation removes the dependence on a user-set number of concepts.

\begin{figure*}[htb]
    \begin{minipage}[c]{0.7\linewidth}
        \begin{minipage}[c]{0.14\linewidth}
            \centering\rotatebox{90}{\scriptsize Time domain}
        \end{minipage}%
        \begin{minipage}[c]{0.86\linewidth}
            \begin{subfigure}[t]{\linewidth}
                \includegraphics[width=\linewidth]{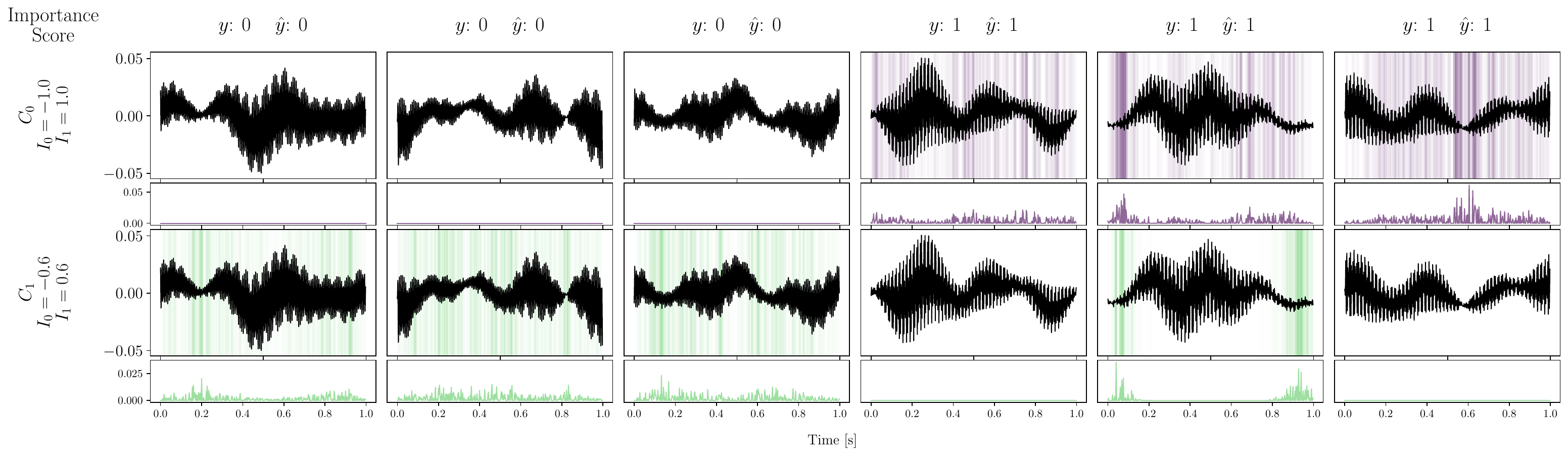}
            \end{subfigure}
        \end{minipage}

        \vspace{0.5em}

        \begin{minipage}[c]{0.14\linewidth}
            \centering\rotatebox{90}{\scriptsize Frequency domain}
        \end{minipage}%
        \begin{minipage}[c]{0.86\linewidth}
            \begin{subfigure}[t]{\linewidth}
                \includegraphics[width=\linewidth]{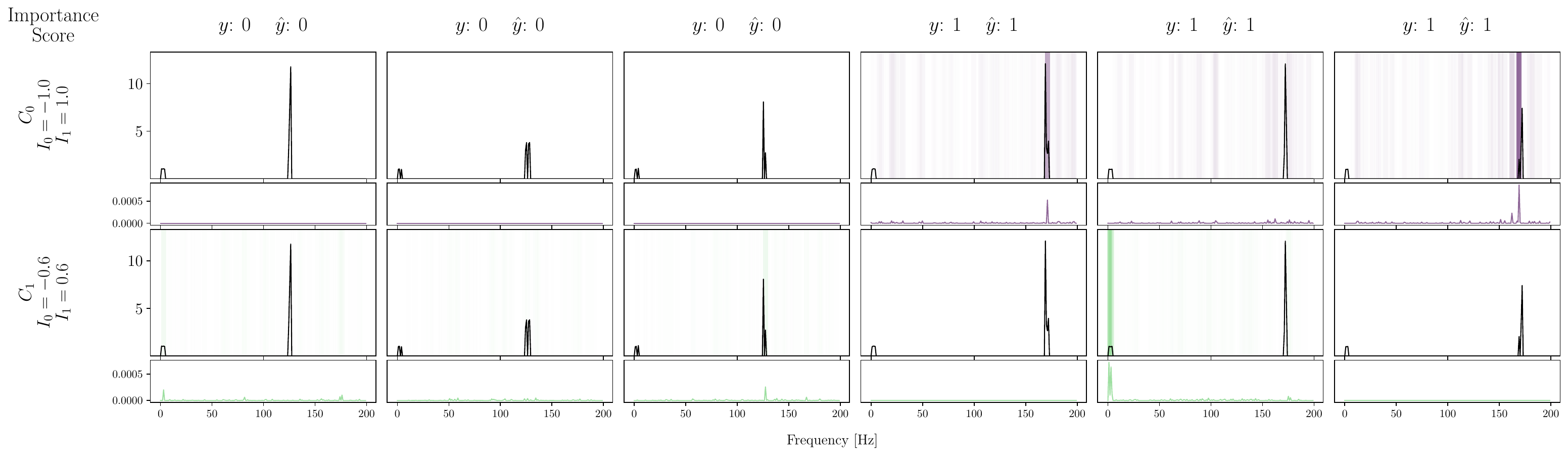}
            \end{subfigure}
        \end{minipage}
    \end{minipage}%
    \hfill
    \begin{minipage}[c]{0.25\linewidth}
        \begin{subfigure}[t]{0.85\linewidth}
            \includegraphics[width=\linewidth]{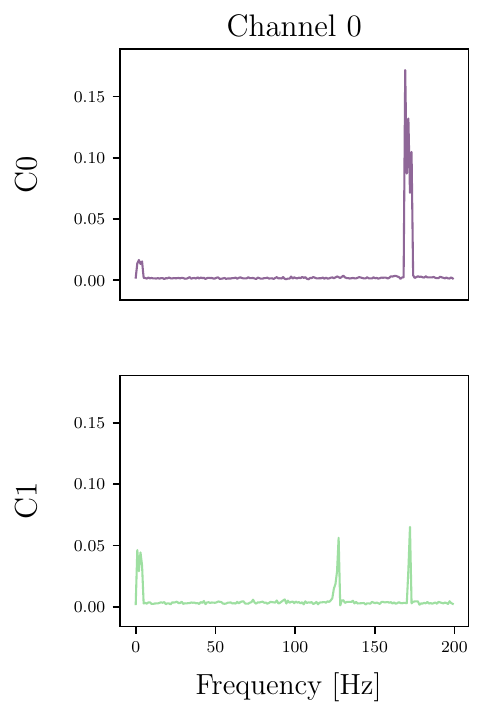}
        \end{subfigure}
    \end{minipage}
    \caption{Concepts extracted from a ResNet1D-18 trained on a \textsc{syntheticFrequency} \texttt{midhighFreq-highFreq} dataset. The two left grids show time-domain (top) and frequency-domain (bottom) masks; within each, rows are concepts, columns are samples, and row labels show importance scores. The right panel shows the global concept-frequency correspondence, obtained by aggregating the frequency masks across samples. Time-domain masks show no interpretable pattern, whereas frequency-domain masks localize in specific bands, as confirmed by the correspondence plot.}
    \label{fig:syntheticFrequency}
\end{figure*}

The \textsc{syntheticFrequency} family tests whether CE methods extract spectral cues exploited by the model.
Among the compared methods, only CENDRe extends to the frequency domain, through its virtual inspection layers, so the baselines are omitted from this analysis.
Figure~\ref{fig:syntheticFrequency} shows that the time-domain masks have no interpretable pattern, whereas the frequency-domain masks concentrate concept mass at specific bands, both per-sample and in the global concept correspondence.
In the \texttt{midhighFreq-highFreq} variant shown, $C_0$ aligns with a single high-frequency band, while $C_1$ mixes mid-high and near-zero content, reflecting how the model composes information from several frequency components.
When a model relies on frequency-related features, CENDRe captures them at both instance and global scale, turning uninterpretable time-domain masks into interpretable frequency explanations.
Quantitatively, Figure~\ref{fig:quant} (bottom row) shows the same CENDRe behavior as the local case.
CENDRe$_{\text{silhouette/HDBSCAN}}$ variants reach slightly lower sRC than $k$-means but higher sIC.
Mean sRC is slightly below \textsc{syntheticLocal}'s, while sIC is comparable.
CENDRe thus extracts concepts in the frequency domain at quality close to the time domain, showing that the localization transfers across domains.

\begin{figure*}[!tb]
    \begin{minipage}[c]{0.49\linewidth}
        \centering{\scriptsize Time domain}
            \begin{subfigure}[t]{\linewidth}
                \includegraphics[width=\linewidth]{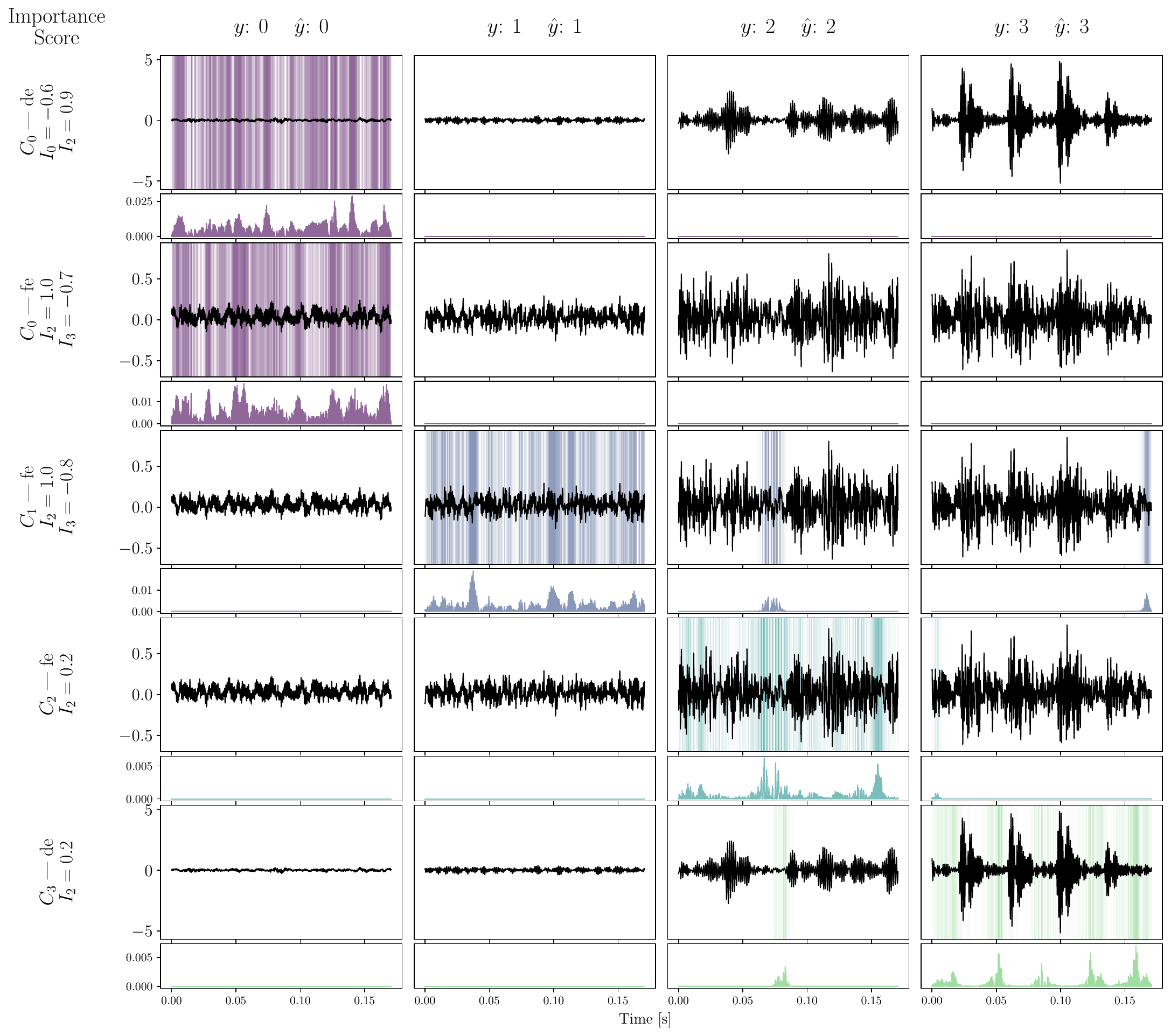}
            \end{subfigure}
    \end{minipage}%
    \begin{minipage}[c]{0.49\linewidth}
        \centering{\scriptsize Frequency domain}
            \begin{subfigure}[t]{\linewidth}
                \includegraphics[width=\linewidth]{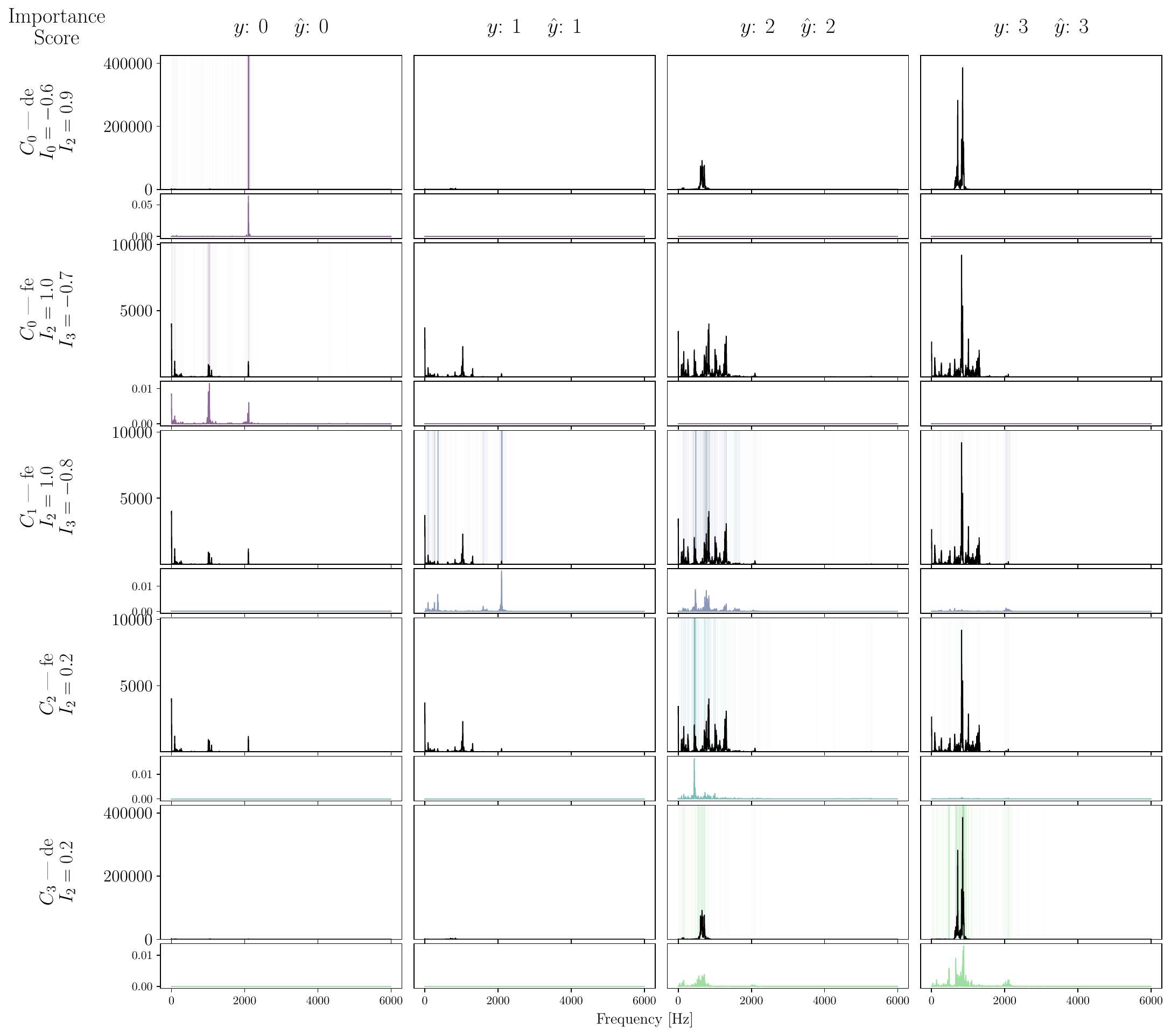}
            \end{subfigure}
    \end{minipage}
    \centering
    \begin{minipage}[c]{0.6\linewidth}
        
        \begin{minipage}[c]{0.14\linewidth}
            \centering\rotatebox{90}{\scriptsize Concept correspondence}
        \end{minipage}%
        \begin{minipage}[c]{0.86\linewidth}
            \begin{subfigure}[t]{\linewidth}
                \includegraphics[width=\linewidth]{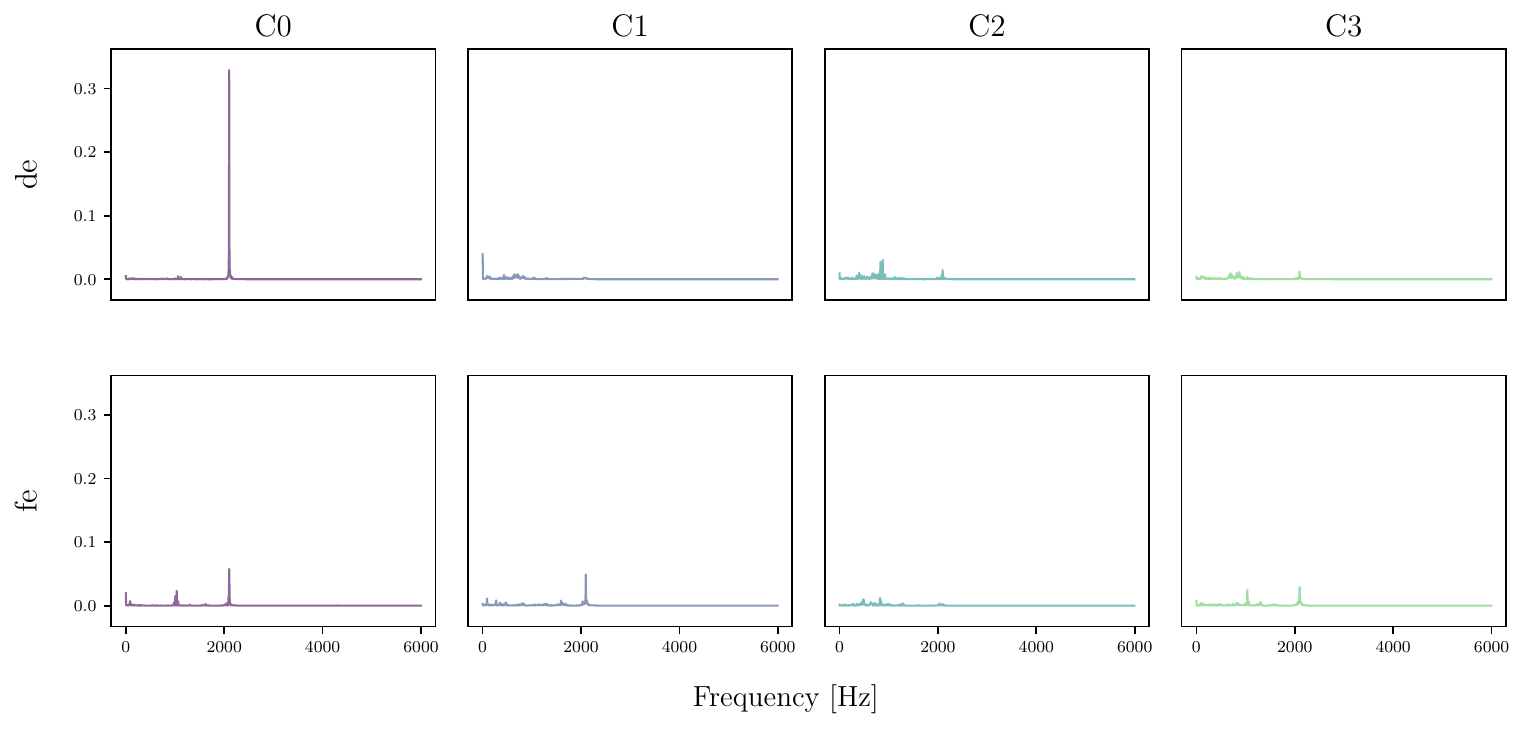}
            \end{subfigure}
        \end{minipage}
    \end{minipage}%
    \caption{Concepts extracted from an InceptionTime10 trained on CWRU. The top grids show time-domain (left) and frequency-domain (right) masks; rows are concepts on channels with high importance and columns are samples. The bottom panel shows the concept-frequency correspondence per channel, with a fan end (fe) and drive end (de) channel. Each concept localizes almost exclusively on one class, with the frequency masks exposing characteristic fault frequencies.}
    \label{fig:cwru}
\end{figure*}

\paragraph{Natural datasets.}
On CWRU (Figure~\ref{fig:cwru}), concepts separate mostly by class. 
$C_0$ aligns with a sharp, dominant peak at $\sim$$2000$~Hz on the drive-end channel, and shows weaker activity at $\sim$$250$ and $\sim$$1000$~Hz on the fan-end channel.
Concepts $C_2$ and $C_3$ correspond more weakly to bands around $800$--$1000$~Hz on the drive-end channel and on the fan-end channel, respectively.
These specific bands give domain experts the information needed to assess whether the model is making predictions for the right frequency reasons.
Analogous results on BearingPD (App.~\ref{sec:app-bearingpd}) and on five datasets of the UCR archive~\cite{UCRArchive2018} (App.~\ref{sec:app-ucr}) support that CENDRe transfers from controlled synthetic settings to real signals across domains. 


%% file: text/60_conclusion.tex
\section{Conclusion}
We have presented CENDRe, a concept extraction method for time-series CNNs that addresses the three limitations of prior time-series CE methods.
First, by propagating  gradients of our contrastive presence score through virtual inspection layers, CENDRe extracts concepts in both time and frequency domains.
The frequency masks recover the ground-truth bands on synthetic data and locate concept-specific bands on the natural bearing-fault datasets, where they support verification by domain experts.
Second, our silhouette-guided aggregation removes the need to fix the number of concepts by hand and delivers results comparable to the best configuration of a manual $K$ sweep in a single run.
This replaces the hyperparameter $K$ with the much less sensitive micro-cluster count $J$.
Third, our gradient-based localization concentrates on the regions that drive the concept, yielding representation correctness comparable to prior methods and higher importance correctness on the synthetic benchmarks.

Being the first CE method to combine time and frequency domain concepts, CENDRe opens new avenues for future research.
The same pipeline can be applied to other invertible representations such as time-frequency distributions. 
Architectures, by contrast, remain a current limitation: CENDRe's per-timestep LADs rely on the approximate translation equivariance of 1D CNNs.
Extending the latent representation to architectures without this property, such as transformers or state-space models, is a natural next step.
Furthermore, the discovered concepts could feed back into the development cycle, for instance to guide model debugging or to flag suspicious reliance on shortcut cues.
Domain-specific user studies, assessing whether CENDRe's masks help experts within their own field, are also a natural direction for future work.

%% file: text/97_appendix_spectral_derivation.tex
\section{Derivation of the spectral concept mask}
\label{sec:app-spectral-derivation}

This appendix derives the frequency-domain CENDRe formulas of Sec.~\ref{sec:methods} as
the sensitivity of a real scalar functional to the log-power spectrum.
The same derivation applies both to the spectral concept mask (setting
$E=\bar\rho_k$) and to the frequency-domain logit sensitivity
(setting $E=y_c(\cdot)$).

\paragraph{Setup.}
Let $x_f=\mathcal{F}(x)\in\mathbb{C}^{F\times D}$ be the rFFT of the
time-domain signal $x$, and let $E:\mathbb{C}^{F\times D}\to\mathbb{R}$ be
any real-valued, differentiable scalar functional.
Writing $x_f=u+iv$ with $u=\Re(x_f)$ and $v=\Im(x_f)$, the power spectrum is
$P=\lvert x_f\rvert^2 = u^2 + v^2 \in\mathbb{R}^{F\times D}_{\ge 0}$.
All operations below are element-wise over the $F\times D$ grid.

\paragraph{Wirtinger preliminaries.}
Treating $x_f$ and $\overline{x_f}$ as formally independent, the Wirtinger derivatives are
\[
\partial_{x_f}E \;\coloneqq\; \tfrac{1}{2}\bigl(\partial_u - i\,\partial_v\bigr)E,
\qquad
\partial_{\overline{x_f}}E \;\coloneqq\; \tfrac{1}{2}\bigl(\partial_u + i\,\partial_v\bigr)E.
\]
For real-valued $E$ we have $\partial_{x_f}E = \overline{\partial_{\overline{x_f}}E}$.
Following the convention adopted by standard automatic-differentiation libraries
(e.g.\ PyTorch), we set
$\nabla_{x_f}E \coloneqq \partial_{\overline{x_f}}E$,
so that the direction of real steepest ascent in the $(u,v)$ parametrization is
$2\nabla_{x_f}E$.

\paragraph{Phase-fixed power perturbation.}
Writing $x_f=re^{i\theta}$ with $r=\sqrt{P}$ and holding the phase $\theta$ fixed,
\[
\frac{dx_f}{dP} \;=\; \frac{e^{i\theta}}{2\sqrt{P}} \;=\; \frac{x_f}{2P},
\qquad
\frac{d\overline{x_f}}{dP} \;=\; \frac{\overline{x_f}}{2P}.
\]

\paragraph{Chain rule.}
The complex chain rule gives
\[
\frac{dE}{dP}
\;=\;
\partial_{x_f}E\odot\frac{dx_f}{dP}
+ \partial_{\overline{x_f}}E\odot\frac{d\overline{x_f}}{dP}
\;=\;
\frac{1}{2P}\Bigl(
\overline{\partial_{\overline{x_f}}E}\odot x_f
+ \partial_{\overline{x_f}}E\odot \overline{x_f}
\Bigr),
\]
where the second equality uses $\partial_{x_f}E = \overline{\partial_{\overline{x_f}}E}$.
Since the two terms in the bracket are complex conjugates of each other,
\[
\frac{dE}{dP}
\;=\;
\frac{1}{P}\,\Re\!\bigl(\overline{x_f}\odot \nabla_{x_f}E\bigr).
\]

\paragraph{Log-power form.}
Multiplying through by $P$ yields the sensitivity with respect to $\log P$, which
avoids the $1/P$ singularity at empty bins and gives a dimensionless quantity:
\begin{equation}
\frac{dE}{d\log P}
\;=\;
\Re\!\bigl(\overline{x_f}\odot \nabla_{x_f}E\bigr).
\label{eq:app-logpower}
\end{equation}

\paragraph{CENDRe spectral mask and logit sensitivity.}
Substituting $E=\bar\rho_k(x_f)$ into \eqref{eq:app-logpower} and
keeping only the positive contributions recovers the spectral concept mask
of Sec.~\ref{sec:methods},
\[
m_k^{\mathrm{freq}}(x)
\;=\;
\mathrm{ReLU}\!\bigl(\Re(\overline{x_f}\odot \nabla_{x_f}\bar\rho_k(x_f))\bigr).
\]
Similarly, setting $E=y_c(x)$ yields the frequency-domain logit sensitivity
$s^{\mathrm{freq}}_{c,\mathrm{ch}}(x) = \Re(\overline{x_{f,\mathrm{ch}}}\odot\nabla_{x_{f,\mathrm{ch}}}y_c(x))$
used in the relevance score.

%% file: text/90_appendix_datasets.tex
\section{Synthetic Dataset Details}
\label{app:datasets}

We construct four synthetic binary-classification time-series datasets with known ground-truth concept masks (primitives), summarized in Table~\ref{tab:datasets}.
\textsc{syntheticLocal} and \textsc{syntheticFrequency} are reported in the main paper, while \textsc{syntheticLmc} and \textsc{syntheticLconf} are included only here, as additional tests for per-channel attribution and confounder robustness.

\begin{table}[!ht]
\caption{The four synthetic datasets. Each is instantiated as several variants pairing primitives or sweeping a confounder ratio.}
\label{tab:datasets}
\footnotesize
\setlength{\tabcolsep}{6pt}
\centering
\begin{tabular}{@{}lcll@{}}
\toprule
\textbf{Name} & $\mathbf{D}$ & \textbf{Variants} & \textbf{Role} \\
\midrule
\textsc{syntheticLocal}     & 1 & 9 & Localized shapes \\
\textsc{syntheticFrequency} & 1 & 6 & Frequency bands \\
\textsc{syntheticLmc}       & 2 & 6 & Per-channel attribution \\
\textsc{syntheticLconf}     & 1 & 6 & Confounder robustness \\
\bottomrule
\end{tabular}
\end{table}

Each variant contains $1{,}000$ samples of length $T=400$.
A base signal is generated by superimposing low-frequency sinusoidal components with additive Gaussian noise, then class-discriminative patterns are injected at random, non-overlapping positions.
Binary ground-truth masks $g_p^{\mathrm{dom}}\in\{0,1\}^{|\mathrm{dom}|\times D}$ (with $|\mathrm{time}|=T$ for time-domain patterns and $|\mathrm{freq}|=F=\lfloor T/2\rfloor+1$ for frequency-domain patterns) record the exact locations of each injected pattern per channel, plus the complementary ``absence'' masks described below.
Shared generation parameters are summarized in Table~\ref{tab:app_gen_params}.

\paragraph{Negatives as primitives.}
A model can learn ``absence of square'' or ``absence of the frequency band'' as a concept just as readily as their presence.
We therefore treat the absence of every involved primitive as a primitive of its own, with a ground-truth mask covering the regions where the primitive is absent.
This ensures that the correctness scores reward concepts that capture absence as well as presence.

\begin{table}[h]
\centering
\caption{Shared generation parameters for all synthetic datasets.}
\label{tab:app_gen_params}
\small
\begin{tabular}{@{}ll@{}}
\toprule
\textbf{Parameter} & \textbf{Value} \\
\midrule
Sample length $T$ & 400 \\
Samples per dataset & 1\,000 \\
Class balance & 50/50 \\
Overlap allowed & No \\
\midrule
\multicolumn{2}{@{}l}{\emph{Shape datasets (\textsc{syntheticLocal} / \textsc{syntheticLmc} / \textsc{syntheticLconf})}} \\
Base signal & Noisy sinusoidal ($f_s{=}1000$\,Hz) \\
Base noise level & 0.05 \\
Base components & 2 sinusoids, $[0.5,\,2]$\,Hz \\
Shape length & 40 time steps \\
Shape amplitude & 1.0 \\
\midrule
\multicolumn{2}{@{}l}{\emph{Frequency datasets (\textsc{syntheticFrequency})}} \\
Base signal & Frequency-domain generator ($f_s{=}400$\,Hz) \\
Base noise level & 0.02 \\
Base components & 8 sinusoids, $[1,\,5]$\,Hz \\
Peaks per band & 5 \\
Peak amplitude & $\mathcal{U}[1.5,\,2.0]$ \\
Random phase & Yes \\
\bottomrule
\end{tabular}
\end{table}

\subsection{\textsc{syntheticLocal} --- Localized Shape Patterns}
\label{app:synthl}

\textsc{syntheticLocal} uses three primitive shapes of length~40:
\begin{itemize}
\item \textbf{Square}: a constant-amplitude block.
\item \textbf{Triangle}: a triangular envelope that peaks at the center and decays linearly to zero at both ends.
\item \textbf{Circle}: a semi-elliptical (half-circle) envelope, producing a smooth bump.
\end{itemize}
Each shape is injected at a uniformly random position, avoiding overlap with previously placed patterns.
Nine variants ($D{=}1$) span three regimes:
\begin{itemize}
\item \emph{Paired} (3 variants). Each class is characterized by a different shape. \textbf{Purpose:} discriminate two class-specific shapes in a clean setting. Variants: \texttt{square-circle}, \texttt{square-triangle}, \texttt{triangle-circle}.
\item \emph{Single-shape vs nothing} (3 variants). One class contains a single shape, the other contains only the noisy base signal. \textbf{Purpose:} probe presence detection of a single primitive against a background. Variants: \texttt{square-nothing}, \texttt{triangle-nothing}, \texttt{circle-nothing}.
\item \emph{Disjunctive vs nothing} (3 variants). One class contains either of two shapes (chosen at random per sample), the other contains only the noisy base signal. \textbf{Purpose:} test whether the method recovers a concept defined as a disjunction of primitives. Variants: \texttt{squareorcircle-nothing}, \texttt{squareortriangle-nothing}, \texttt{triangleorcircle-nothing}.
\end{itemize}

\begin{figure}[h]
\centering
\includegraphics[width=\linewidth]{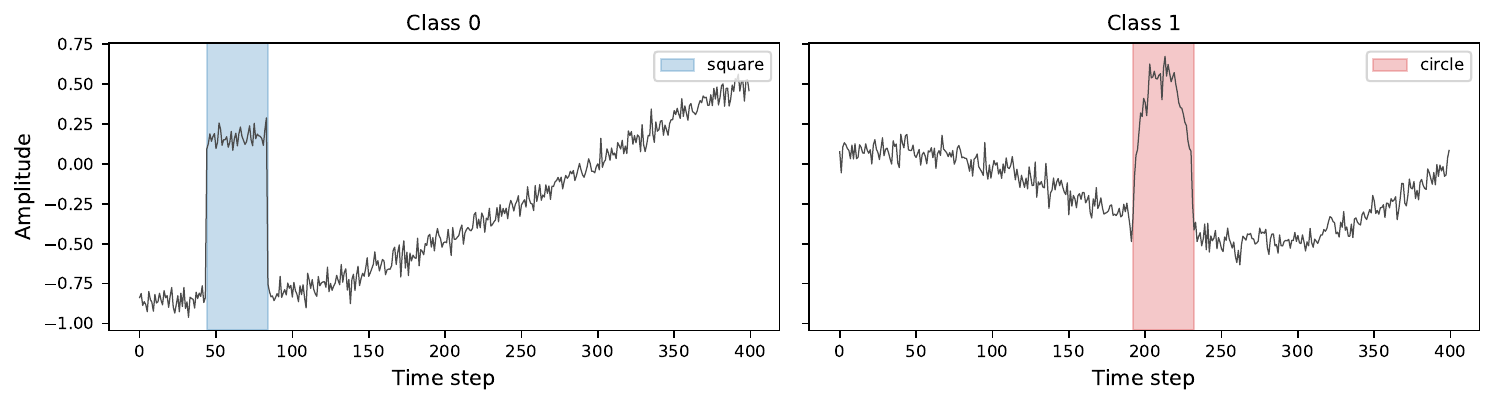}
\caption{Example samples from the \textsc{syntheticLocal} \texttt{square-circle} variant.  Class~0 contains a square pattern (blue region) and class~1 a circle pattern (red region), each injected at a random position into a noisy sinusoidal base signal.}
\label{fig:synthl_example}
\end{figure}

Figure~\ref{fig:synthl_example} shows an example from the \texttt{square-circle} variant.

\subsection{\textsc{syntheticFrequency} --- Frequency Band Patterns}
\label{app:synthf}

\begin{figure}[h]
\centering
\includegraphics[width=\linewidth]{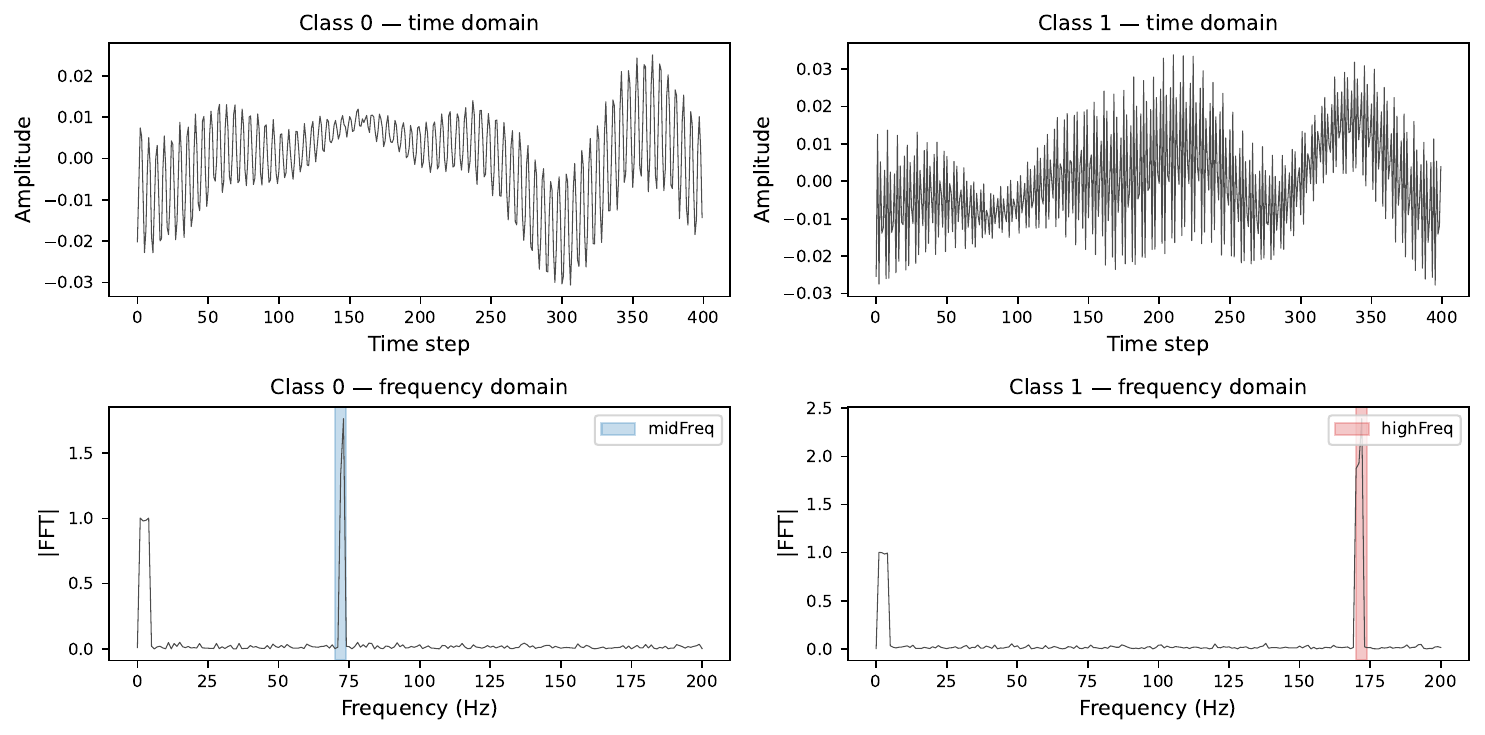}
\caption{Example samples from the \textsc{syntheticFrequency} \texttt{midFreq-highFreq} variant.  Top: time-domain signals.  Bottom: FFT magnitudes with ground-truth frequency bands highlighted (midFreq in blue, highFreq in red).  Class~0 contains the midFreq pattern, class~1 the highFreq pattern.}
\label{fig:synthf_example}
\end{figure}

\textsc{syntheticFrequency} uses frequency-domain patterns drawn from three bands:
\begin{itemize}
\item \textbf{midFreq}: $[70,\,75]$\,Hz
\item \textbf{midhighFreq}: $[125,\,130]$\,Hz
\item \textbf{highFreq}: $[170,\,175]$\,Hz
\end{itemize}
Each band is realized by injecting $5$ sinusoidal peaks with amplitudes drawn from $\mathcal{U}[1.5,\,2.0]$ and random phases into the discrete Fourier transform of the base signal, then inverting back to the time domain.
Ground-truth masks are defined in the frequency domain, marking the bins within each band.
Six variants ($D{=}1$) span two regimes:
\begin{itemize}
\item \emph{Pure paired} (3 variants). Each class is characterized by a different band, with no other peaks. \textbf{Purpose:} clean spectral discrimination, the frequency analogue of the paired regime in \textsc{syntheticLocal}. Variants: \texttt{midFreq-highFreq}, \texttt{midFreq-midhighFreq}, \texttt{midhighFreq-highFreq}.
\item \emph{Paired + confounder} (3 variants). Each pure-paired base is augmented with the third (unused) band injected at a $50/50$ ratio, equally likely in both classes. \textbf{Purpose:} test whether the method ignores a spectrally prominent but uninformative band and focuses on the causal pair. Variants: \texttt{midFreq-highFreq\_conf-midhighFreq-50-50}, \texttt{midFreq-midhighFreq\_conf-highFreq-50-50}, \texttt{midhighFreq-highFreq\_conf-midFreq-50-50}.
\end{itemize}
Figure~\ref{fig:synthf_example} shows an example from the \texttt{midFreq-highFreq} variant.

We deliberately keep \textsc{syntheticFrequency} single-channel.
With discriminative frequencies present in some channels and absent in others, the cross-channel structure is dominated by the low-frequency base signal, and a CNN trained on such a setup is too likely to learn local time-domain shapes induced by these residuals rather than the discriminative bands themselves.
The synthetic benchmarks are only informative when model behavior is strongly predictable from the dataset construction, a property that is already strained on simpler datasets, which the multi-channel frequency setting would defeat.

\subsection{\textsc{syntheticLmc} --- Multi-Channel concept localization}
\label{app:synthlmc}

\textsc{syntheticLmc} reuses the three shape primitives of \textsc{syntheticLocal} but signals are two-channel ($D{=}2$).
Six variants in the following format exist:
\begin{itemize}
\item \emph{Random-channel single-shape} (3 variants). One class contains a single shape injected into a uniformly chosen channel, the other contains only the noisy base signal in both channels. \textbf{Purpose:} test whether the method recovers a channel-agnostic concept, defined by shape rather than channel index. Variants: \texttt{square-nothing\_ch2\_random}, \texttt{triangle-nothing\_ch2\_random}, \texttt{circle-nothing\_ch2\_random}.
\item \emph{Specific-channel paired} (3 variants). Each of two class-discriminative shapes is assigned to a designated channel, one shape per channel. \textbf{Purpose:} test per-channel attribution, whether the method assigns each concept to the correct channel rather than mixing evidence across them. Variants: \texttt{square-circle\_ch2\_specificChannels}, \texttt{square-triangle\_ch2\_specificChannels}, \texttt{triangle-circle\_ch2\_specificChannels}.
\end{itemize}

\subsection{\textsc{syntheticLconf} --- Confounder Robustness}
\label{app:synthlconf}

\textsc{syntheticLconf} reuses the three shape primitives of \textsc{syntheticLocal} ($D{=}1$) and adds a non-discriminative confounder shape on top of each paired base, with controlled class--confounder correlation.
Six variants combine two paired bases with three confounder ratios:
\begin{itemize}
\item \emph{Bases.} \texttt{square-circle} with \texttt{triangle} as confounder, and \texttt{triangle-circle} with \texttt{square} as confounder.
\item \emph{Ratios.} $40/60$, $50/50$, and $60/40$, where the ratio gives the proportion of confounder presence in class~0 versus class~1.
\end{itemize}
\textbf{Purpose:} test robustness to spurious correlations. At $50/50$ the confounder carries no label information, while at $40/60$ and $60/40$ it does, so a method that latches onto the prominent shape rather than the causal pair is penalized more strongly at the off-balance ratios.

%% file: text/92_appendix_models.tex
\section{Implementation details}
\label{app:models}

This appendix collects the implementation details needed to reproduce CENDRe and the baselines: backbone architectures, the probe layers used for concept extraction, and the CENDRe clustering hyperparameters.

\subsection{Model architectures}
\label{app:models-arch}
Table~\ref{tab:app_models} lists the exact hyperparameters of the three 1D CNN architectures used throughout the experiments (\S\ref{sec:models}).
All three are direct 1D adaptations of the corresponding standard image-classification architectures, with 2D convolutions replaced by 1D convolutions and 2D pooling by 1D pooling.

\begin{table}[h]
\centering
\caption{Architecture hyperparameters used for the three 1D CNN backbones. Defaults follow each architecture's reference implementation.}
\label{tab:app_models}
\small
\setlength{\tabcolsep}{6pt}
\begin{tabular}{@{}ll@{}}
\toprule
\textbf{Model} & \textbf{Configuration} \\
\midrule
\textbf{InceptionTime10}~\cite{ismail2020inceptiontime}
  & 10 inception modules, 32 filters per branch, base kernel size $40$, \\
  & with bottleneck layer and residual connections every three modules. \\
\midrule
\textbf{ResNet1D-18}~\cite{he2016deep}
  & Four stages with $[2,2,2,2]$ residual blocks, expansion factor $1$. \\
\midrule
\textbf{DenseNet1D-121}~\cite{huang2017densely}
  & Block configuration $[6,12,24,16]$, growth rate $32$, compression factor $0.5$, \\
  & $64$ initial features. \\
\bottomrule
\end{tabular}
\end{table}

\subsection{Probe layers }

The choice of which layers to analyze constitutes the most critical hyperparameter for CE. Table~\ref{tab:layers} reports the specific layers selected for each network. In ECLAD-ts, probe layers are sampled at uniform intervals throughout the architecture. MultiVISION instead relies on a single bottleneck layer, which prevents the analysis of blocks that residual connections might bypass. CENDRe adopts the same probe layers as ECLAD-ts. For MultiVISION, the neuron activation extraction threshold is set to the $0.99$ quantile of the layer under examination, and activations are grouped via $k$-means clustering.

\begin{table}[h]
    \caption{Layers used for CE with each method and model.}
    \label{tab:layers}
    \scriptsize
    \setlength{\tabcolsep}{3pt}
    \centering
    \begin{minipage}[t]{0.567\textwidth}
    \centering
    \begin{tabular}{@{}l|p{6.05cm}@{}}
    \toprule
         Model & ECLAD-ts and CENDRe \\
    \midrule
         InceptionTime10 & \texttt{model.inception\_block.inception\_layers.$n_b$}\allowbreak\texttt{.bottleneck}, $n_b\in\{6,7,8,9\}$ \\
         ResNet1D-18     & \texttt{model.layers.$n_b$.1.relu} $n_b \in \{0,1,2,3\}$ \\
         DenseNet1D-121  & \texttt{model.features.transition$n_b$.conv}, $n_b \in \{1,2,3\}$;\newline \texttt{model.features.denseblock4.}\allowbreak\texttt{block.15.conv2} \\
    \bottomrule
    \end{tabular}
    \end{minipage}
    \hfill
    \begin{minipage}[t]{0.42\textwidth}
    \centering
    \begin{tabular}{@{}l|p{4.1cm}@{}}
    \toprule
         Model & MultiVISION \\
    \midrule
         InceptionTime10 & \texttt{model.inception\_block.}\allowbreak\texttt{inception\_layers.9.bottleneck} \\
         ResNet1D-18     & \texttt{model.layers.2.1.relu} \\
         DenseNet1D-121  & \texttt{model.features.transition3.conv} \\
    \bottomrule
    \end{tabular}
    \end{minipage}
\end{table}

\subsection{CENDRe clustering hyperparameters}
\label{app:cendre-clustering}
The second stage of CENDRe$_\text{silhouette}$ and CENDRe$_\text{HDBSCAN}$ runs on the $J=50$ micro-centroids returned by mini-batch $k$-means.
CENDRe$_\text{silhouette}$ uses average linkage and selects $K\in\{2,\dots,10\}$ by the silhouette score, both evaluated under cosine distance.
CENDRe$_\text{HDBSCAN}$ uses $\mathrm{min\_cluster\_size}=5$ under cosine distance.
Micro-centroids that HDBSCAN labels as noise are excluded from $\bar{\mathcal{J}}_k$ when computing $\rho_k$.

%% file: text/95_appendix_additional_results.tex
\newcommand{\reportpanels}[3]{%
    \begin{minipage}[c]{0.75\linewidth}
        \begin{minipage}[c]{0.14\linewidth}
            \centering\rotatebox{90}{\scriptsize Time domain}
        \end{minipage}%
        \begin{minipage}[c]{0.86\linewidth}
            \includegraphics[width=\linewidth]{#1}
        \end{minipage}

        \vspace{0.5em}

        \begin{minipage}[c]{0.14\linewidth}
            \centering\rotatebox{90}{\scriptsize Frequency domain}
        \end{minipage}%
        \begin{minipage}[c]{0.86\linewidth}
            \includegraphics[width=\linewidth]{#2}
        \end{minipage}
    \end{minipage}%
    \hfill
    \begin{minipage}[c]{0.24\linewidth}
        \includegraphics[width=\linewidth]{#3}
    \end{minipage}%
}

\section{Additional results}
\label{sec:app-additional}
This appendix extends Section~\ref{sec:results} in two directions.
\S\ref{sec:app-additional-synthetic} broadens the synthetic evaluation: it compares all three CENDRe variants across three architectures, visualizes concepts under two further invertible transforms, breaks the quantitative comparison down per dataset with significance tests, and extends it to two dataset families that test multichannel localization and confounder robustness.
\S\ref{sec:app-additional-natural} extends the qualitative evaluation to five real-world datasets from the UCR archive and to a second bearing-fault dataset, BearingPD, on two architectures.

\subsection{Synthetic datasets} 
\label{sec:app-additional-synthetic}
We extend the synthetic evaluation of Section~\ref{sec:results} along four lines.
We first repeat the qualitative comparison of Figure~\ref{fig:syntheticLocal} across all three CENDRe variants and three architectures.
We then visualize concepts under two further invertible transforms, the short-time Fourier and the wavelet transform (\S\ref{sec:app-domains}).
Next, we break the family-level quantitative comparison of Figure~\ref{fig:quant} down per dataset and test the resulting ordering for statistical significance (\S\ref{sec:app-distributional}, \S\ref{sec:app-significance}).
Finally, we extend the quantitative comparison to two appendix-only dataset families, \textsc{syntheticLmc} and \textsc{syntheticLconf}, testing multichannel localization and confounder robustness.

\subsubsection{Qualitative comparison across CENDRe variants and architectures}
\label{sec:app-variants}
\begin{figure*}[!tb]
    \centering
    \parbox[c]{0.05\linewidth}{}%
    \parbox[c]{0.315\linewidth}{\centering\scriptsize ResNet1D-18}%
    \parbox[c]{0.315\linewidth}{\centering\scriptsize InceptionTime10}%
    \parbox[c]{0.315\linewidth}{\centering\scriptsize DenseNet1D-121}\par
    \vspace{2pt}
    \parbox[c]{0.05\linewidth}{\centering\rotatebox{90}{\scriptsize CENDRe$_\textrm{silhouette}$}}%
    \parbox[c]{0.315\linewidth}{\centering\includegraphics[width=\linewidth]{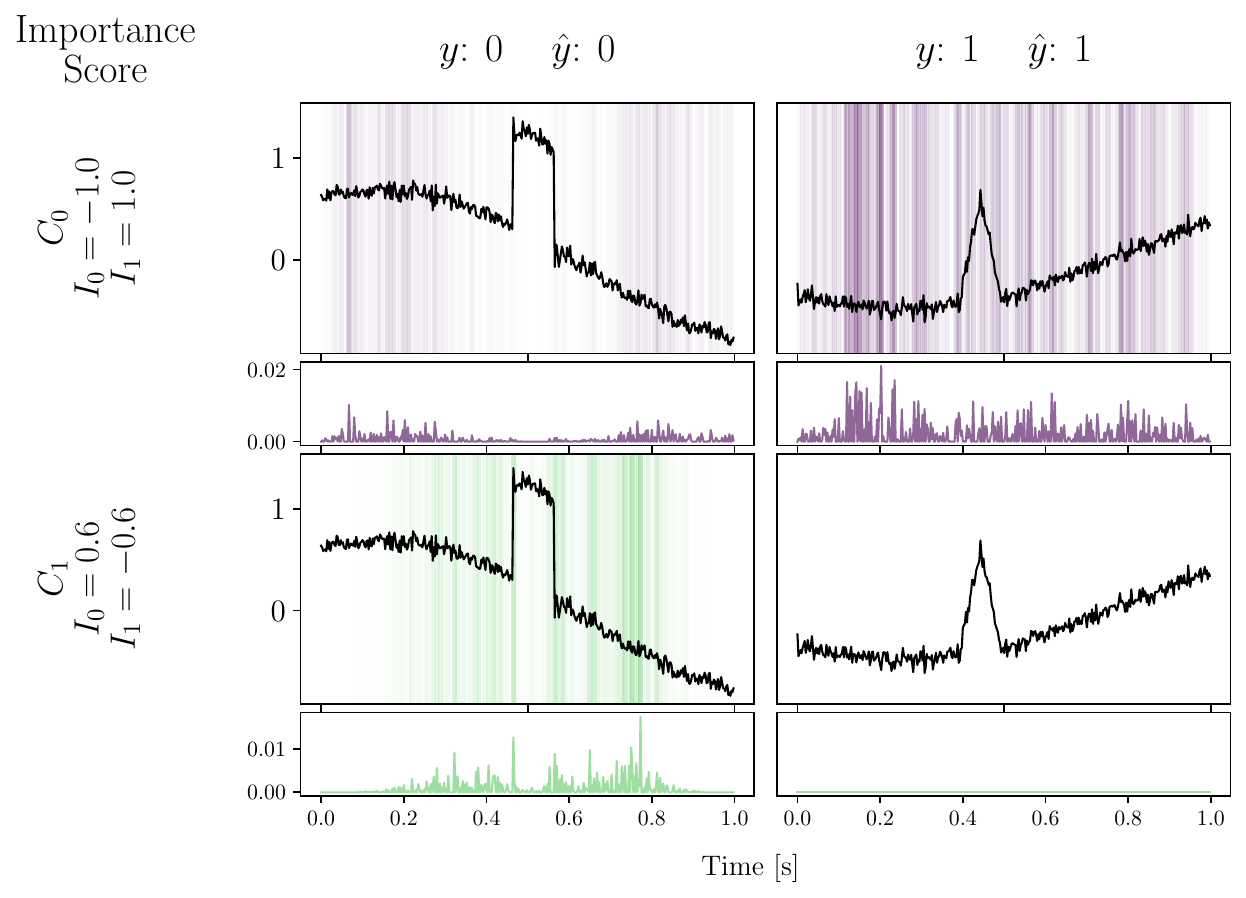}}%
    \parbox[c]{0.315\linewidth}{\centering\includegraphics[width=\linewidth]{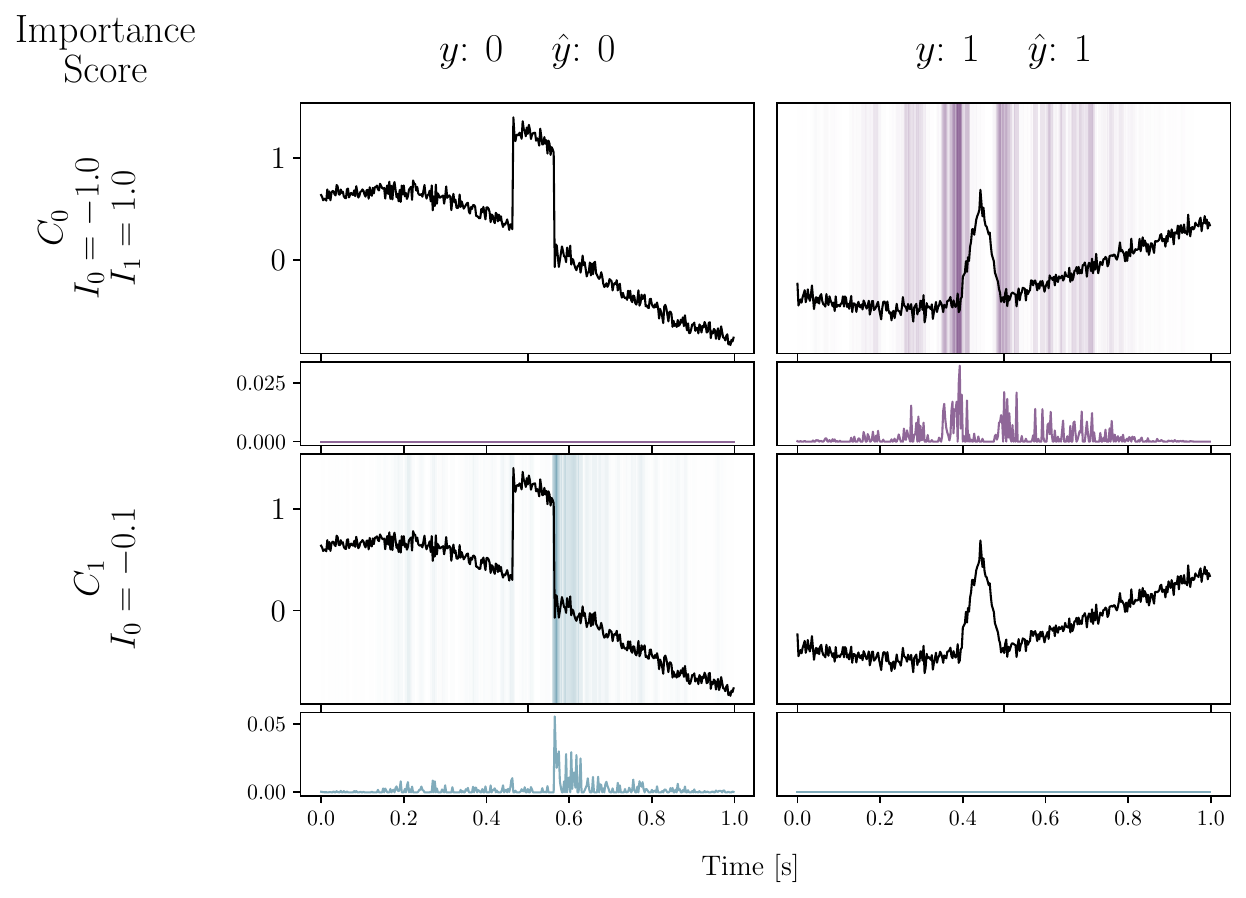}}%
    \parbox[c]{0.315\linewidth}{\centering\includegraphics[width=\linewidth]{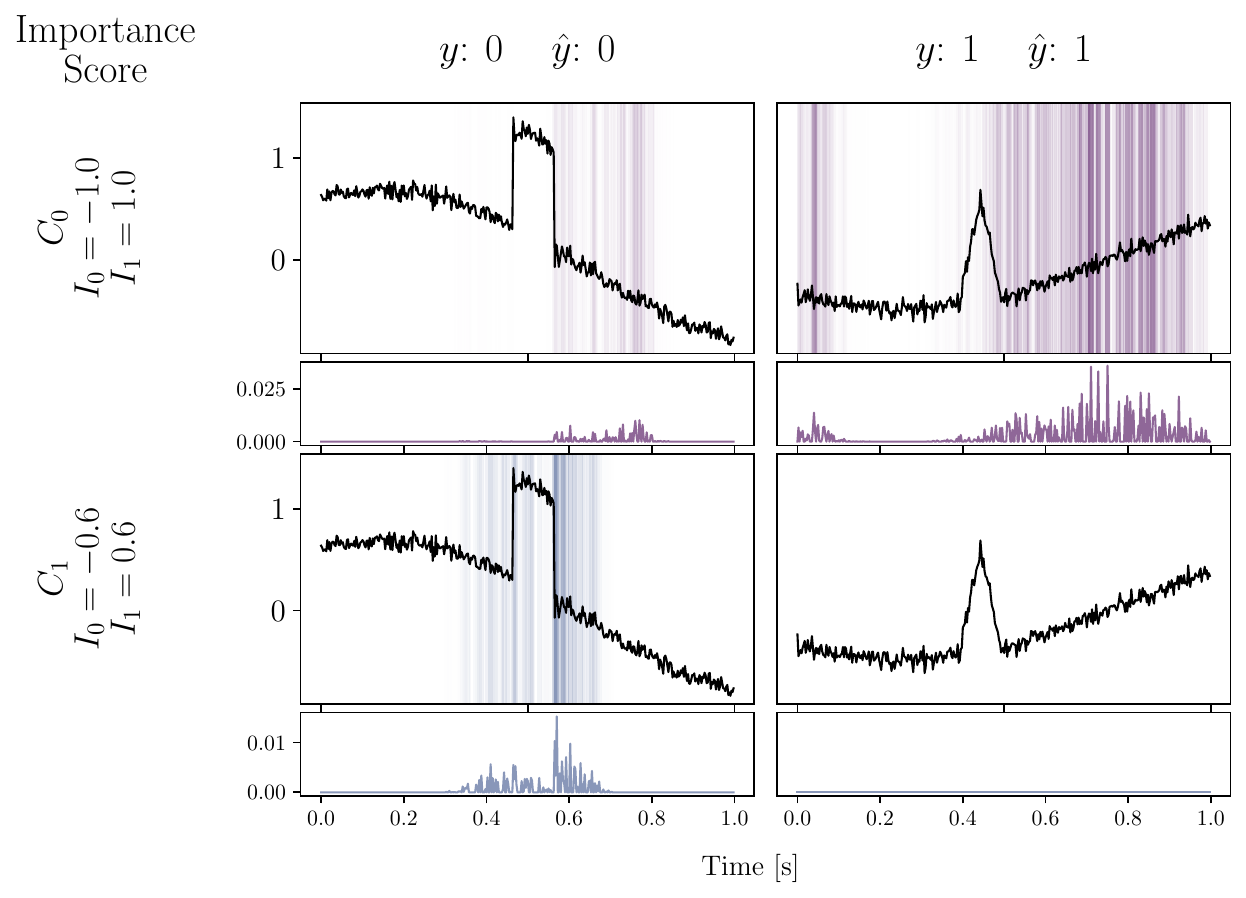}}\par
    \vspace{-20pt}
    \parbox[c]{0.05\linewidth}{\centering\rotatebox{90}{\scriptsize CENDRe$_\textrm{HDBSCAN}$}}%
    \parbox[c]{0.315\linewidth}{\centering\includegraphics[width=\linewidth]{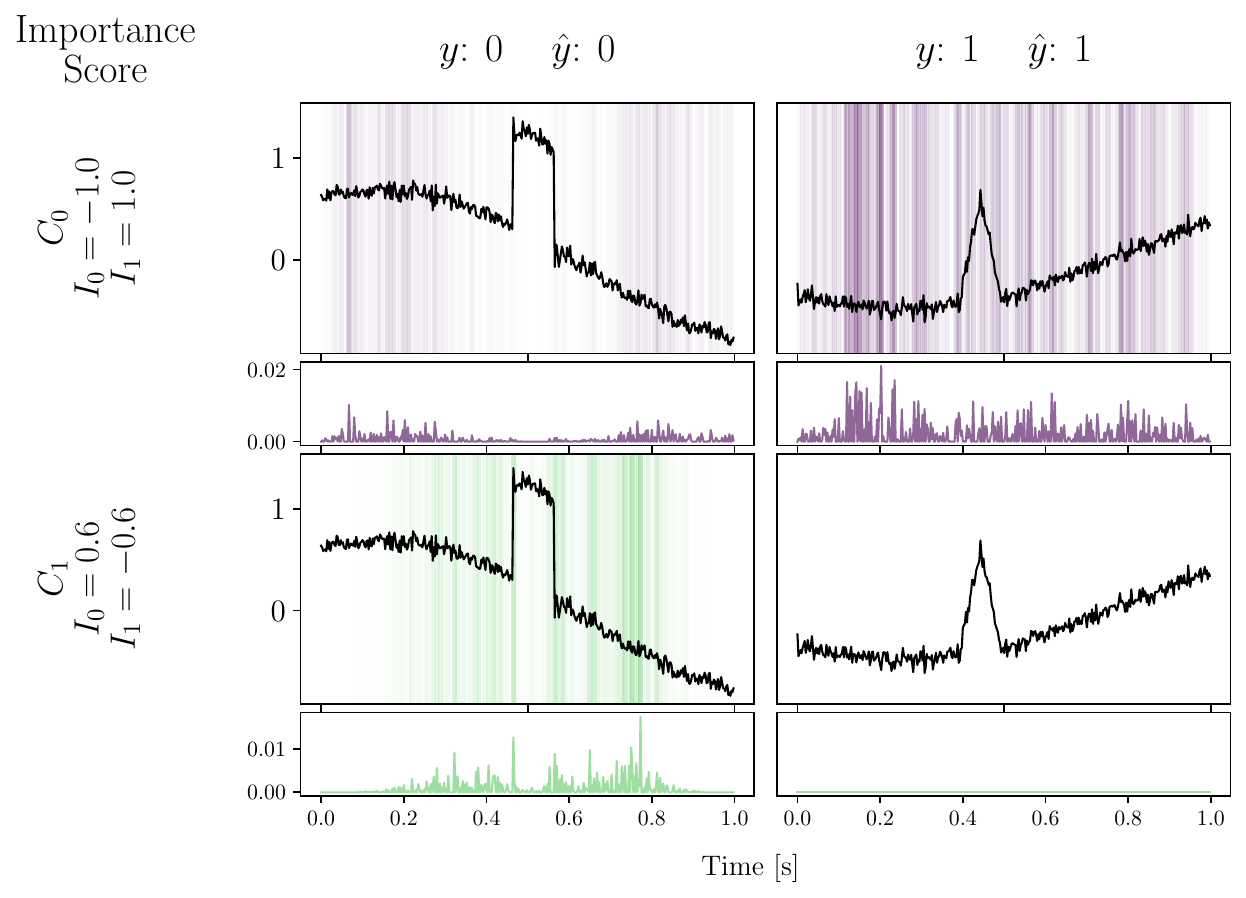}}%
    \parbox[c]{0.315\linewidth}{\centering\includegraphics[width=\linewidth]{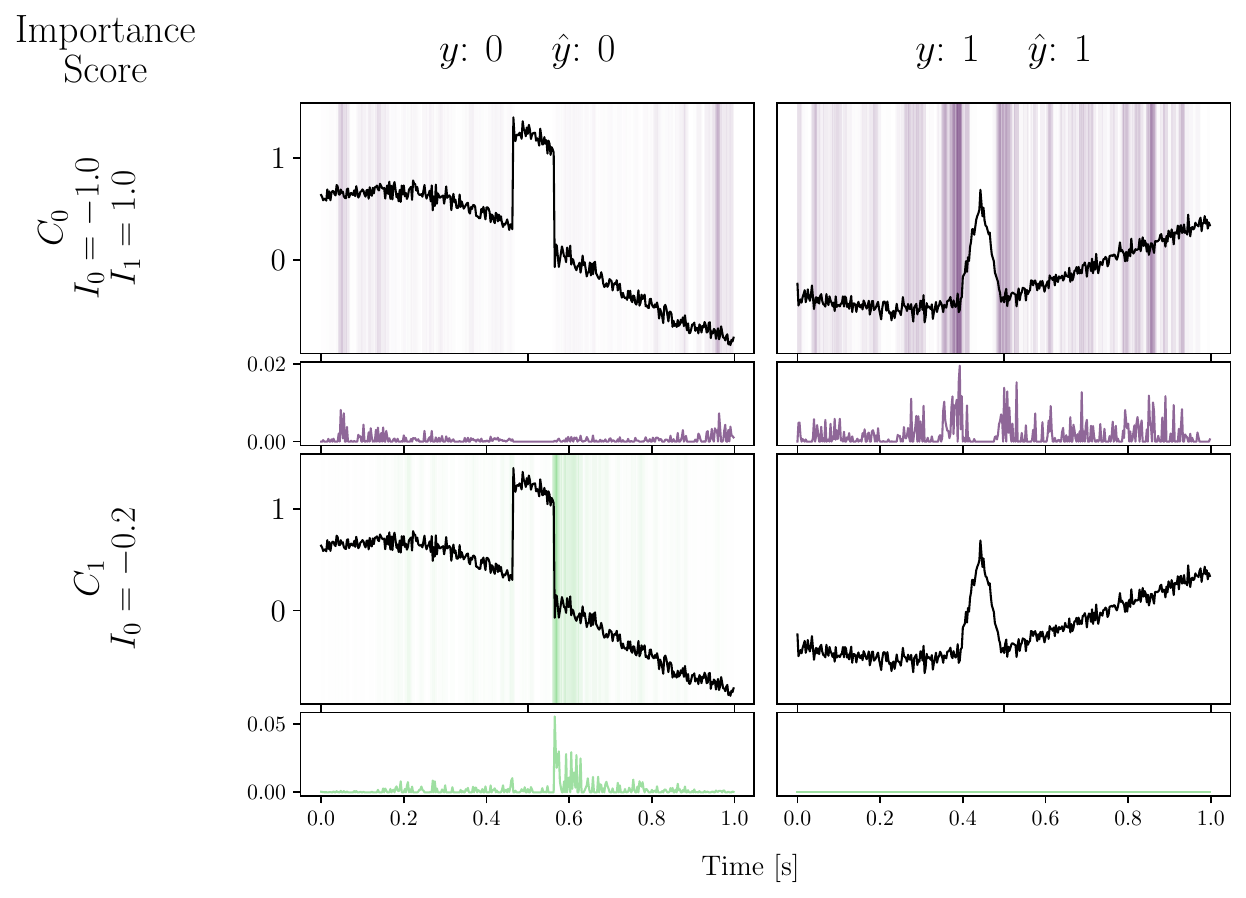}}%
    \parbox[c]{0.315\linewidth}{\centering\includegraphics[width=\linewidth]{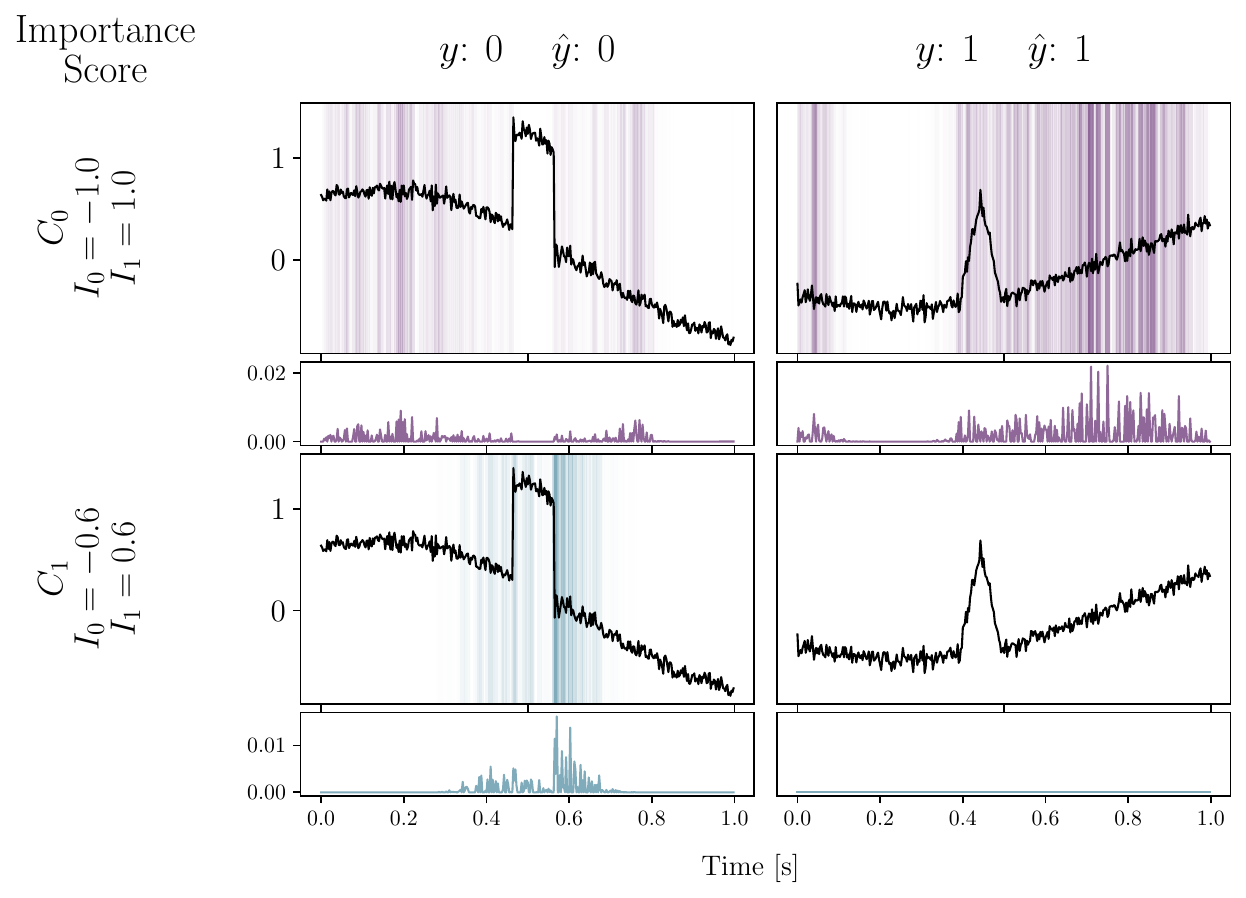}}\par
    \vspace{-20pt}
    \parbox[c]{0.05\linewidth}{\centering\rotatebox{90}{\scriptsize CENDRe$_\textrm{kMeans}$}}%
    \parbox[c]{0.315\linewidth}{\centering\includegraphics[width=\linewidth]{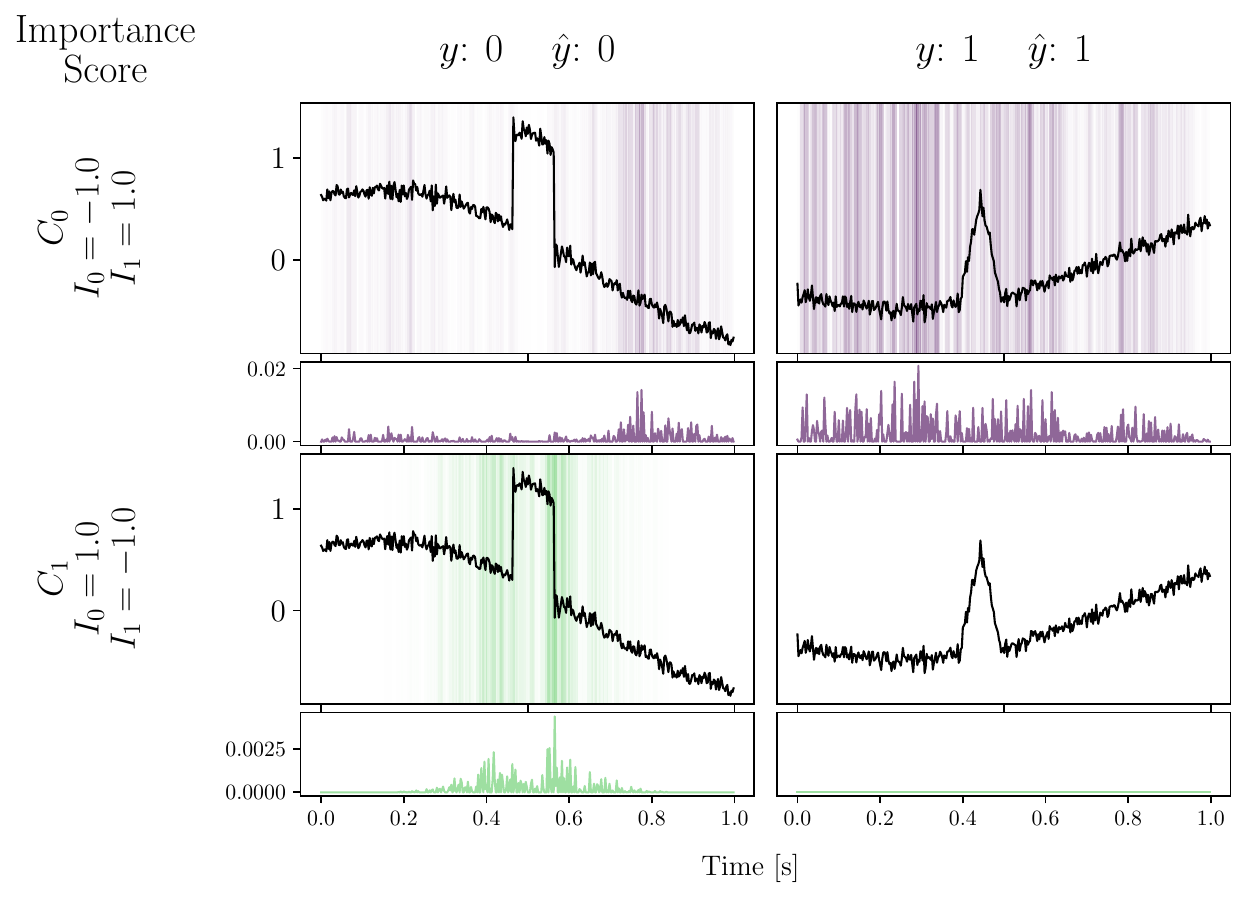}}%
    \parbox[c]{0.315\linewidth}{\centering\includegraphics[width=\linewidth]{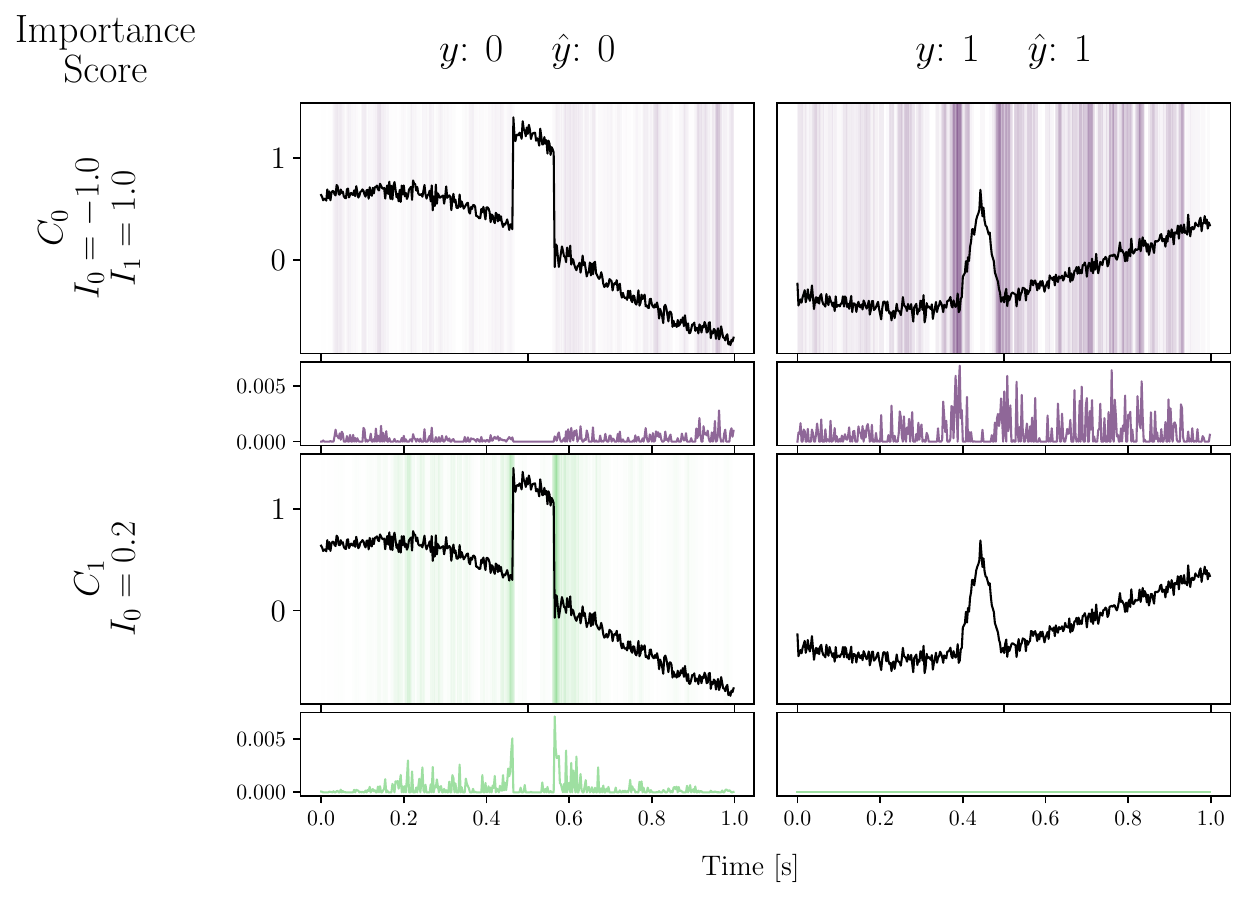}}%
    \parbox[c]{0.315\linewidth}{\centering\includegraphics[width=\linewidth]{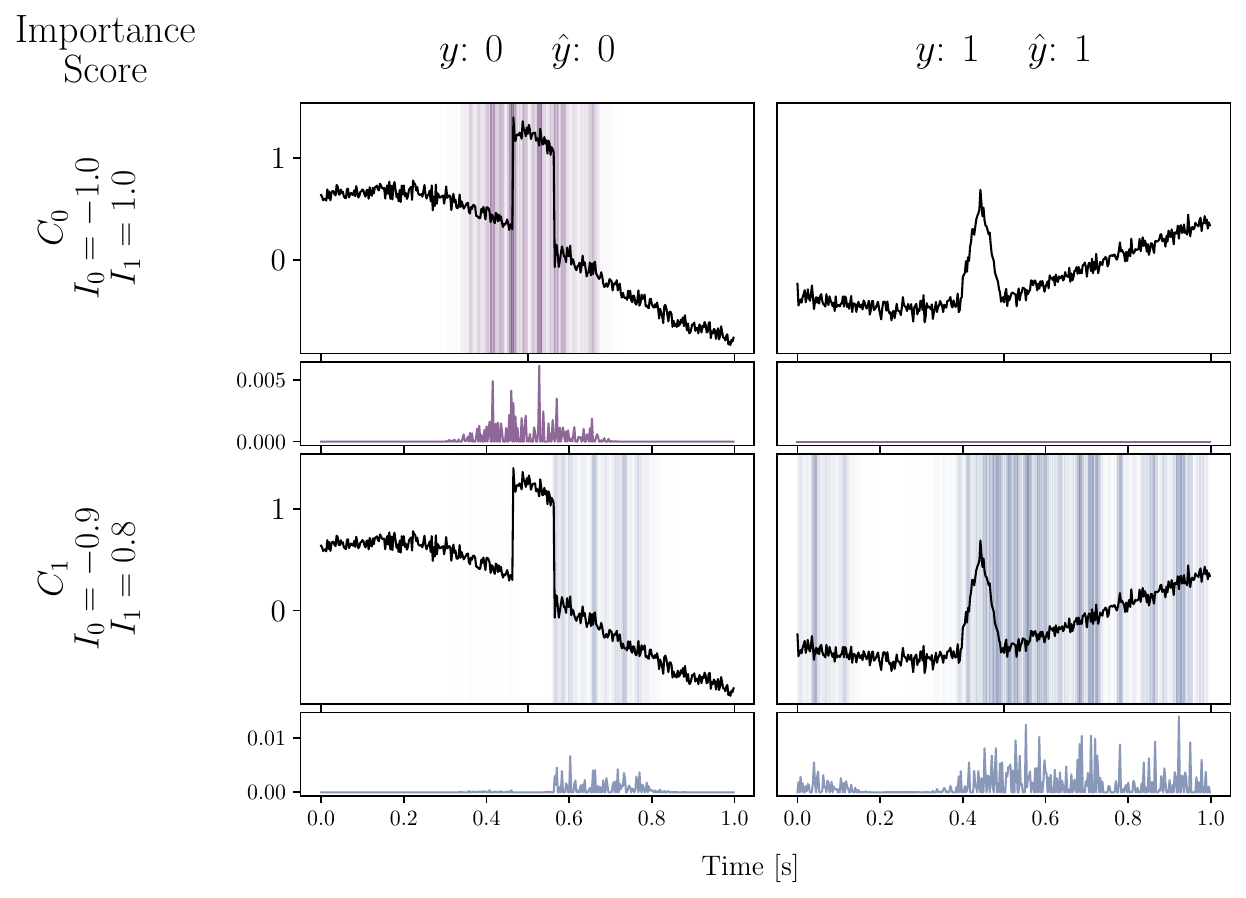}}\par
    \vspace{-20pt}
    \parbox[c]{0.05\linewidth}{\centering\rotatebox{90}{\scriptsize ECLAD-ts}}%
    \parbox[c]{0.315\linewidth}{\centering\includegraphics[width=\linewidth]{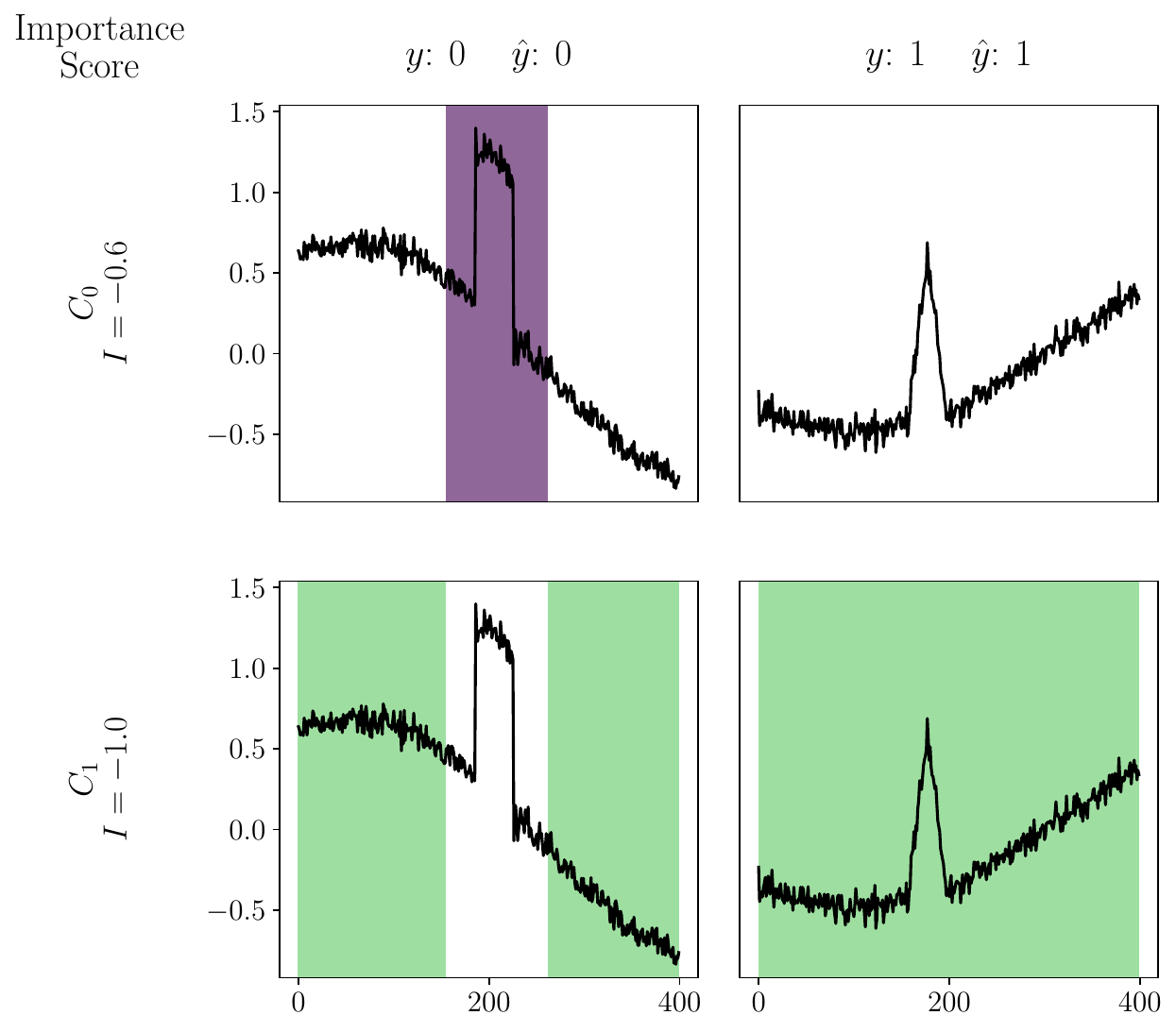}}%
    \parbox[c]{0.315\linewidth}{\centering\includegraphics[width=\linewidth]{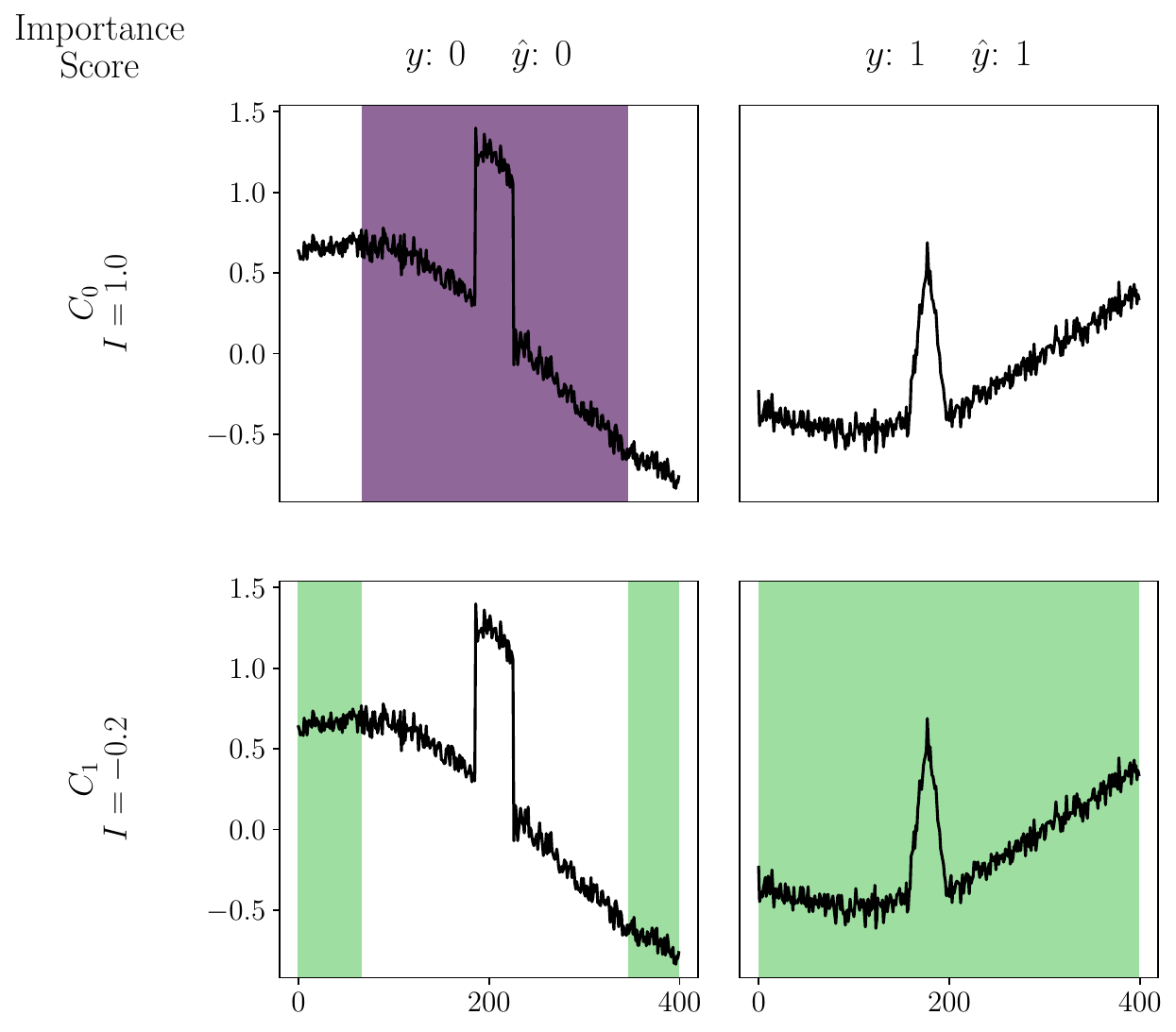}}%
    \parbox[c]{0.315\linewidth}{\centering\includegraphics[width=\linewidth]{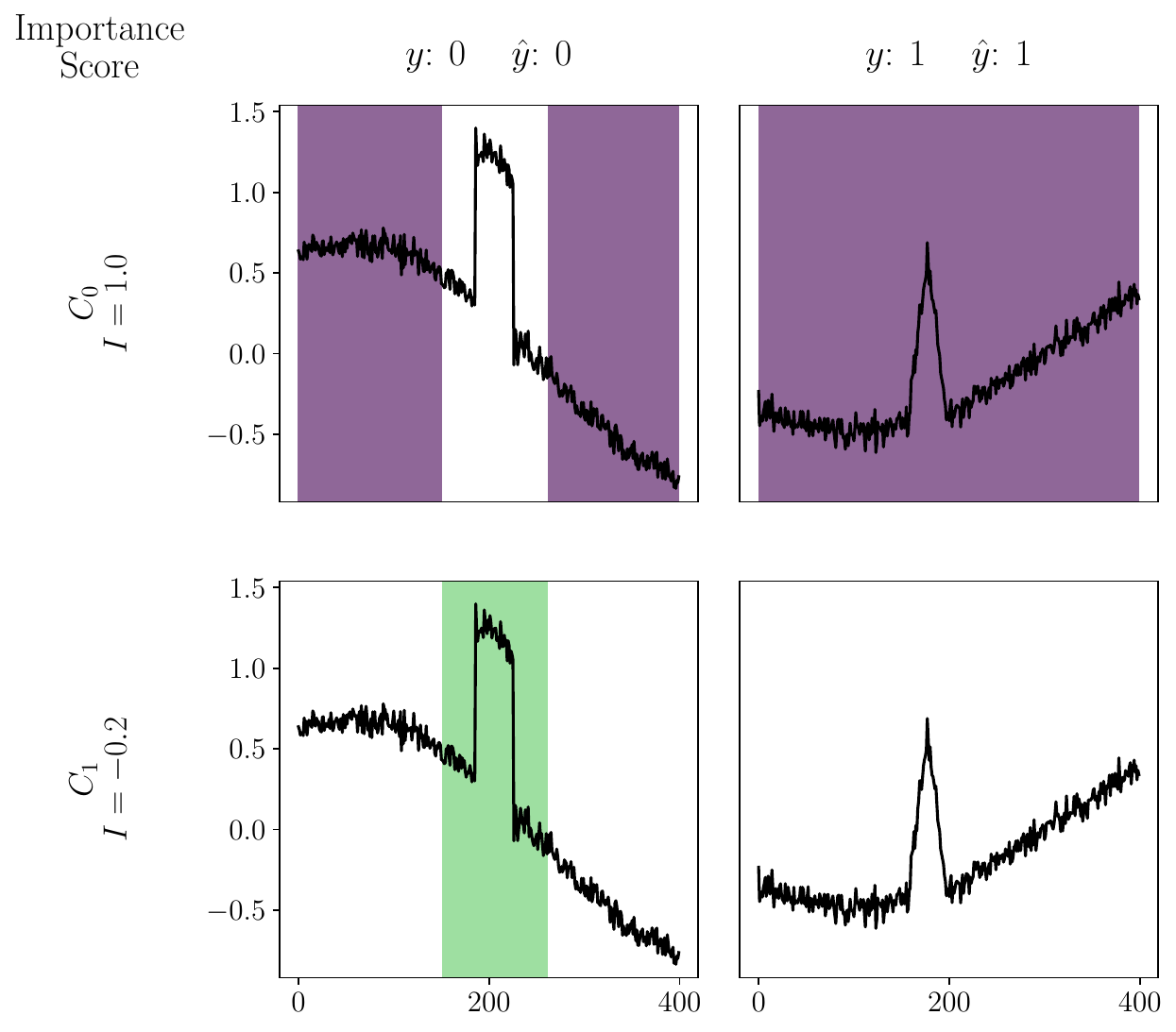}}\par
    \vspace{-5pt}
    \parbox[c]{0.05\linewidth}{\centering\rotatebox{90}{\scriptsize MultiVISION}}%
    \parbox[c]{0.315\linewidth}{\centering\includegraphics[width=\linewidth]{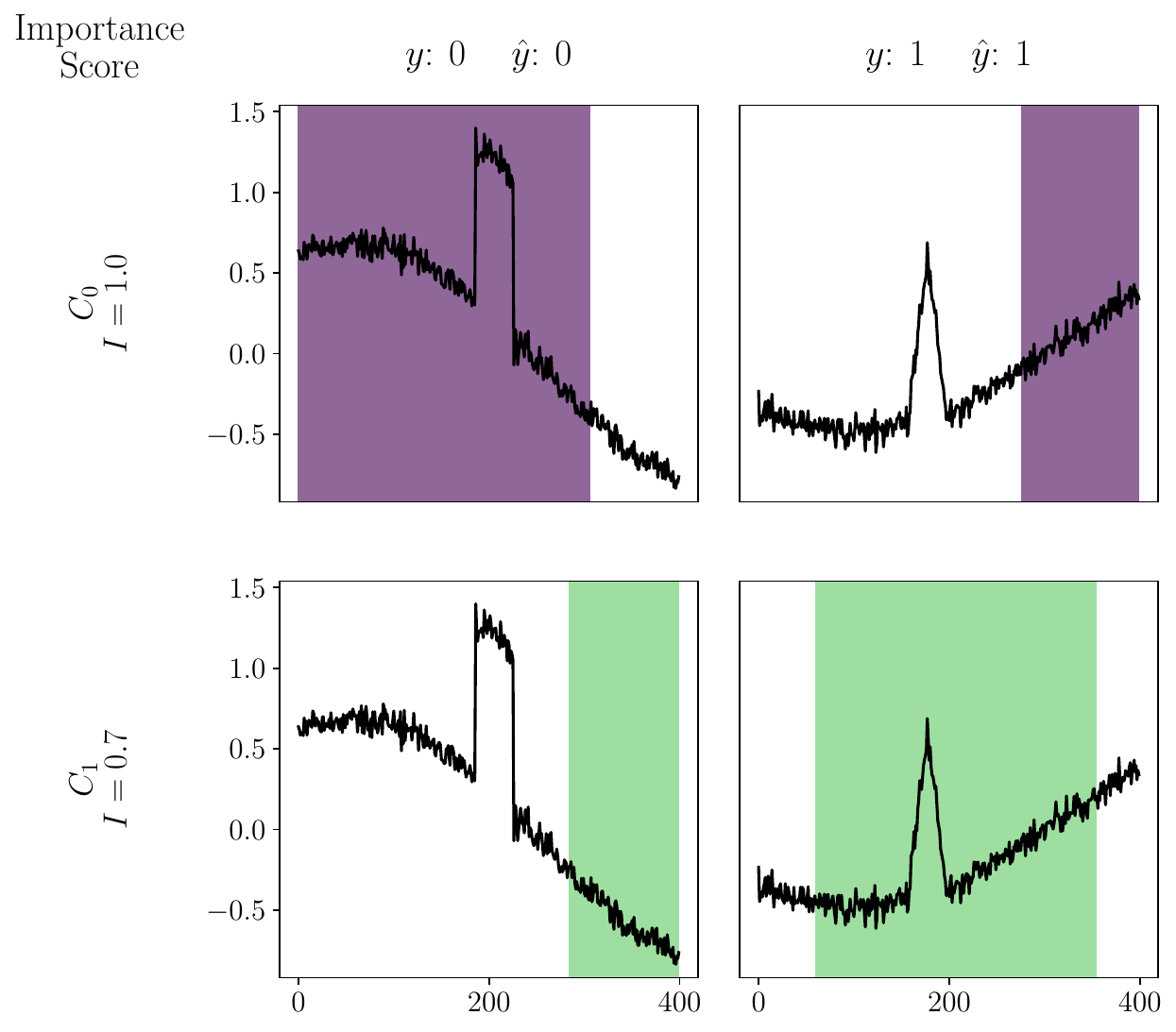}}%
    \parbox[c]{0.315\linewidth}{\centering\includegraphics[width=\linewidth]{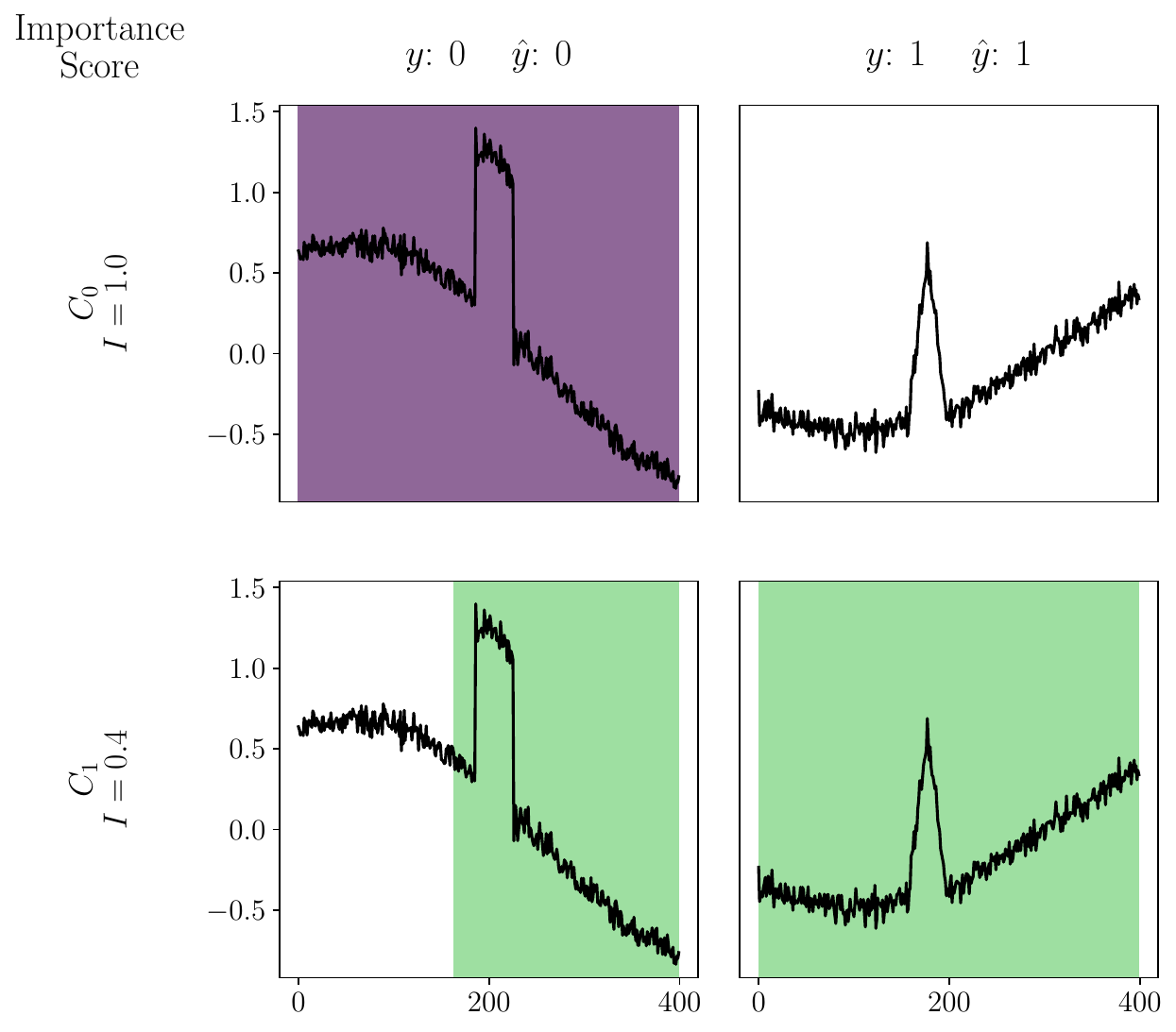}}%
    \parbox[c]{0.315\linewidth}{\centering\includegraphics[width=\linewidth]{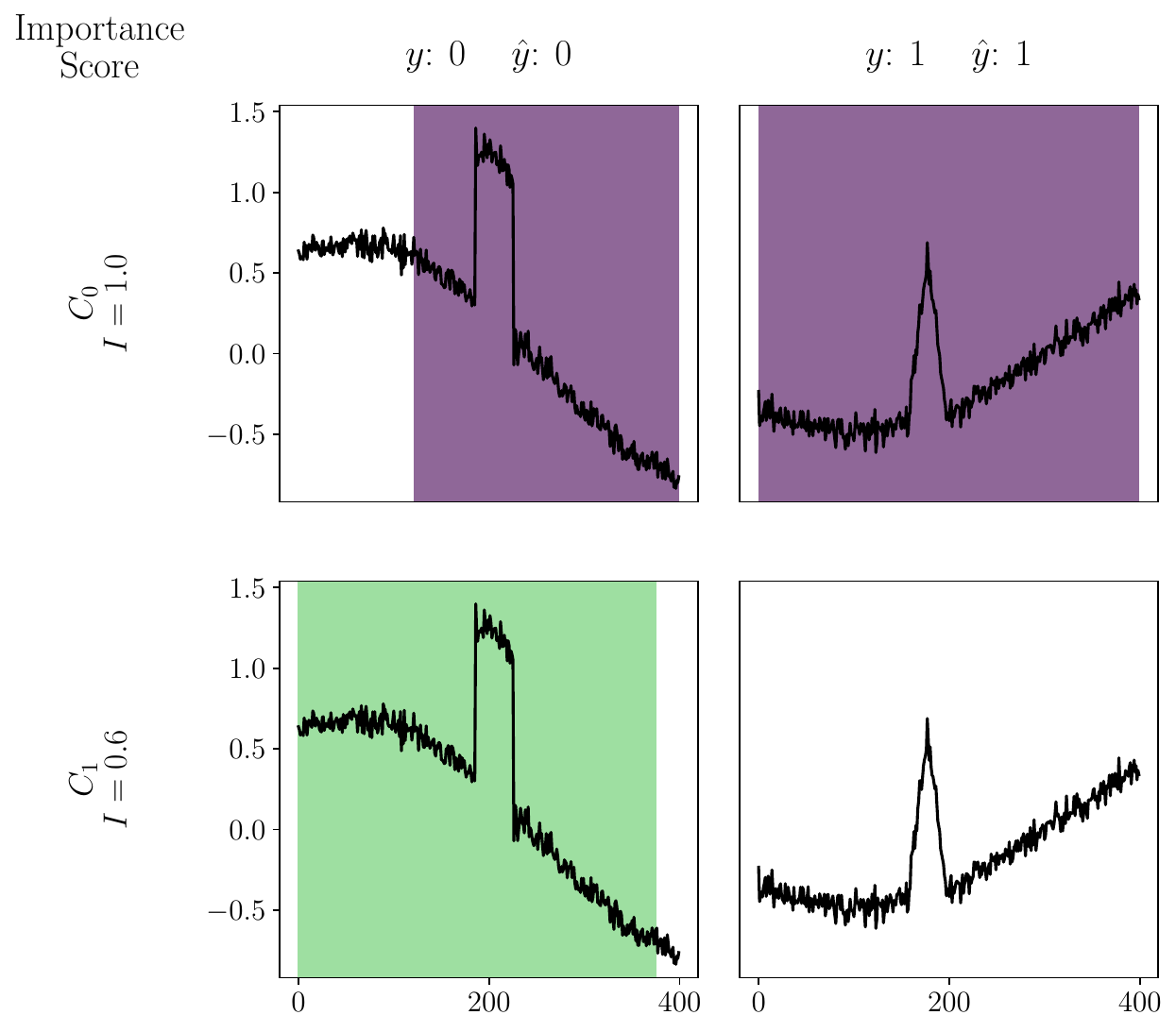}}\par

    \caption{Concepts extracted by CENDRe$_\textrm{silhouette}$, CENDRe$_\textrm{HDBSCAN}$,
    CENDRe$_\textrm{kMeans}$, ECLAD-ts, and MultiVISION (rows) from a ResNet1D-18, an
    InceptionTime10, and a DenseNet1D-121 (columns, in that order) trained on the same
    \textsc{syntheticLocal} square-triangle dataset as Figure~\ref{fig:syntheticLocal}.
    For space, each panel shows the two most important concepts; within a panel, rows are
    concepts, columns are one class-0 and one class-1 sample, and row labels report the
    importance scores. Across architectures, ECLAD-ts isolates the square as a solid block, 
    CENDRe recovers the same regions as ECLAD-ts but as graded masks whose largest values partially sit at the edges, 
    and MultiVISION shows no correspondence to either primitive. 
    On DenseNet1D-121 the CENDRe variants extract additional
    low-importance concepts (not shown) that further partition the background.}
    \label{fig:syntheticLocalVariants}
\end{figure*}

Section~\ref{sec:results} reports the silhouette-guided aggregation variant (CENDRe$_{\text{silhouette}}$) as the default, extracted from a single ResNet1D-18.
Figure~\ref{fig:syntheticLocalVariants} extends that view in two directions, adding the two remaining CENDRe variants and repeating the extraction on two further architectures.
It shows the concepts extracted by the five CE methods (rows) from a ResNet1D-18, an InceptionTime10, and a DenseNet1D-121 (columns) trained on the same \textsc{syntheticLocal} square-triangle dataset as Figure~\ref{fig:syntheticLocal}.
CENDRe$_{\text{silhouette}}$ and CENDRe$_{\text{HDBSCAN}}$ set the number of concepts automatically; for CENDRe$_{k\text{Means}}$, ECLAD-ts, and MultiVISION we report the run whose number of concepts maximizes the RC score.
For space, each panel shows the two most important concepts.

The three CENDRe variants stay close to each other within every column, pointing to similar regions with comparable importance scores.
Reporting CENDRe$_{\text{silhouette}}$ alone in the main text therefore loses little.
Against the baselines, the comparison reproduces the one in Section~\ref{sec:results} on all three architectures.
ECLAD-ts marks the square as a uniform block and its complement as a second concept, while CENDRe spreads graded mass over similar regions, with the largest values at the edges of the shapes.
MultiVISION splits the signal into broad contiguous segments whose boundaries follow neither primitive.

All three architectures yield concepts that map onto the primitives, with slight differences between them.
On the ResNet1D-18 and the InceptionTime10, every method separates the discriminative shapes from the background.
The DenseNet1D-121 follows the same pattern, with the CENDRe variants splitting the background further into additional low-importance concepts, left out of the figure for space.

\subsubsection{Visualizations in other natural domains} 
\label{sec:app-domains}
In Section~\ref{sec:methods}, we instantiate the transform $\phi$ as a Fourier transform to visualize the concepts in the frequency domain, 
but mention how other invertible transforms could be used instead.
While we don't test them quantitatively, such transforms are straightforward to implement and can be used to visualize the same concepts in other domains, 
e.g., wavelet transforms or short-time Fourier transforms for time-frequency analysis.
We have implemented such transforms by creating virtual inspection layers for short-time Fourier transform (STFT) and discrete wavelet transform (DWT).
The STFT builds on the implementation of PyTorch~\cite{pytorch}, and the DWT on the differentiable filter banks of the PyTorch Wavelet Toolbox~\cite{wolter2024ptwt}, which in turn builds on PyWavelets~\cite{lee2019pywavelets}.
The STFT $\mathcal{S}:\mathbb{R}^{T\times D}\to\mathbb{C}^{F\times N\times D}$ applies the rFFT to overlapping, windowed segments of length $W$ with hop size $H$, producing $F\coloneqq\lfloor W/2\rfloor+1$ frequency bins across $N$ time frames.
Because $\mathcal{S}$ is complex-valued like $\mathcal{F}$, the sensitivity in Eq.~\ref{eq:masks} carries over unchanged after substituting $x_s\coloneqq\mathcal{S}(x)$ for $x_f$.
Its derivation only requires a complex transform output, not a specific transform (App.~\ref{sec:app-spectral-derivation}).
We invert $\mathcal{S}$ exactly by choosing an analysis window that satisfies the nonzero-overlap-add condition.
The DWT $\mathcal{W}:\mathbb{R}^{T\times D}\to\mathbb{R}^{T\times D}$ decomposes $x$ into approximation and detail coefficients through an orthogonal wavelet filter bank.
This filter bank admits an exact inverse by construction.
Since $\mathcal{W}$ is real-valued, unlike $\mathcal{F}$ and $\mathcal{S}$, its mask reduces to the plain ReLU gradient, following the same pattern as $m_k^{\mathrm{time}}$ in Eq.~\ref{eq:masks}.
We selected STFT and DWT because both are exactly invertible, matching the requirement virtual inspection layers place on $\phi$.
Other time-frequency transforms, such as the continuous wavelet transform, lack an exact inverse and are left for future work.

\begin{figure}[!htb]
    \centering
    \begin{subfigure}[t]{0.49\linewidth}
        \includegraphics[width=\linewidth]{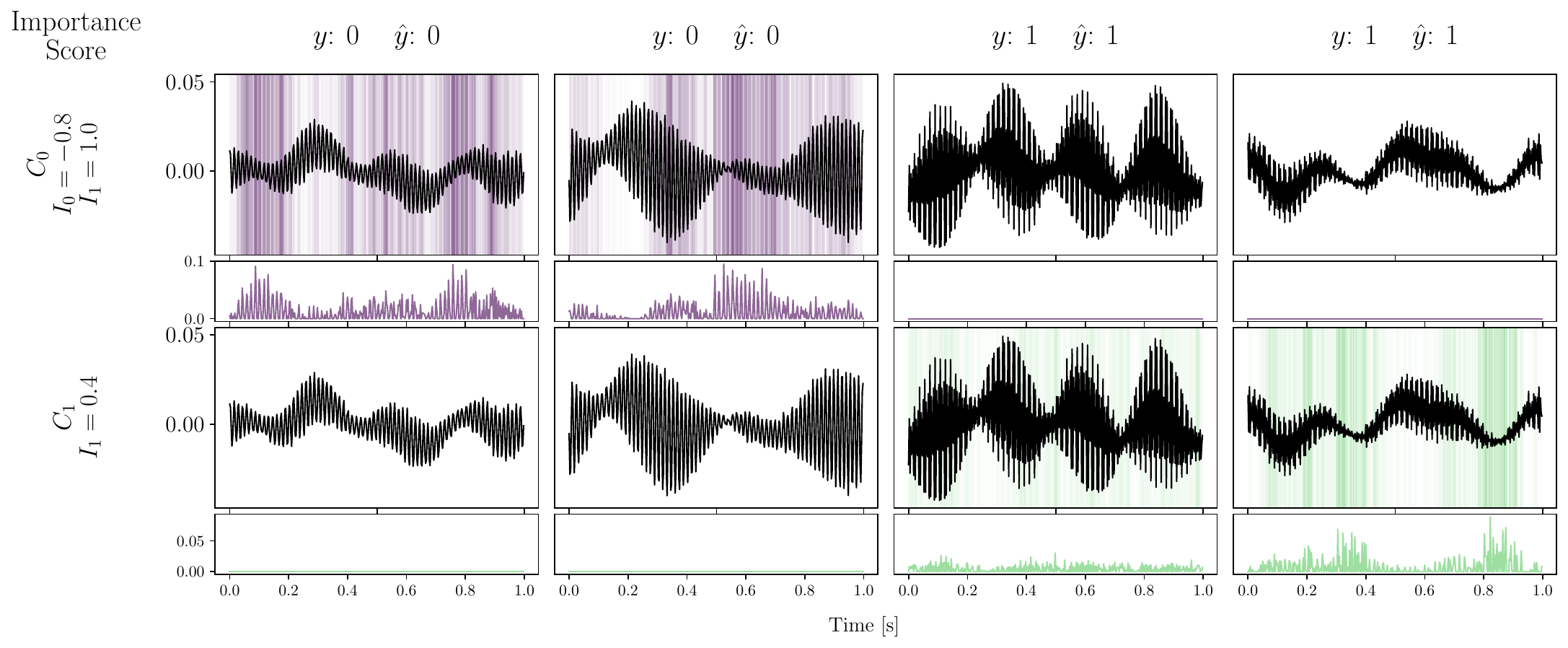}
        \caption{Time domain}
        \label{fig:tfdomains-ts}
    \end{subfigure}\hfill
    \begin{subfigure}[t]{0.49\linewidth}
        \includegraphics[width=\linewidth]{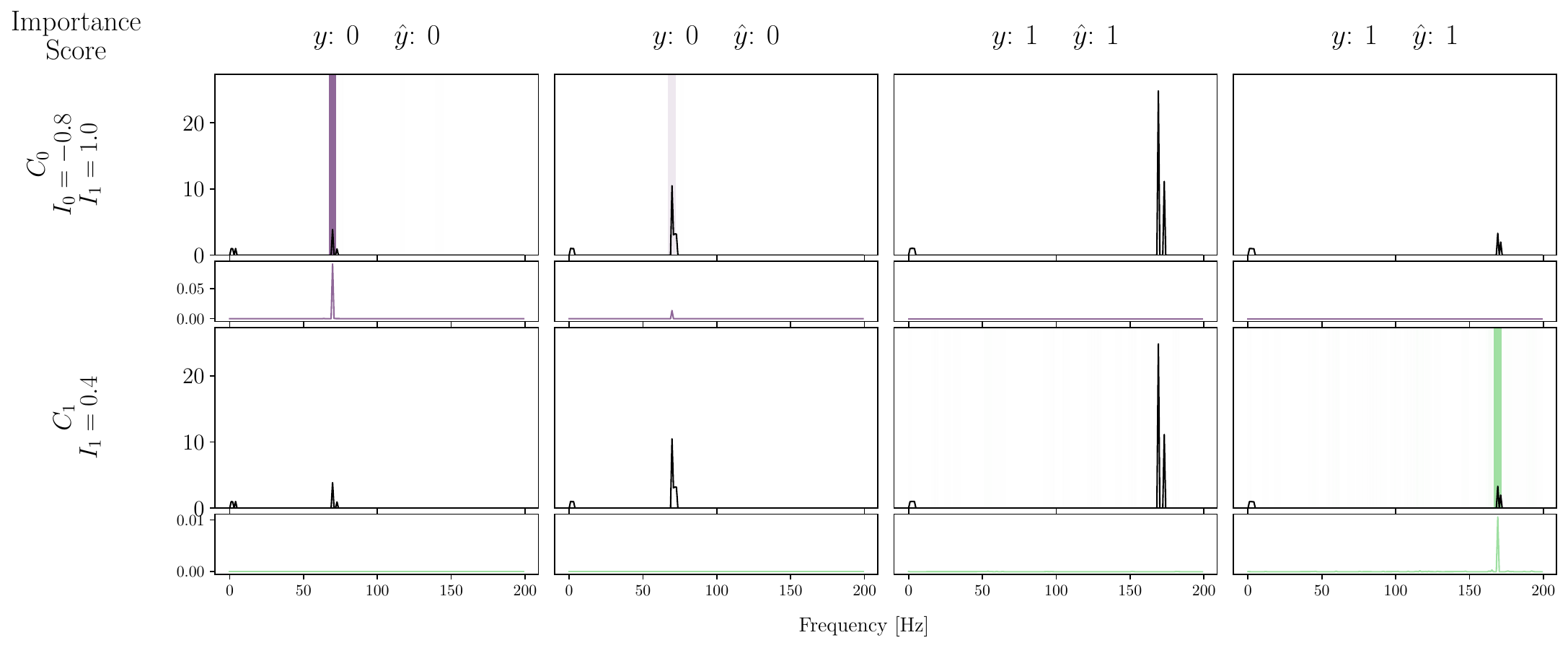}
        \caption{Frequency domain (rFFT)}
        \label{fig:tfdomains-f}
    \end{subfigure}
    \vspace{0.75em}
    \begin{subfigure}[t]{0.49\linewidth}
        \includegraphics[width=\linewidth]{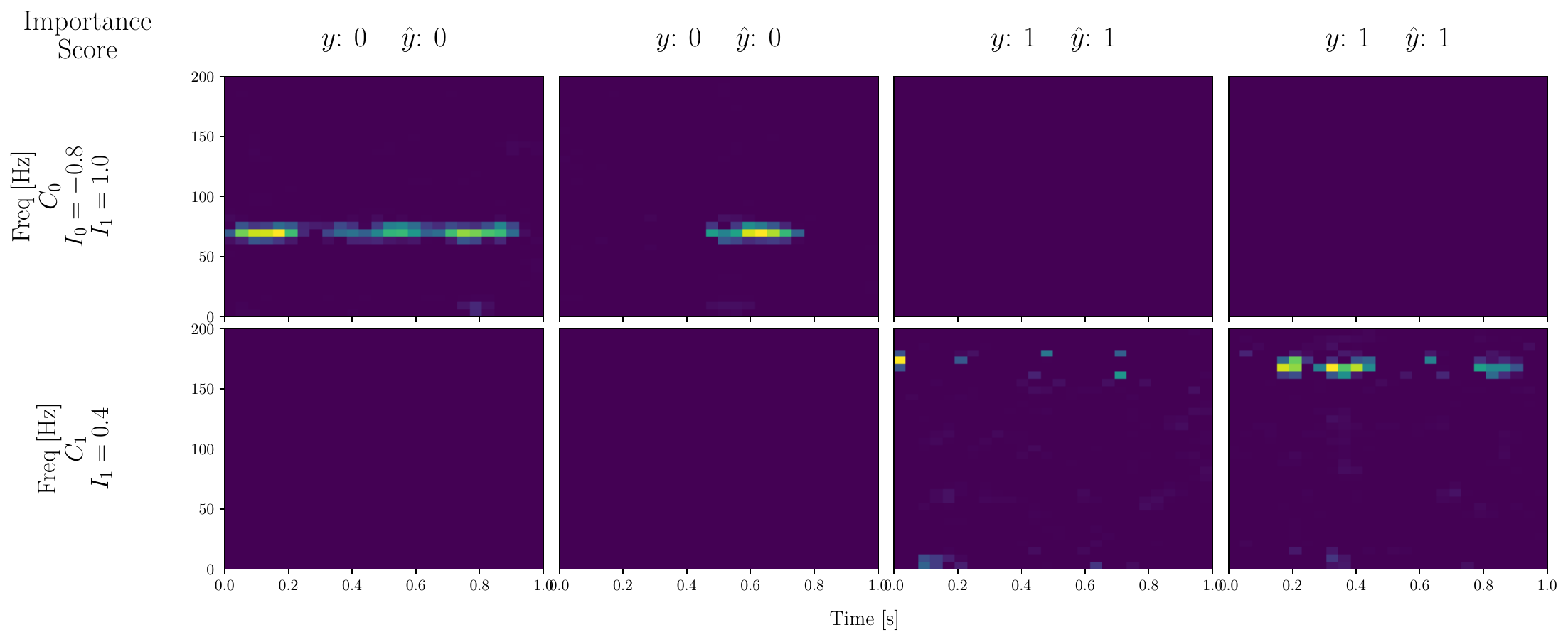}
        \caption{Time-frequency domain (STFT)}
        \label{fig:tfdomains-stft}
    \end{subfigure}\hfill
    \begin{subfigure}[t]{0.49\linewidth}
        \includegraphics[width=\linewidth]{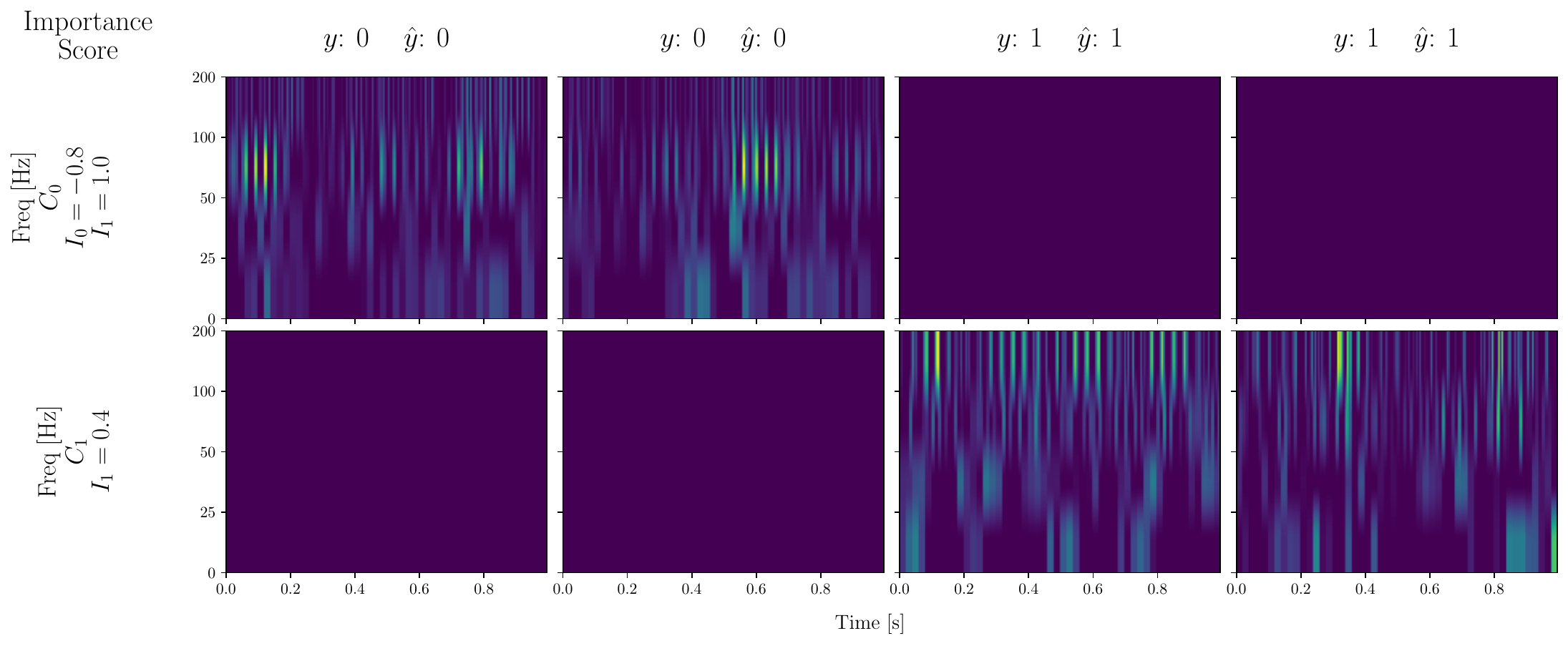}
        \caption{Time-frequency domain (wavelet)}
        \label{fig:tfdomains-wt}
    \end{subfigure}
        \caption{The same two concepts, extracted with CENDRe$_\textrm{silhouette}$ from an InceptionTime10 trained on the \textsc{syntheticFrequency} \texttt{midFreq-highFreq} variant, visualized under four choices of the transform $\phi$.
    The time-domain masks (a) carry no interpretable pattern, while all three spectral views place $C_0$ in the ground-truth midFreq band at $[70,75]$\,Hz.
    The rFFT (b) resolves that band as a sharp peak and also places $C_1$ in the highFreq band at $[170,175]$\,Hz, though only on one of the class 1 samples.
    The STFT (c) highlights $C_1$ in both class one samples, and shows that it is active only at isolated time frames, whereas $C_0$ persists across the full record.
    The DWT (d) reproduces the same time structure at coarser frequency resolution, since its highest dyadic band spans $100$ to $200$\,Hz and cannot separate the highFreq band inside it.}
    \label{fig:tfdomains}
\end{figure}

In Figure~\ref{fig:tfdomains} we show qualitative results of reports in the time-frequency domain using such transforms.
Across the four views, both concepts land where the ground truth places them, but the choice of $\phi$ decides how granular that visualization is.
The FFT already isolates $C_0$ in the midFreq band, while its $C_1$ mask spreads over a wide range with no dominant peak.
The STFT resolves $C_1$ into the highFreq band and shows it active at isolated time frames rather than across the whole record.
Comparing $C_0$ in the FFT and STFT views shows that the lower activation magnitude of the concept corresponds with it being active in a shorter timespan. 
The DWT recovers the same time structure, but its highest band spans $100$ to $200$\,Hz, so the highFreq band cannot be separated within it.

\subsubsection{Quantitative evaluation: summary per dataset} 
\label{sec:app-distributional}
In Figure~\ref{fig:quant} of Section~\ref{sec:results}, each box summarizes the sRC and sIC values across all datasets of a given family.
Tables~\ref{tab:distributional_summary_sRC} and~\ref{tab:distributional_summary_sIC} report the same runs resolved per dataset, as mean $\pm$ standard deviation with a $95\%$ confidence interval over the $33$ runs per cell ($3$ architectures $\times$ $11$ seeds), with the best method per dataset in bold.
The interval is the standard $95\%$ Student-$t$ confidence interval for the mean, $\bar{y} \pm t_{0.975,\,n-1}\,s/\sqrt{n}$. 
The family-level ranking of Figure~\ref{fig:quant} holds dataset by dataset and is not an artifact of pooling.
\begin{table}[!hbtp]
\centering\scriptsize
\setlength{\tabcolsep}{3pt}
\caption{Distributional summary (mean $\pm$ std, 95\% CI) of sRC across CE methods and datasets.}
\resizebox{\textwidth}{!}{\input{tables/table_distributional_summary_rma_sRC}}
\label{tab:distributional_summary_sRC}

\caption{Distributional summary (mean $\pm$ std, 95\% CI) of sIC across CE methods and datasets.}
\resizebox{\textwidth}{!}{\input{tables/table_distributional_summary_rma_sIC}}
\label{tab:distributional_summary_sIC}
\end{table}

On \textsc{syntheticLocal}, ECLAD-ts attains the highest sRC on all $9$ datasets, with a family mean of $0.741$ $[0.731, 0.751]$ separated from CENDRe$_{k\text{Means}}$'s $0.704$ $[0.695, 0.714]$, the next-best method. 
CENDRe$_{\text{silhouette}}$ and CENDRe$_{\text{HDBSCAN}}$ trail ECLAD-ts by $0.09$ and $0.11$ sRC respectively, on every dataset and with comparable spread, while MultiVISION is last on sRC everywhere.
The ordering reverses in sIC, where CENDRe$_{\text{HDBSCAN}}$ is best on all nine datasets (family mean $0.858$), ahead of CENDRe$_{\text{silhouette}}$ ($0.792$) and MultiVISION ($0.722$), with CENDRe$_{k\text{Means}}$ at $0.658$ and ECLAD-ts at $0.556$.
ECLAD-ts is also the least stable method under sIC, with per-dataset standard deviations up to $0.305$ against $0.140$ for the two automatic CENDRe variants.
Its sRC lead therefore does not transfer to sIC: broader masks that cover the ground-truth primitive, selected at the $K$ with the best sRC, do not track the model's per-timestep importance signal.

On \textsc{syntheticFrequency}, only the CENDRe variants are applicable, since the baselines have no frequency-domain counterpart, which the tables mark as ``--''.
Both automatic variants lose about $0.06$ sRC to CENDRe$_{k\text{Means}}$ and gain about $0.09$ sIC, on every dataset of the family, the same trade observed on \textsc{syntheticLocal}.
CENDRe$_{\text{silhouette}}$ and CENDRe$_{\text{HDBSCAN}}$ behave similarly on \textsc{syntheticFrequency} (family means within $0.001$ sRC and sIC of one another) though with $11$ seeds their per-run scores no longer coincide everywhere, matching exactly in $42\%$ of the (dataset, architecture, seed) combinations. 
Compared with \textsc{syntheticLocal}, mean sRC is lower ($0.510$ against $0.651$ for CENDRe$_{\text{silhouette}}$) while mean sIC is unchanged ($0.792$ in both cases).
Frequency-domain concepts thus track the model's importance signal as well as time-domain ones, while matching the ground-truth band less tightly.

\subsubsection{Quantitative evaluation: statistical significance}
\label{sec:app-significance}
\begin{figure}[!hbt]
    \centering
    \begin{subfigure}[t]{\linewidth}
        \centering
        \includegraphics[width=\linewidth]{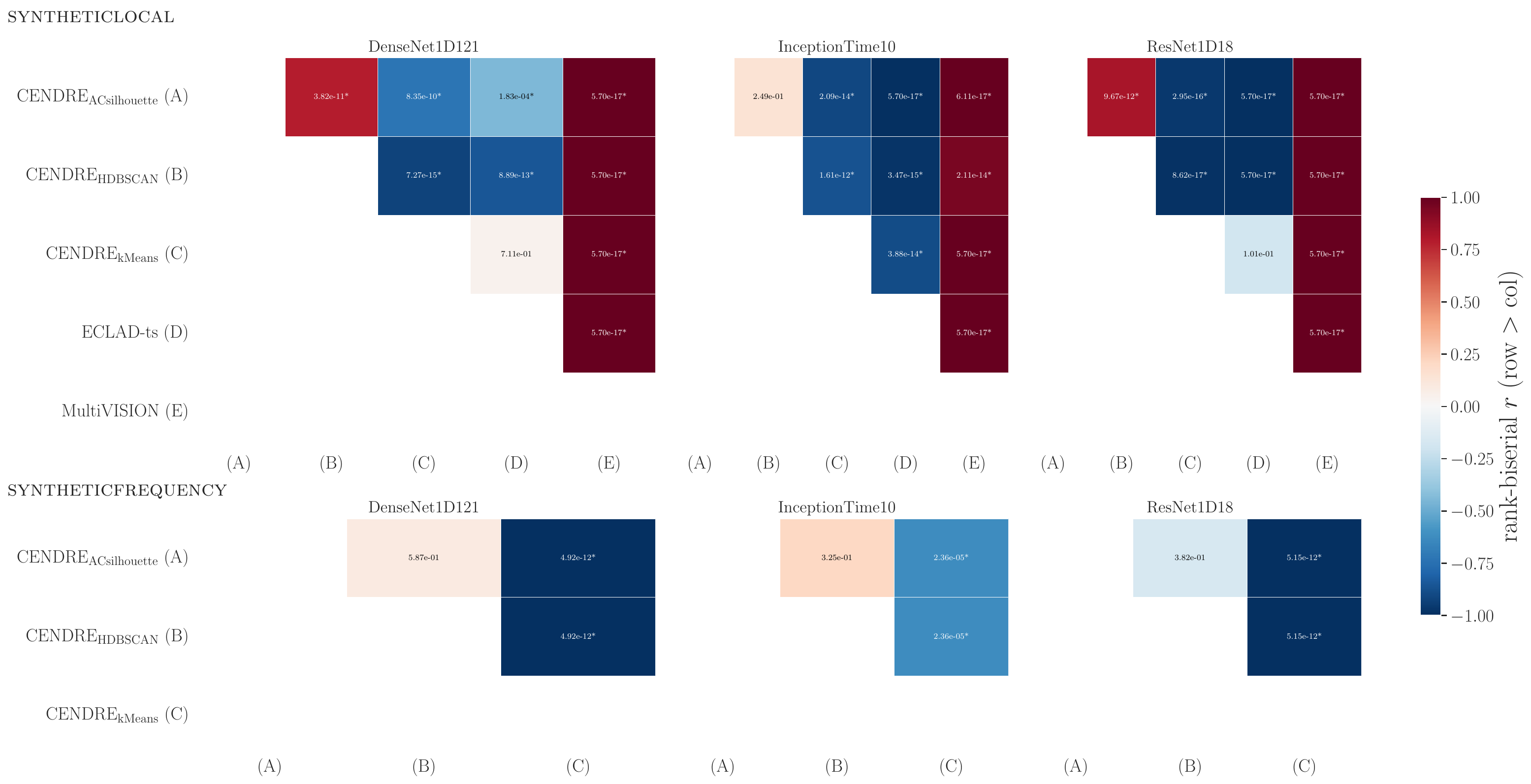}
        \caption{sRC}
        \label{fig:pairwise-sRC}
    \end{subfigure}

    \begin{subfigure}[t]{\linewidth}
        \centering
        \includegraphics[width=\linewidth]{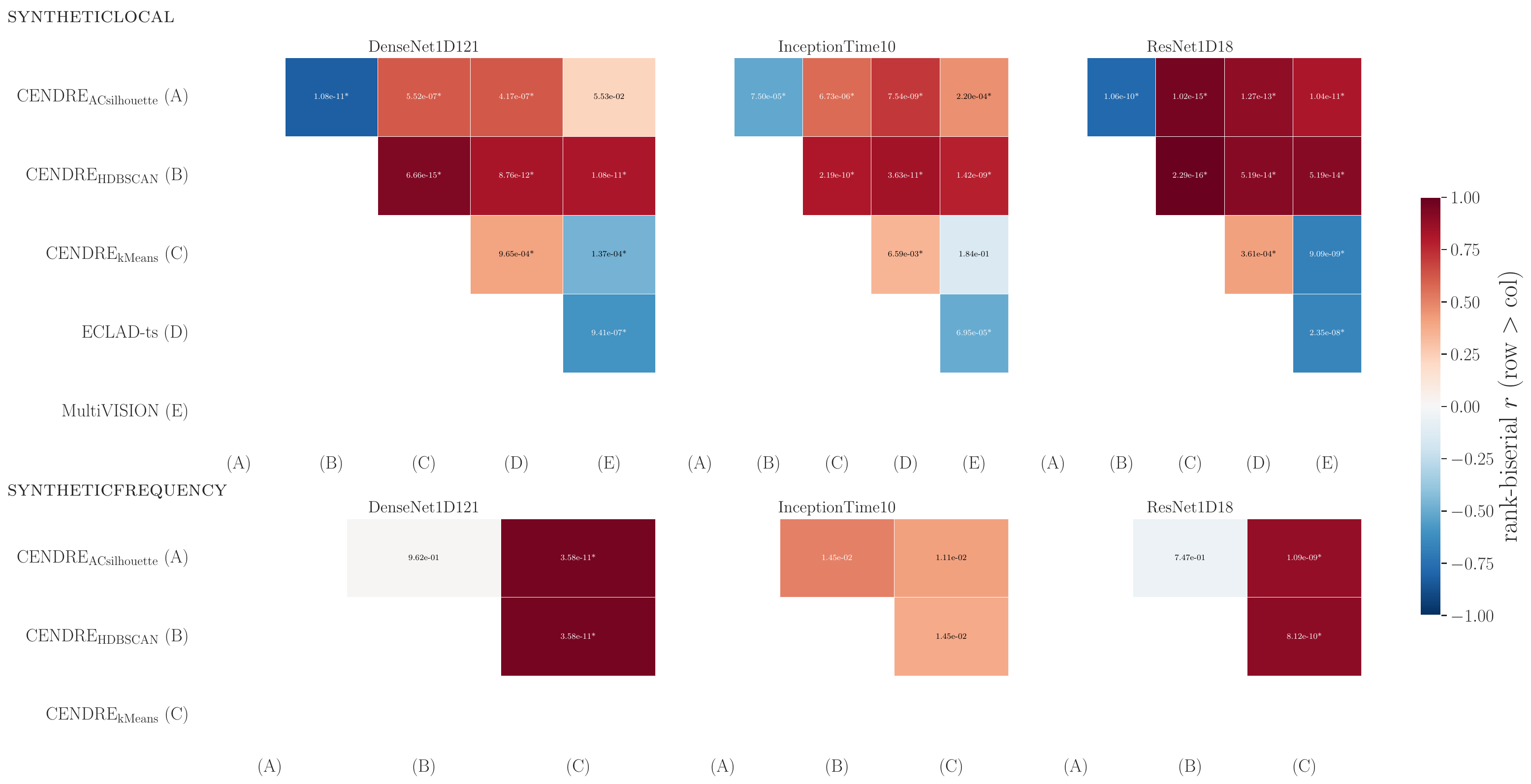}
        \caption{sIC}
        \label{fig:pairwise-sIC}
    \end{subfigure}
    \caption{Pairwise Holm-corrected Wilcoxon signed-rank tests between CE methods, per architecture, on the runs of Figure~\ref{fig:quant}, for (a) sRC and (b) sIC.
    Within each panel, the upper block is \textsc{syntheticLocal} and the lower block \textsc{syntheticFrequency}, and the three column groups are the DenseNet1D-121, the InceptionTime10, and the ResNet1D-18.
    Only the upper triangle is shown, since the test is symmetric.
    Cell color gives the matched-pairs rank-biserial correlation $r$ (red for row above column, blue for row below), and cell text the Holm-adjusted $p$-value, with an asterisk marking $p<0.01$.
    On \textsc{syntheticFrequency}, only the three CENDRe variants enter the comparison, since the baselines have no frequency-domain counterpart.}
    \label{fig:pairwise}
\end{figure}

Section~\ref{sec:results} compares the methods by reading the boxplots of Figure~\ref{fig:quant}, where an ordering is visible but not tested.
We test that ordering here, with no additional experiments.
Each run is one combination of dataset, architecture, and seed, and every applicable method is evaluated on all of them, so the scores form matched pairs.
We test each architecture separately over $11$ seeds, giving $99$ matched pairs per method pair on \textsc{syntheticLocal} ($9$ datasets $\times$ $11$ seeds) and $66$ on \textsc{syntheticFrequency} ($6$ datasets $\times$ $11$ seeds).
Runs in which CENDRe$_{\text{HDBSCAN}}$ returned no concepts are dropped from the pairs involving it.
We compare each pair with a two-sided Wilcoxon signed-rank test~\cite{wilcoxon1992individual}, and Holm-correct the $p$-values over all pairs within a family, metric, and architecture~\cite{holm1979simple}.
We accompany each test with the matched-pairs rank-biserial correlation $r$~\cite{kerby2014simple}, the share of pairs favoring the row method minus the share favoring the column method.
Figure~\ref{fig:pairwise} reports both, with cell color giving $r$ (red for row above column, blue for row below) and cell text the Holm-adjusted $p$-value.
On \textsc{syntheticFrequency}, only the three CENDRe variants enter the comparison, since the baselines have no frequency-domain counterpart.
Of the $78$ tests in Figure~\ref{fig:pairwise} ($10$ method pairs on \textsc{syntheticLocal} and $3$ on \textsc{syntheticFrequency}, per architecture and metric), $65$ reach $p<0.01$, and the sign of $r$ follows the ordering of \S\ref{sec:app-distributional} throughout.
The tests therefore confirm the reading of Figure~\ref{fig:quant}.
The $13$ pairs that do not separate concentrate in three places, with two isolated exceptions.

The first is CENDRe$_{\text{silhouette}}$ against CENDRe$_{\text{HDBSCAN}}$ on \textsc{syntheticFrequency}, which separates on neither metric nor any architecture ($p \geq 0.014$ across all six tests).
This matches their family means in \S\ref{sec:app-distributional}, which differ by less than $0.001$ in both metrics.
On \textsc{syntheticLocal} the two criteria separate everywhere except in sRC on the InceptionTime10 ($p=0.25$).
Both methods select the number of centroids automatically and visualize concepts by the same means, so similar behaviour on some datasets is expected.

The second is CENDRe$_{k\text{Means}}$ against ECLAD-ts in sRC on \textsc{syntheticLocal}, indistinguishable on the DenseNet1D-121 ($p=0.71$) and the ResNet1D-18 ($p=0.10$) but separated on the InceptionTime10.
The gap of $0.035$ sRC between their family means is therefore carried by a single architecture.
In sIC the same pair separates on all three, with ECLAD-ts behind.
The two methods share the same clustering step, so their scores differ mainly through the mode of concept visualization.
The visualizations converge when the CENDRe masks are dense, which can occur on individual architecture-dataset combinations.

The third is the InceptionTime10 in sIC on \textsc{syntheticFrequency}, where no pair of CENDRe variants reaches the $1\%$ level ($p$ between $0.011$ and $0.015$).
The other two architectures separate CENDRe$_{k\text{Means}}$ from both automatic variants there, so this reflects the backbone rather than the metric.

The two remaining cells both involve MultiVISION in sIC on \textsc{syntheticLocal}: against CENDRe$_{\text{silhouette}}$ on the DenseNet1D-121 ($p=0.055$) and against CENDRe$_{k\text{Means}}$ on the InceptionTime10 ($p=0.18$).
Both are consistent with MultiVISION ranking mid-field on sIC rather than last, and neither affects the ordering among the CENDRe variants or against ECLAD-ts.

\subsubsection{Multichannel localization and confounder robustness}
\label{sec:app-mc-conf}
\begin{figure*}[!tb]
    \begin{minipage}{.49\linewidth}
        \begin{subfigure}[t]{\linewidth}
            \includegraphics[width=\linewidth]{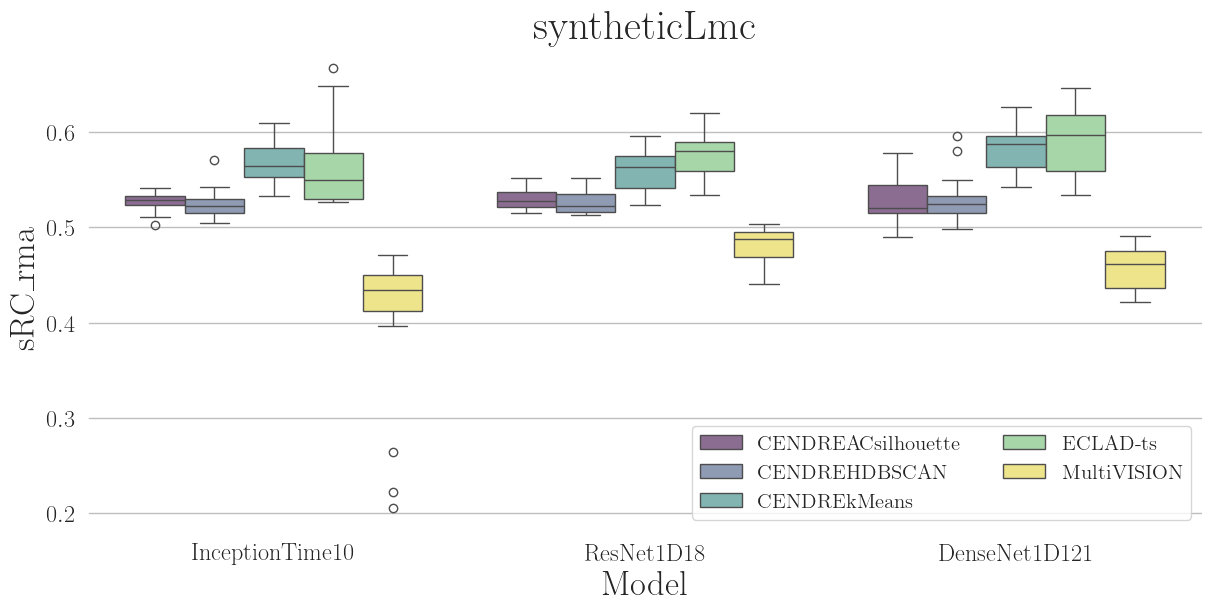}
        \end{subfigure}
    \end{minipage}%
    \begin{minipage}{.49\linewidth}
        \begin{subfigure}[t]{\linewidth}
            \includegraphics[width=\linewidth]{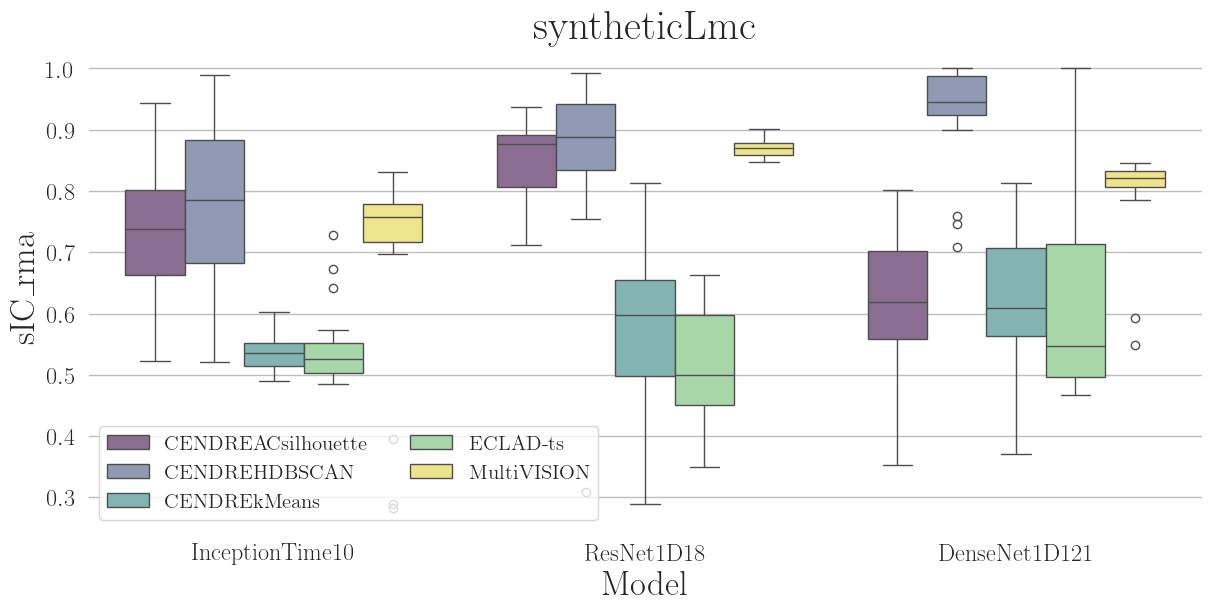}
        \end{subfigure}
    \end{minipage}

    \begin{minipage}{.49\linewidth}
        \begin{subfigure}[t]{\linewidth}
            \includegraphics[width=\linewidth]{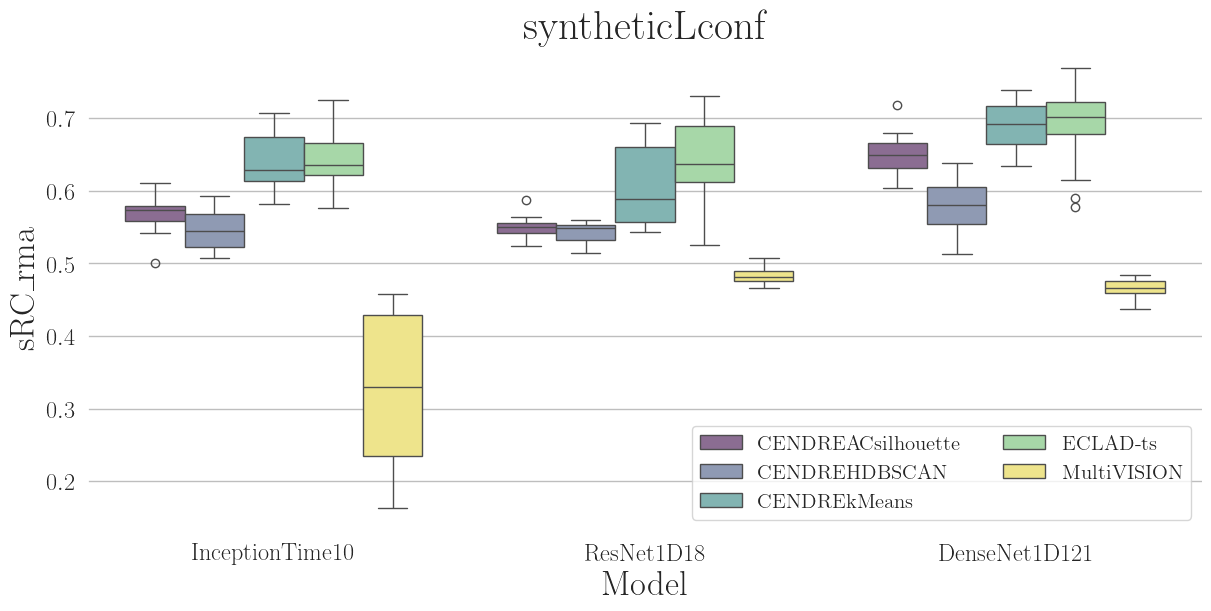}
        \end{subfigure}
    \end{minipage}%
    \begin{minipage}{.49\linewidth}
        \begin{subfigure}[t]{\linewidth}
            \includegraphics[width=\linewidth]{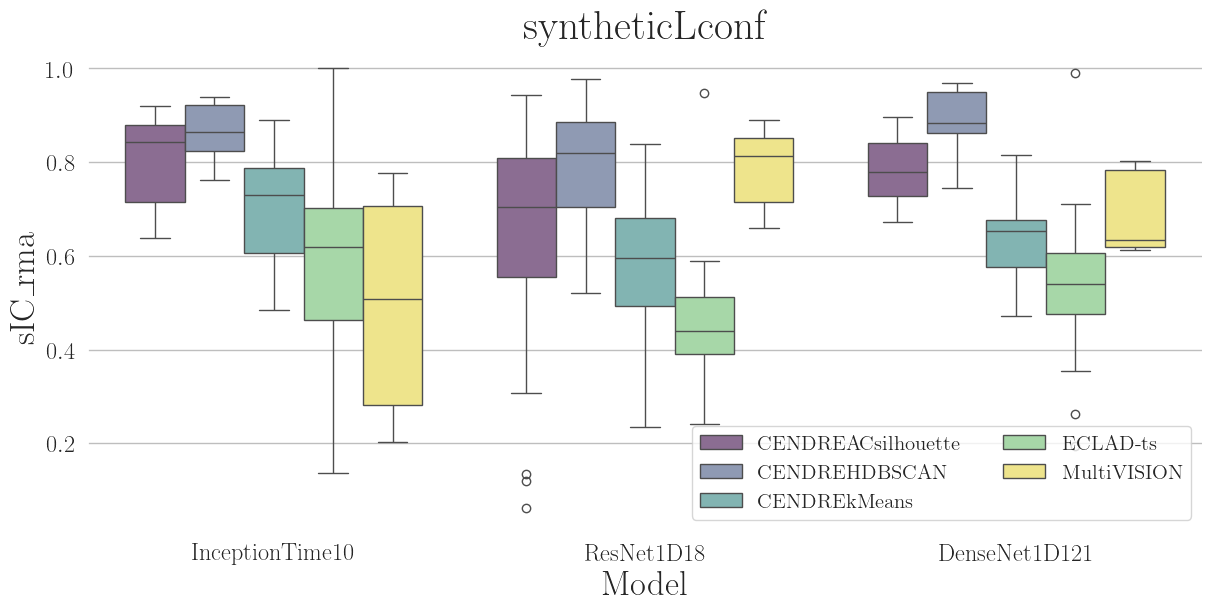}
        \end{subfigure}
    \end{minipage}
    \caption{Quantitative evaluation on \textsc{syntheticLmc} (top, per-channel attribution) and \textsc{syntheticLconf} (bottom, paired with confounders). The ranking mirrors Figure~\ref{fig:quant}: CENDRe$_\text{silhouette}$ and CENDRe$_\text{HDBSCAN}$ achieve sRC close to ECLAD-ts and higher sIC, CENDRe$_{k\text{Means}}$ tracks ECLAD-ts on both metrics, and MultiVISION is weak on sRC.}
    \label{fig:quant_mc_conf}
\end{figure*}

We extend the quantitative evaluation of Figure~\ref{fig:quant} to two synthetic datasets: \textsc{syntheticLmc} ($D{=}2$, per-channel attribution) and \textsc{syntheticLconf} ($D{=}1$, paired with a third shape as confounder at $40/60$, $50/50$, and $60/40$ ratios).
Figure~\ref{fig:quant_mc_conf} reports sRC and sIC across all variants and the three architectures.
These two families run over $3$ seeds ($0$--$2$), rather than the $11$ of the main families.
The ranking is similar to the main paper. CENDRe$_\text{silhouette}$ and CENDRe$_\text{HDBSCAN}$ trade slightly lower sRC for noticeably higher sIC than ECLAD-ts, CENDRe$_{k\text{Means}}$ is similar to ECLAD-ts on both metrics, and MultiVISION underperforms on sRC.

Across these settings, CENDRe$_{\text{HDBSCAN}}$ reaches slightly higher sIC than CENDRe$_{\text{silhouette}}$, consistent with density-based clustering being better suited to the non-spherical structures one expects in latent spaces.
We still report silhouette-guided aggregation as the default in the main text because HDBSCAN occasionally requires per-dataset tuning of its neighborhood parameters when its single-pass run fails to find any dense region (see failure rates below); on more complex problems where this tuning effort is acceptable, HDBSCAN may be the stronger choice.
CENDRe$_{\text{HDBSCAN}}$ occasionally fails to converge, with HDBSCAN labeling all $50$ micro-centroids as noise, on $11/297$ \textsc{syntheticLocal} and $5/54$ \textsc{syntheticLconf} runs (all on InceptionTime10 or ResNet1D-18, concentrated on \texttt{triangle-circle} bases). 
No failures occur on \textsc{syntheticFrequency} ($0/198$) or \textsc{syntheticLmc}
($0/54$).
Failed runs are excluded from the boxplot statistics, so the boxplots are conditional on convergence.

\subsection{Natural datasets}
\label{sec:app-additional-natural}
We complement the CWRU results of Section~\ref{sec:results} with two further natural-data evaluations.
We extend the qualitative evaluation to five univariate datasets from the UCR archive, covering cardiac, automotive, motion, robotic, and semiconductor signals (\S\ref{sec:app-ucr}).
We then extract concepts from a second bearing-fault dataset, Paderborn Bearing (BearingPD), on two architectures, to test whether CENDRe recovers the same fault-relevant frequency bands under a different rig and sensor setup.

\subsubsection{UCR}
\label{sec:app-ucr}
We extend the qualitative evaluation to five univariate datasets of the UCR archive~\cite{UCRArchive2018}, covering cardiac, automotive, motion, robotic, and semiconductor signals.
For each dataset we train the three architectures of Section~\ref{sec:setup} and extract concepts from one of them, using CENDRe$_{\text{silhouette}}$ with the hyperparameters of Section~\ref{sec:ce-hparams}.
Each report shows the time-domain masks, the frequency-domain masks, and the global concept-frequency correspondence.
The archive distributes most of its datasets without the sampling rate of the original recording.
Where the rate is undocumented we plot the frequency axis in cycles per sample, under $f_s = 1$, rather than assume a rate.
Where it is documented we state it in the corresponding paragraph and label the time and frequency axes in seconds and Hz.
The choice affects the labeling of the axes alone, since the masks are computed on the sampled series either way.

\paragraph{ECG200}~\cite{UCRArchive2018} traces the electrical activity recorded during a single heartbeat, on one electrocardiogram channel.
The two classes are a normal heartbeat and a myocardial infarction.
Series have $T = 96$ samples.
CENDRe$_{\text{silhouette}}$ extracts $K = 4$ concepts, which partition the beat into consecutive segments.
$C_0$ and $C_1$ carry the largest and opposite-signed importance, with $C_0$ activating on the plateau that follows the main negative deflection, mostly in the $y=1$ series, and $C_1$ on the deflection itself.
$C_2$ and $C_3$ cover the late and the early part of the beat at lower importance.
In frequency, all four concepts keep their mass below $0.25$ cycles per sample and peak near $0.02$, so the model reads the morphology of the beat rather than high-frequency content.
\begin{figure}[!htb]
    \centering
    \reportpanels
        {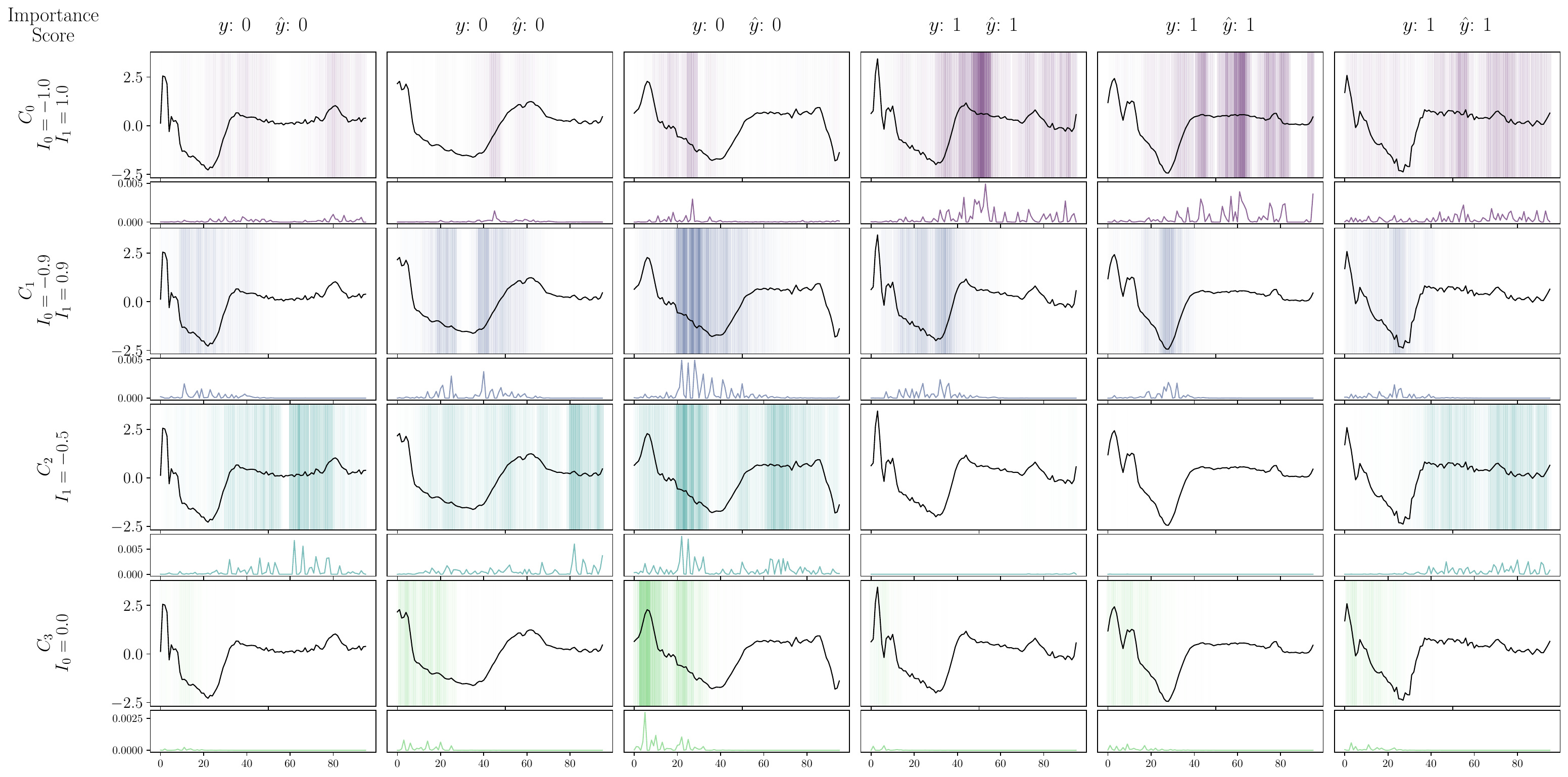}
        {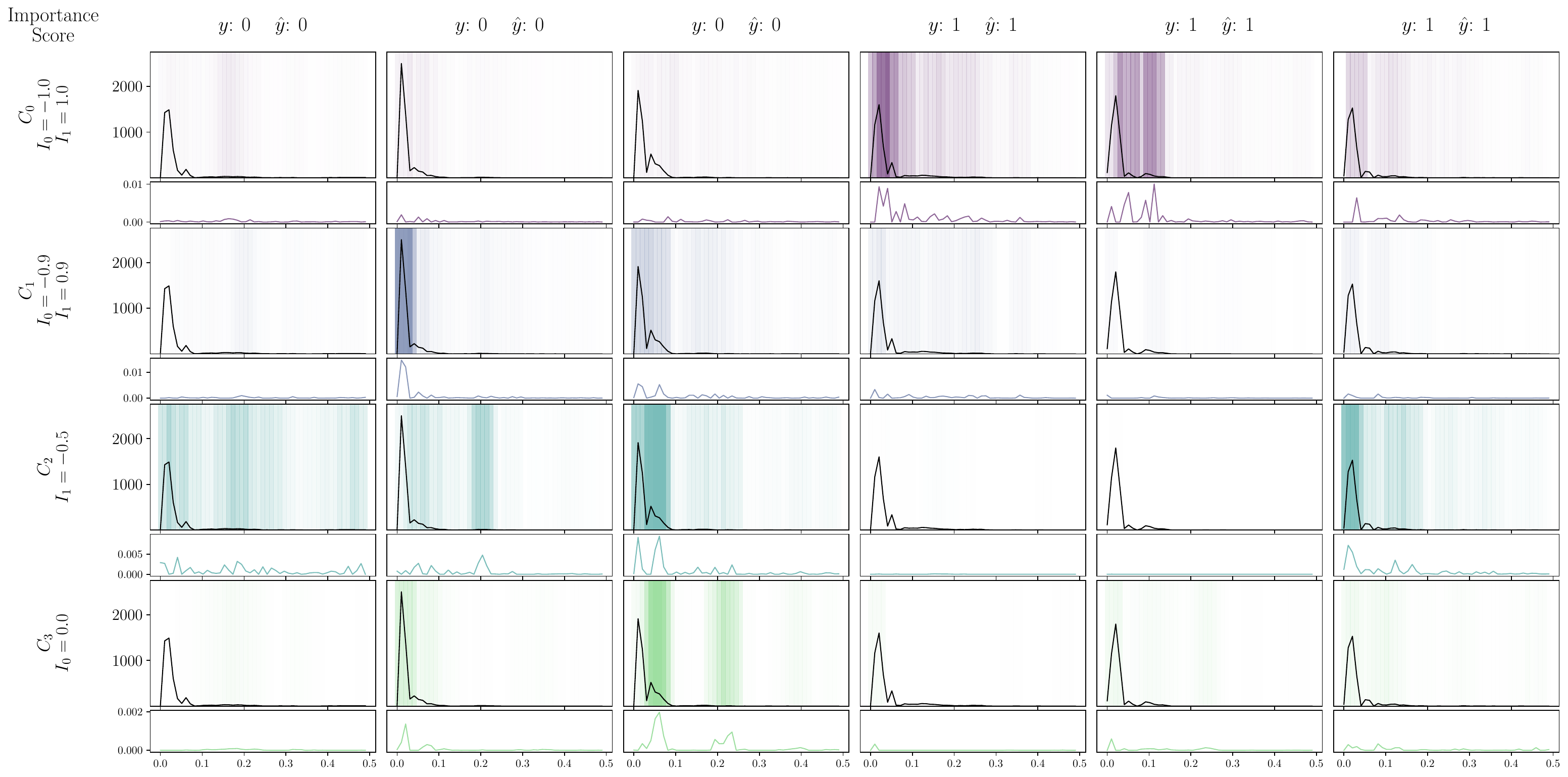}
        {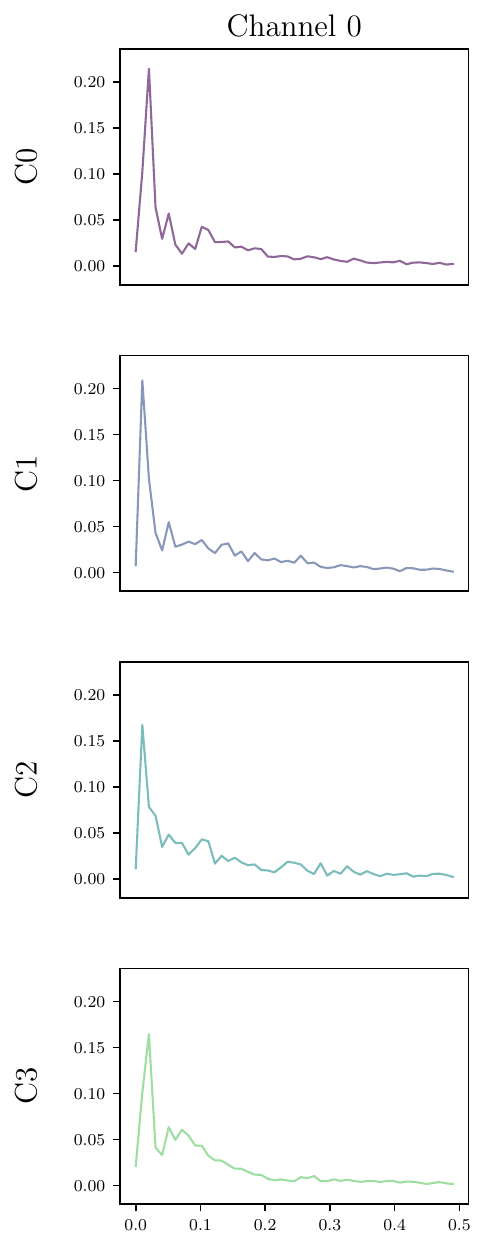}
    \caption{Concept extraction from a DenseNet1D-121 trained on ECG200 (two classes: normal heartbeat, myocardial infarction), reaching $91.67\%$ test accuracy.
    Top left: time-domain concept masks.
    Bottom left: frequency-domain concept masks.
    Right: global concept-frequency correspondence.
    The concepts split the beat into consecutive segments, and none of them reads content above $0.25$ cycles per sample.
    }
    \label{fig:ucr-ecg200}
\end{figure}

\paragraph{FordA}~\cite{UCRArchive2018} contains engine noise measurements from an automotive subsystem, with the task of deciding whether a given symptom is present.
Series have $T = 500$ samples, recorded under typical operating conditions and with minimal contamination.
CENDRe$_{\text{silhouette}}$ extracts $K = 2$ concepts with opposite class importance.
$C_0$ concentrates on short high-amplitude bursts and $C_1$ on the remaining oscillation, the two alternating along the window instead of splitting it into contiguous segments.
Both confine their frequency mass to the band below $0.15$ cycles per sample, which carries the dominant spectral peak of the signal, and both peak near $0.05$.
Within that band the two profiles differ in shape: $C_0$ concentrates its mass around the $0.05$ peak and decays quickly above it, 
while $C_1$ spreads a broader shoulder that persists toward $0.15$.
\begin{figure}[!htb]
    \centering
    \reportpanels
        {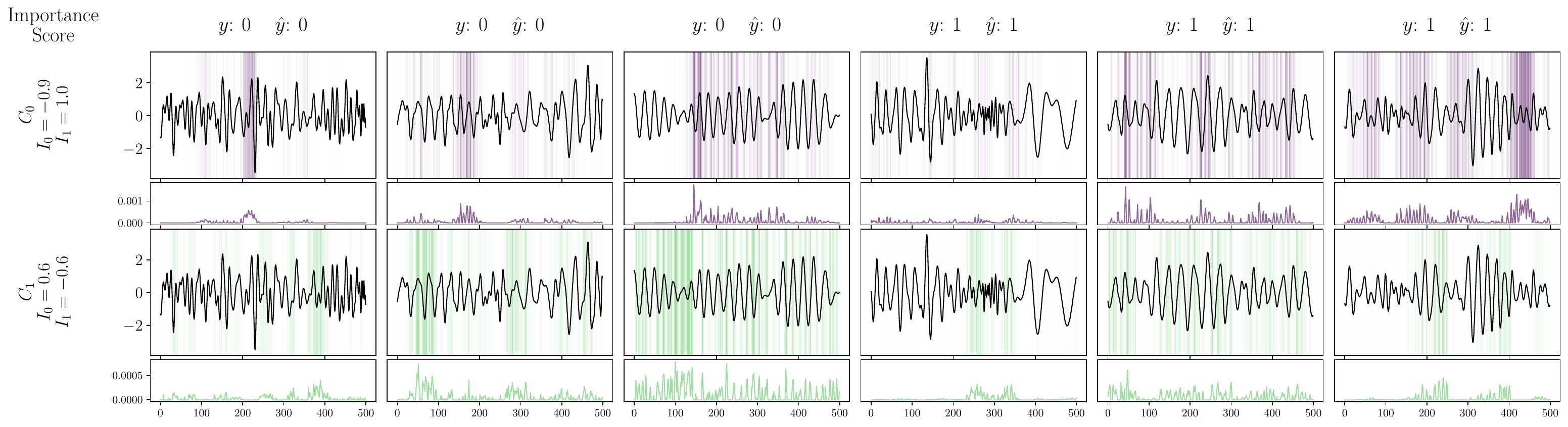}
        {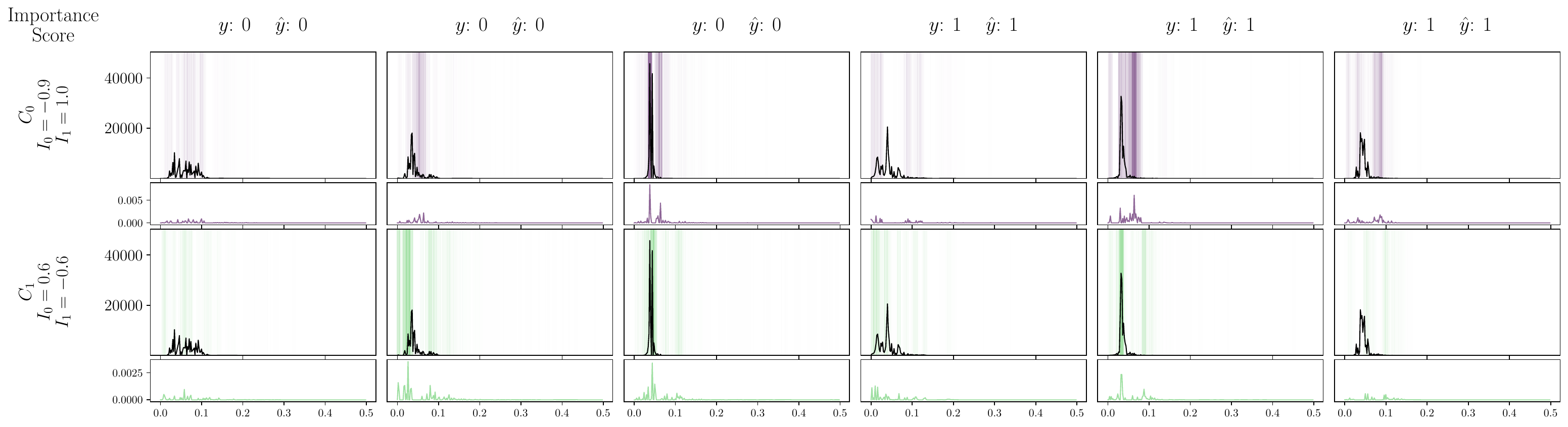}
        {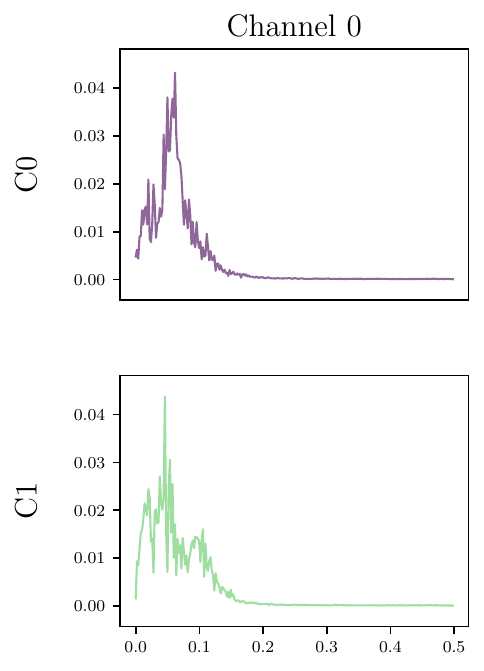}
    \caption{Concept extraction from an InceptionTime10 trained on FordA (two classes: symptom present, symptom absent), reaching $92.48\%$ test accuracy.
    Top left: time-domain concept masks.
    Bottom left: frequency-domain concept masks.
    Right: global concept-frequency correspondence.
    Both concepts peak near $0.05$ cycles per sample; $C_0$ concentrates its mass around that peak, while $C_1$ spreads a broader shoulder across the band below $0.15$.
    }
    \label{fig:ucr-forda}
\end{figure}

\paragraph{GunPoint}~\cite{UCRArchive2018} tracks the horizontal position of the centroid of an actor's right hand, extracted from video of one of two motions.
In the gun class the actor draws a replica gun from a hip holster, points it at a target for about one second, and returns it.
In the point class the actor performs the same motion with an extended index finger and no prop.
Series have $T = 150$ samples from two actors.
The motion was recorded at $30$\,Hz, so the axes are labeled in seconds and in Hz over $[0, 15]$.
CENDRe$_{\text{silhouette}}$ extracts $K = 4$ concepts, all of which localize on the transitions of the hand trajectory and none on the plateau where the hand is held on target.
$C_0$ and $C_1$ cover the rising edge around $1$--$2$\,s and $C_2$ the falling edge around $3$--$4$\,s in the $y=0$ series, while $C_3$ covers both edges in the $y=1$ series.
Every concept peaks between $0.3$ and $0.6$\,Hz and carries no mass above $2$\,Hz, on the order of the one-second gesture.
The extraction recovers the draw and the return as separate concepts, which is the distinction the two motions are defined by, as has been observed in literature~\cite{hills2014classification}.
\begin{figure}[!htb]
    \centering
    \reportpanels
        {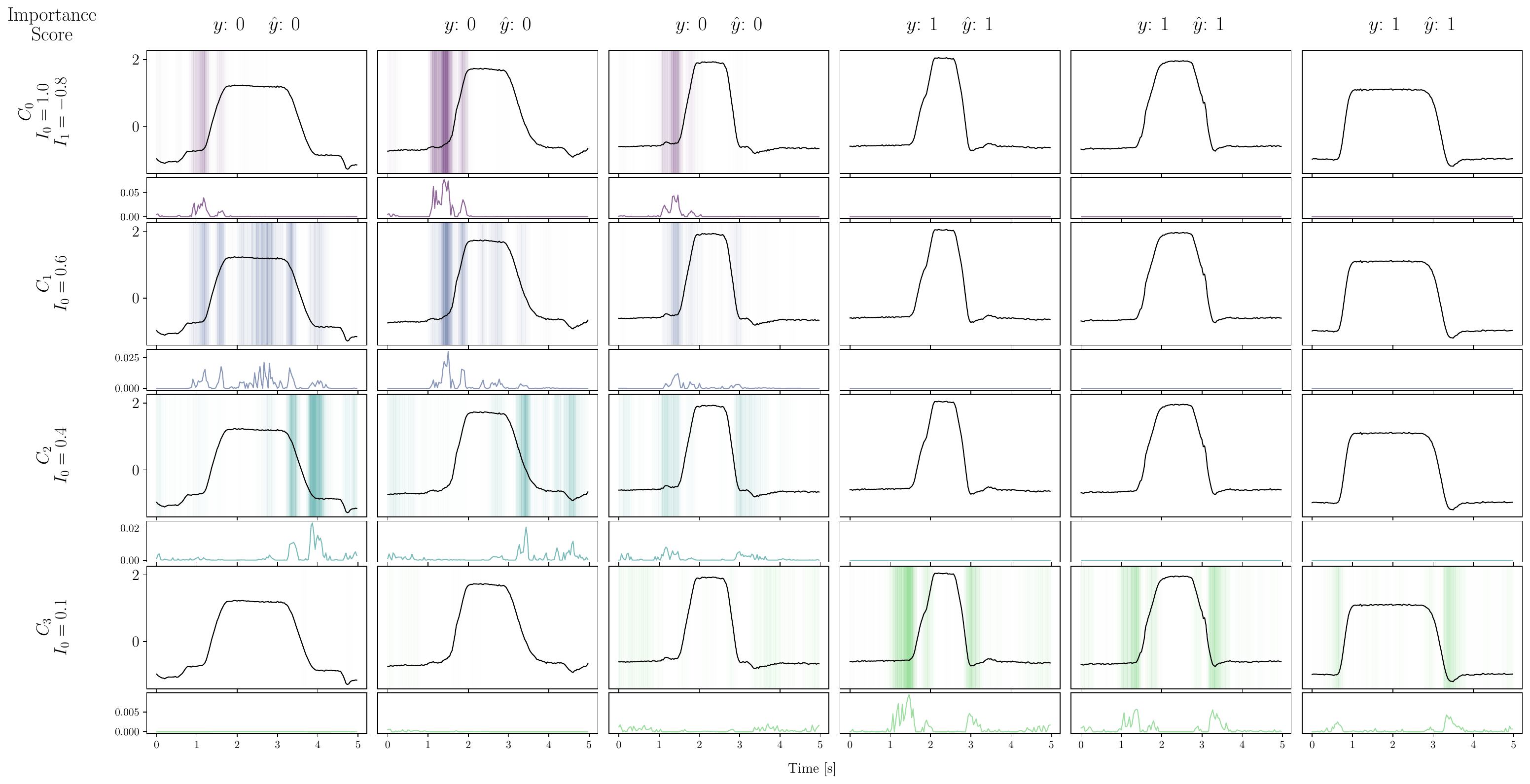}
        {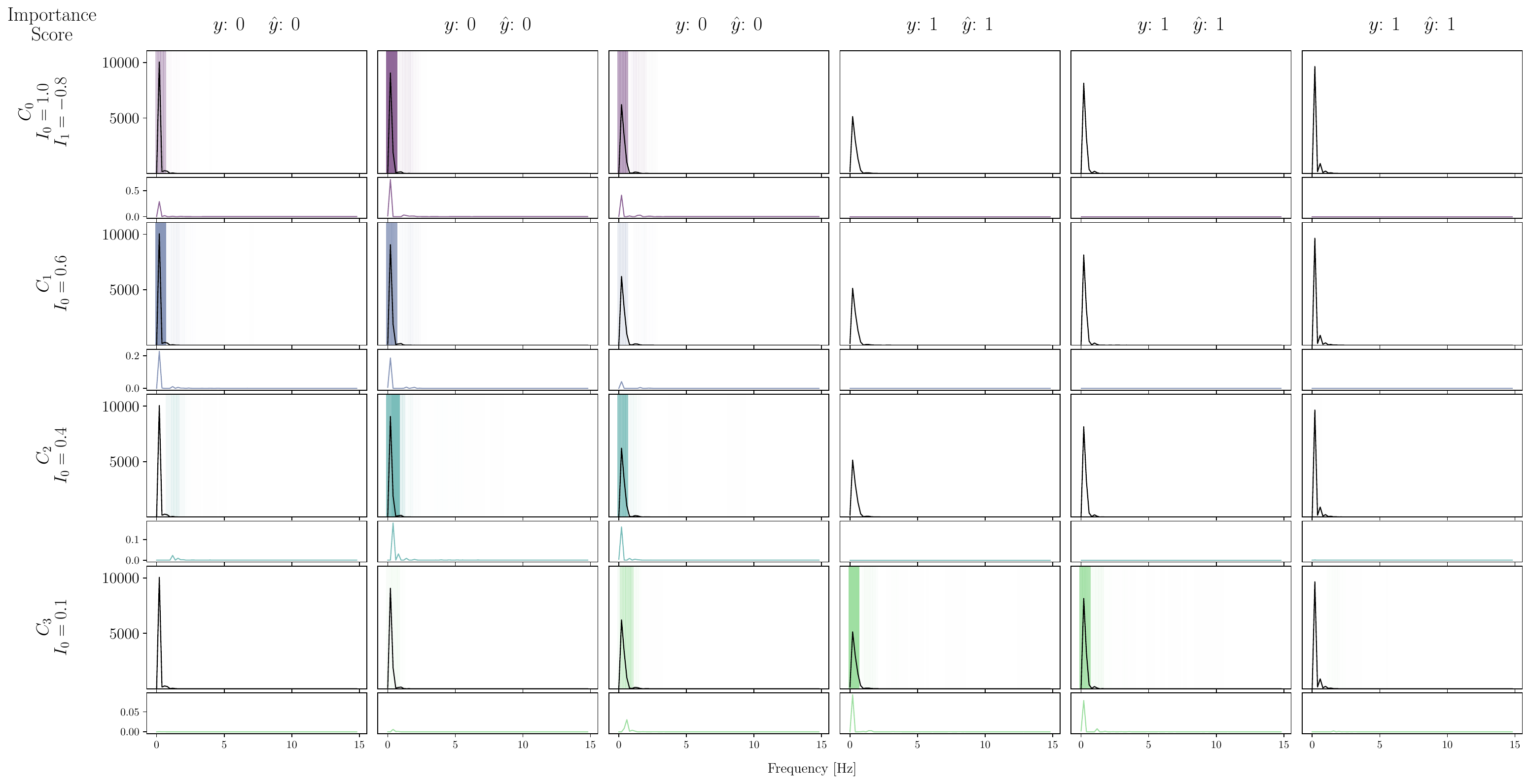}
        {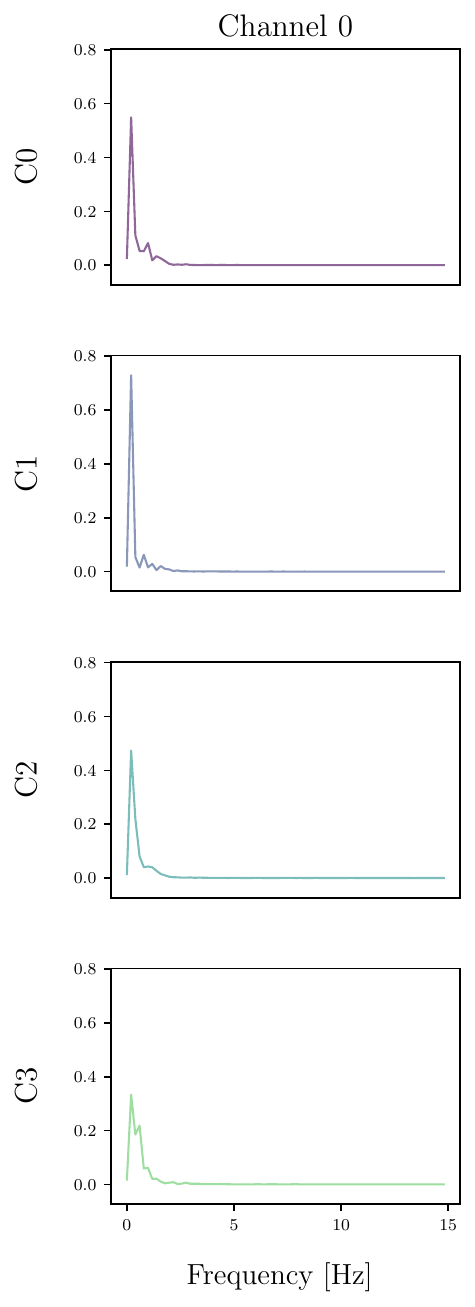}
    \caption{Concept extraction from an InceptionTime10 trained on GunPoint (two classes: gun draw, point), reaching $100\%$ test accuracy.
    Top left: time-domain concept masks.
    Bottom left: frequency-domain concept masks.
    Right: global concept-frequency correspondence.
    All four concepts localize on the draw and the return of the hand, and none on the plateau in between.
    }
    \label{fig:ucr-gunpoint}
\end{figure}

\paragraph{SonyAIBORobotSurface1}~\cite{UCRArchive2018} contains accelerometer readings from a Sony AIBO robot, taken while the robot walks on one of two surfaces.
The two classes are cement and carpet.
The robot carries roll, pitch, and yaw accelerometers, of which the archive distributes the $x$ axis alone.
Series have $T = 70$ samples.
The accelerometer samples at $125$\,Hz, so the axes are labeled in seconds and in Hz over $[0, 62.5]$.
CENDRe$_{\text{silhouette}}$ extracts $K = 2$ concepts that separate by class rather than by band.
$C_0$ activates on the $y=1$ series and $C_1$ on the $y=0$ series, each spread over the whole $0.55$\,s window instead of a contiguous segment.
Both concentrate their mass between $5$ and $25$\,Hz, peak near $8$\,Hz, and decay above $30$\,Hz, in the range of the step cadence of the walk.
Within that band the two profiles differ in shape: $C_0$ splits its mass into two sharp peaks near $8$ and $14$\,Hz, while $C_1$ pairs the $8$\,Hz peak with a lower but longer ridge that extends to about $30$\,Hz.
\begin{figure}[!htb]
    \centering
    \reportpanels
        {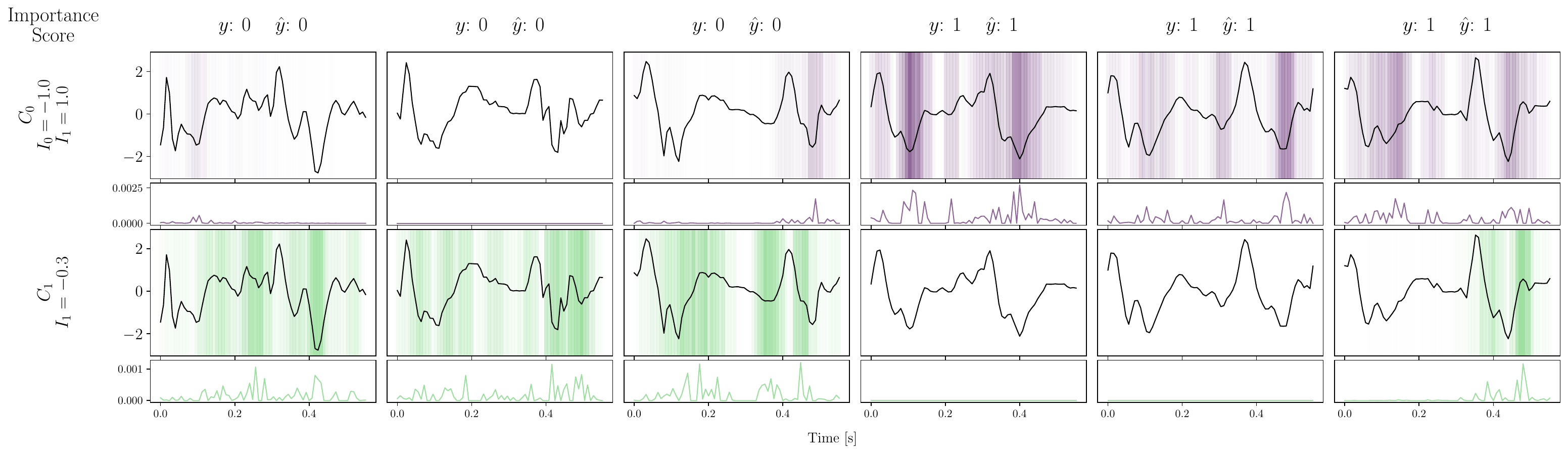}
        {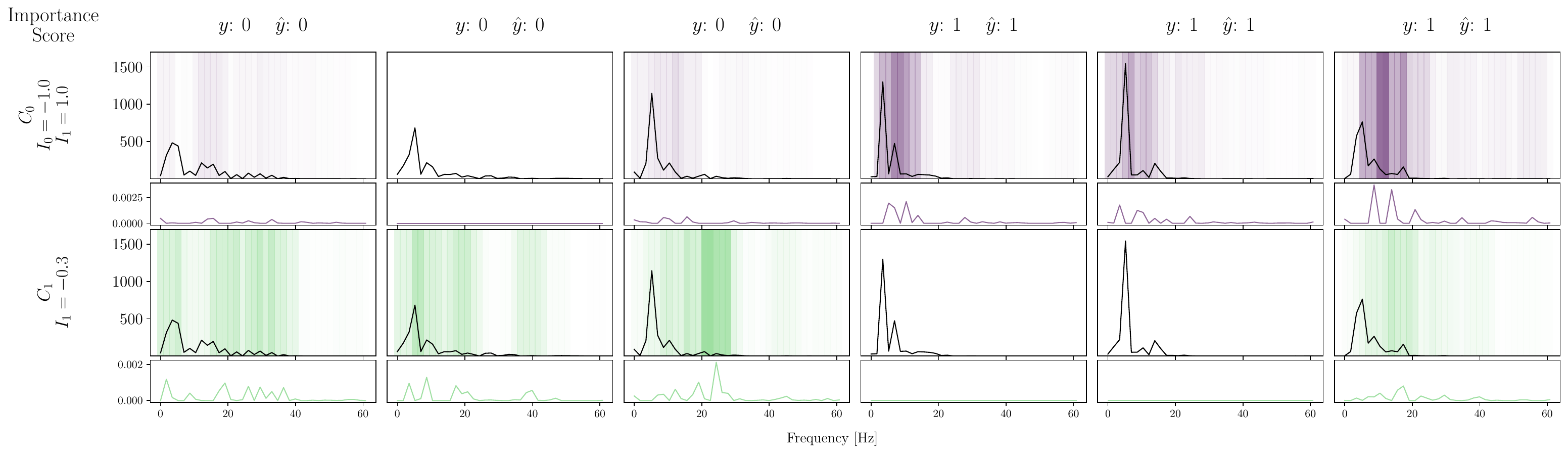}
        {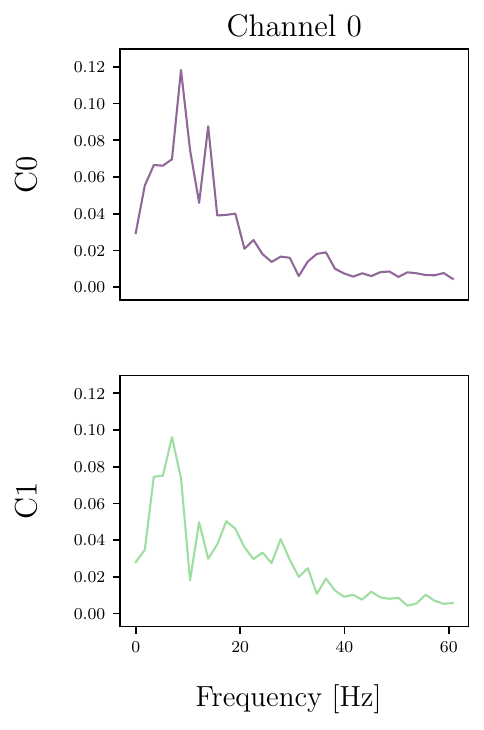}
    \caption{Concept extraction from a DenseNet1D-121 trained on SonyAIBORobotSurface1 (two classes: cement, carpet), reaching $97.50\%$ test accuracy.
    Top left: time-domain concept masks.
    Bottom left: frequency-domain concept masks.
    Right: global concept-frequency correspondence.
    Both concepts peak near $8$\,Hz within the gait band; $C_0$ adds a second sharp peak near $14$\,Hz, while $C_1$ spreads a lower ridge up to about $30$\,Hz.
    }
    \label{fig:ucr-sony}
\end{figure}

\paragraph{Wafer}~\cite{UCRArchive2018} contains inline process control measurements from semiconductor fabrication, each series recorded by one sensor while one tool processes one silicon wafer.
The two classes are normal and abnormal, with abnormal covering $10.7\%$ of the training and $12.1\%$ of the test series.
Series have $T = 152$ samples.
CENDRe$_{\text{silhouette}}$ extracts $K = 5$ concepts.
Four of them cover consecutive stages of the process trace, with $C_1$ and $C_4$ on the initial transient, $C_3$ on the plateau, and $C_2$ on the final ramp.
The fifth, $C_0$, carries the largest importance magnitude with opposite sign across the two classes and spans the whole window of the $y=1$ series.
The frequency masks are broader than on the other four datasets: $C_2$, $C_3$, and $C_4$ peak below $0.03$ cycles per sample, while $C_0$ and $C_1$ cover the full band, $C_1$ as a near-uniform comb.
That comb is the signature of the step-like transients of the trace rather than of a narrowband feature, which is what a purely spectral summary of this model would misread.
\begin{figure}[!htb]
    \centering
    \reportpanels
        {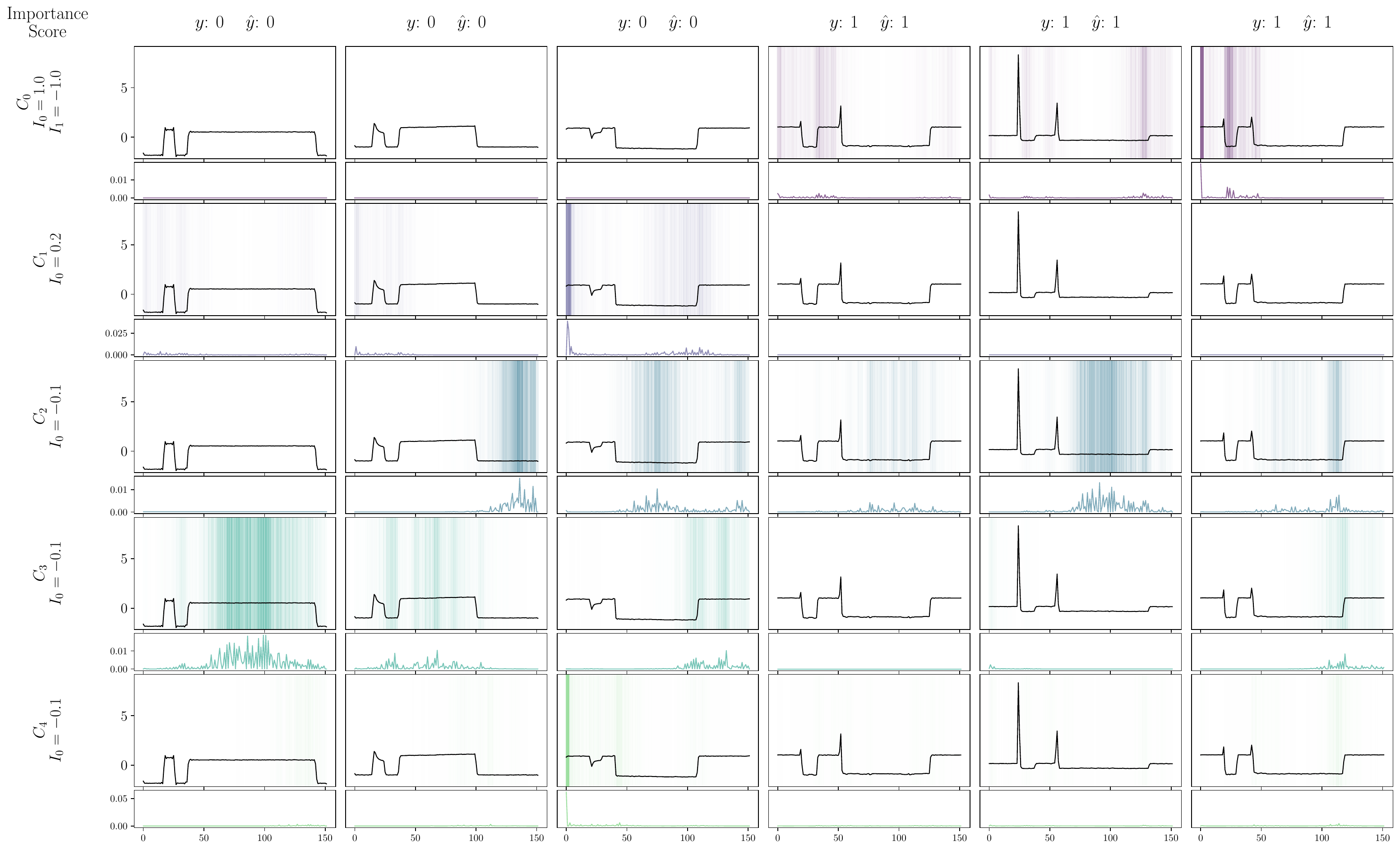}
        {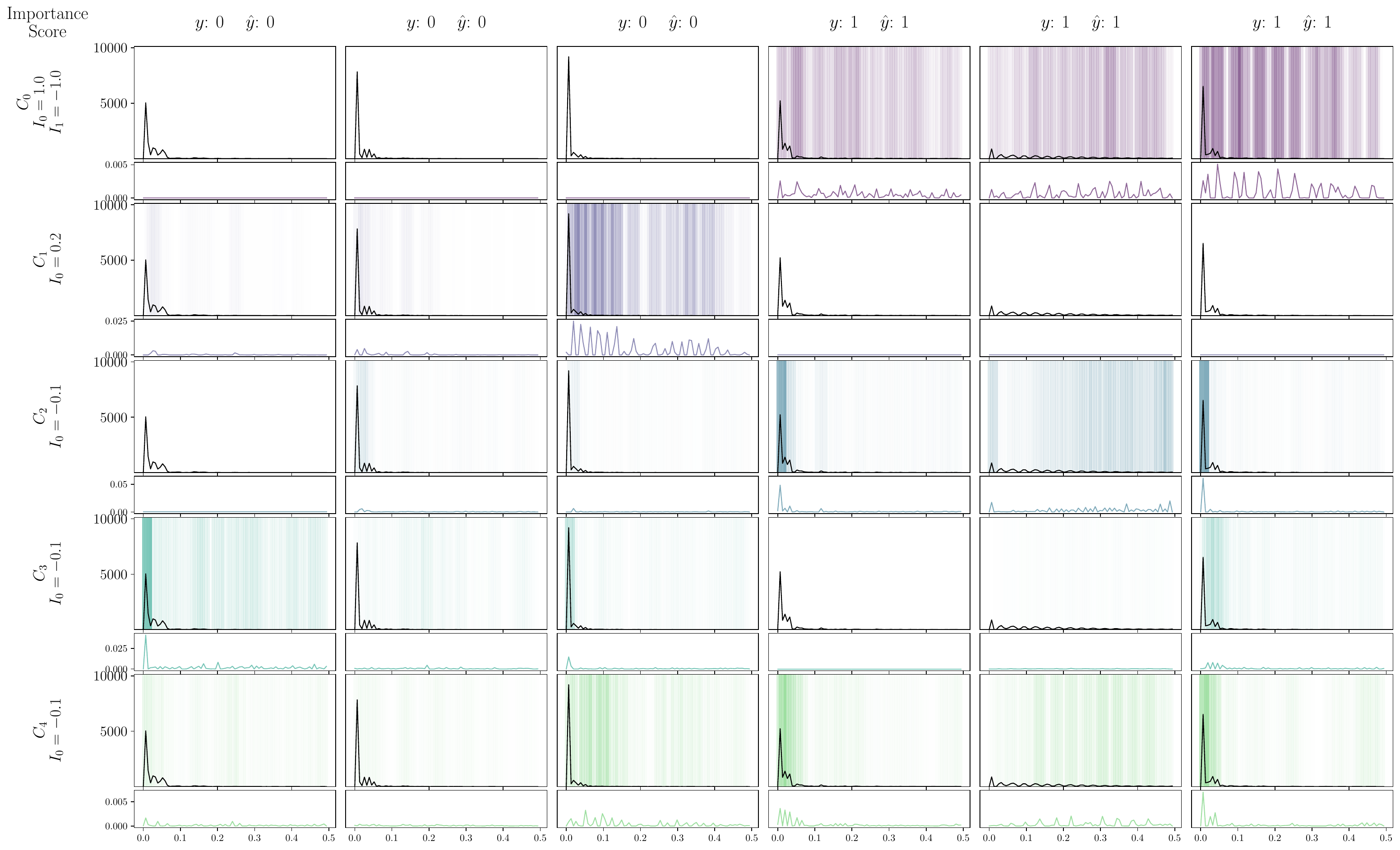}
        {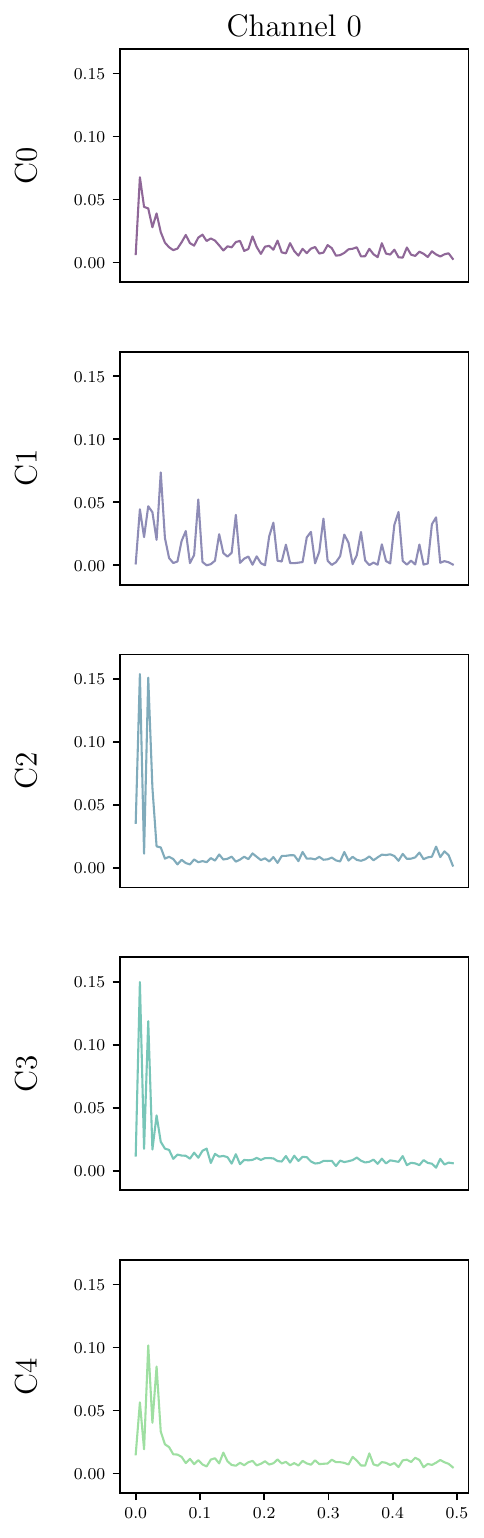}
    \caption{Concept extraction from a ResNet1D-18 trained on Wafer (two classes: normal, abnormal), reaching $100\%$ test accuracy.
    Top left: time-domain concept masks.
    Bottom left: frequency-domain concept masks.
    Right: global concept-frequency correspondence.
    Four concepts follow the stages of the process trace, while $C_0$ spans the whole window and carries the strongest class-discriminative importance.
    }
    \label{fig:ucr-wafer}
\end{figure}

\subsubsection{BearingPD}
\label{sec:app-bearingpd}
We complement the CWRU example in the main text with concept extraction on a second natural bearing-fault dataset.
\emph{Paderborn Bearing (BearingPD)}~\cite{lessmeier2016condition} contains accelerometer recordings from a motor test rig at Paderborn University, covering healthy bearings and bearings with artificially induced or naturally degraded inner and outer race damage; we use a three-class setup (healthy, inner-race fault, outer-race fault).
Recordings are sampled at $f_s = 64$\,kHz on a single vibration channel and segmented into non-overlapping windows of $T = 5000$ samples.
Training follows Section~\ref{sec:setup}, with two additions: 
Inputs are standardized by the global mean and standard deviation of the training set, and Gaussian noise of standard deviation $0.01$ is added to the standardized signal during training as augmentation.

\begin{figure}[p]
    \centering
    \begin{subfigure}[t]{\linewidth}
        \centering
        \reportpanels
            {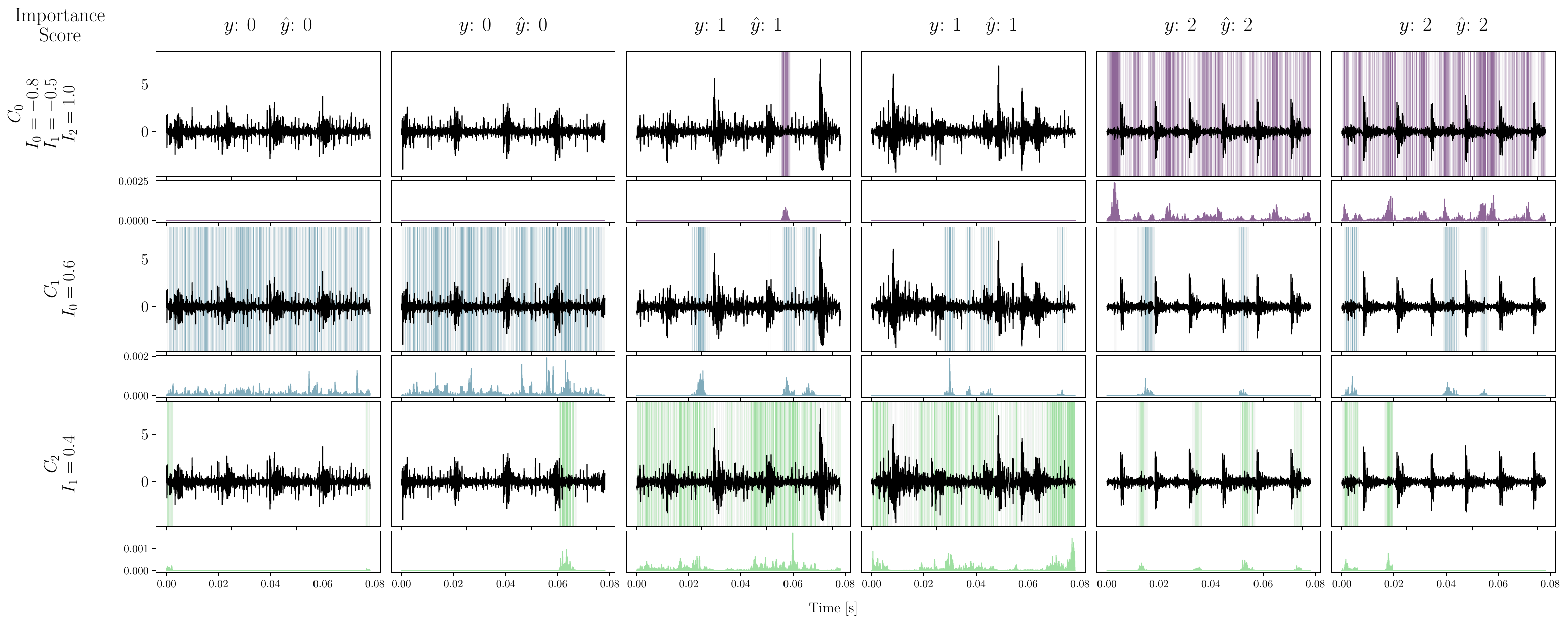}
            {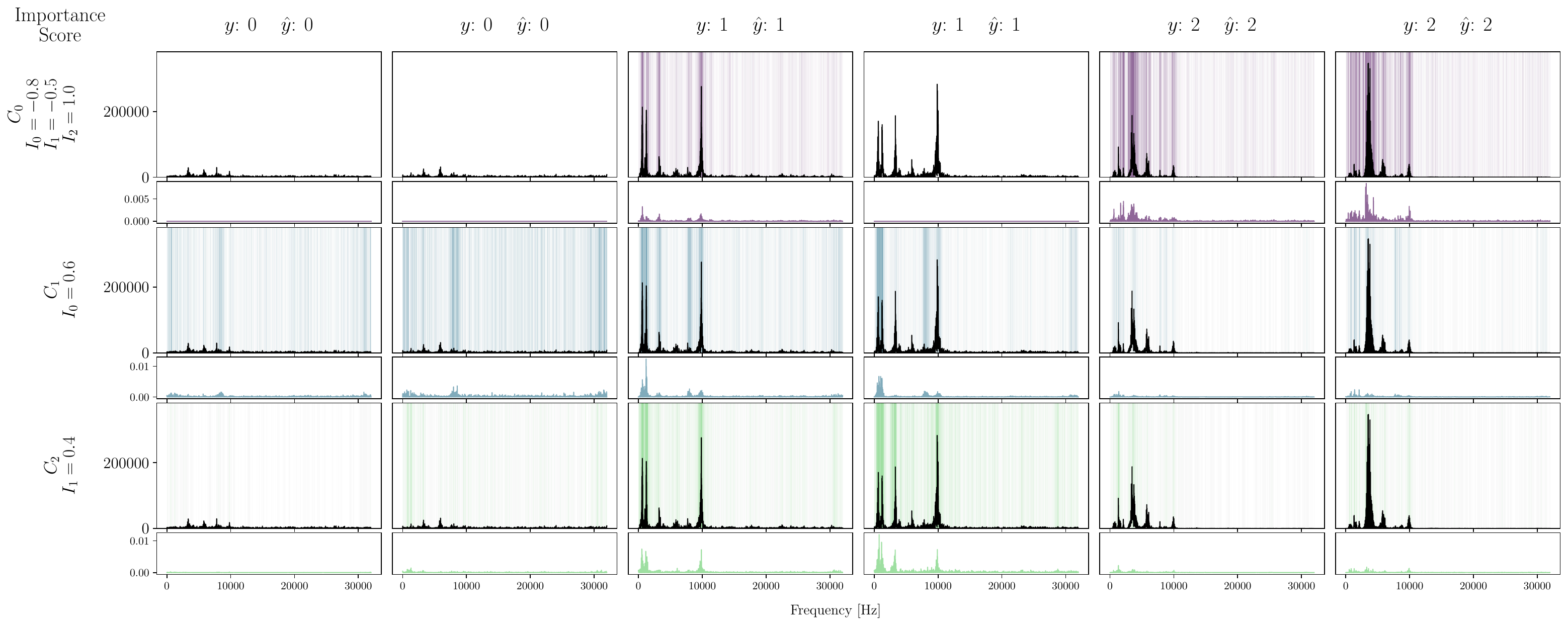}
            {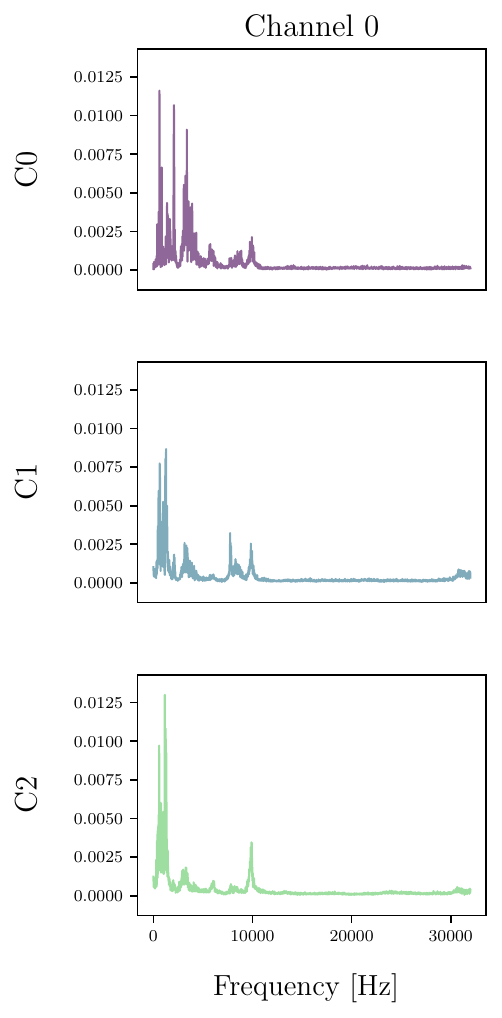}
        \caption{InceptionTime10, where $C_0$ carries the $2$ and $3.4$\,kHz bands alongside the low pair, while $C_1$ and $C_2$ concentrate on the low pair.}
        \label{fig:bearingPD}
    \end{subfigure}

    \vspace{1em}

    \begin{subfigure}[t]{\linewidth}
        \centering
        \reportpanels
            {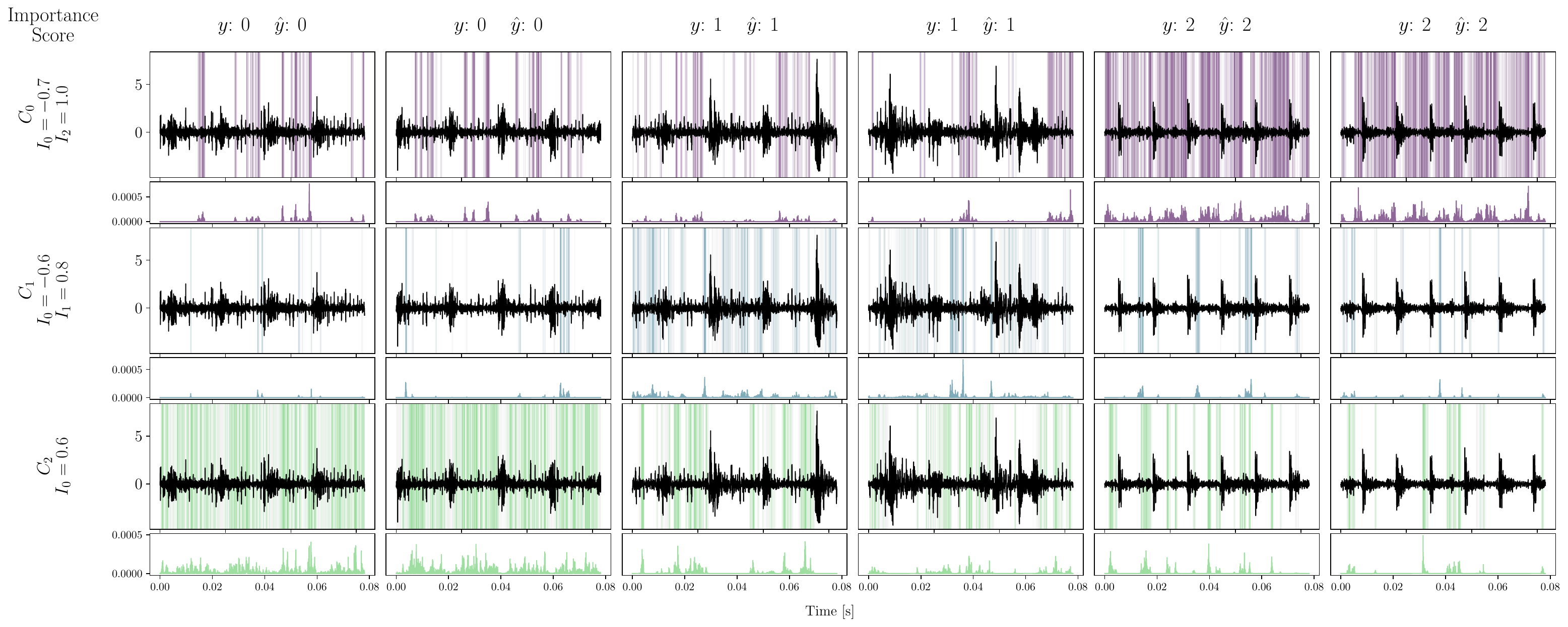}
            {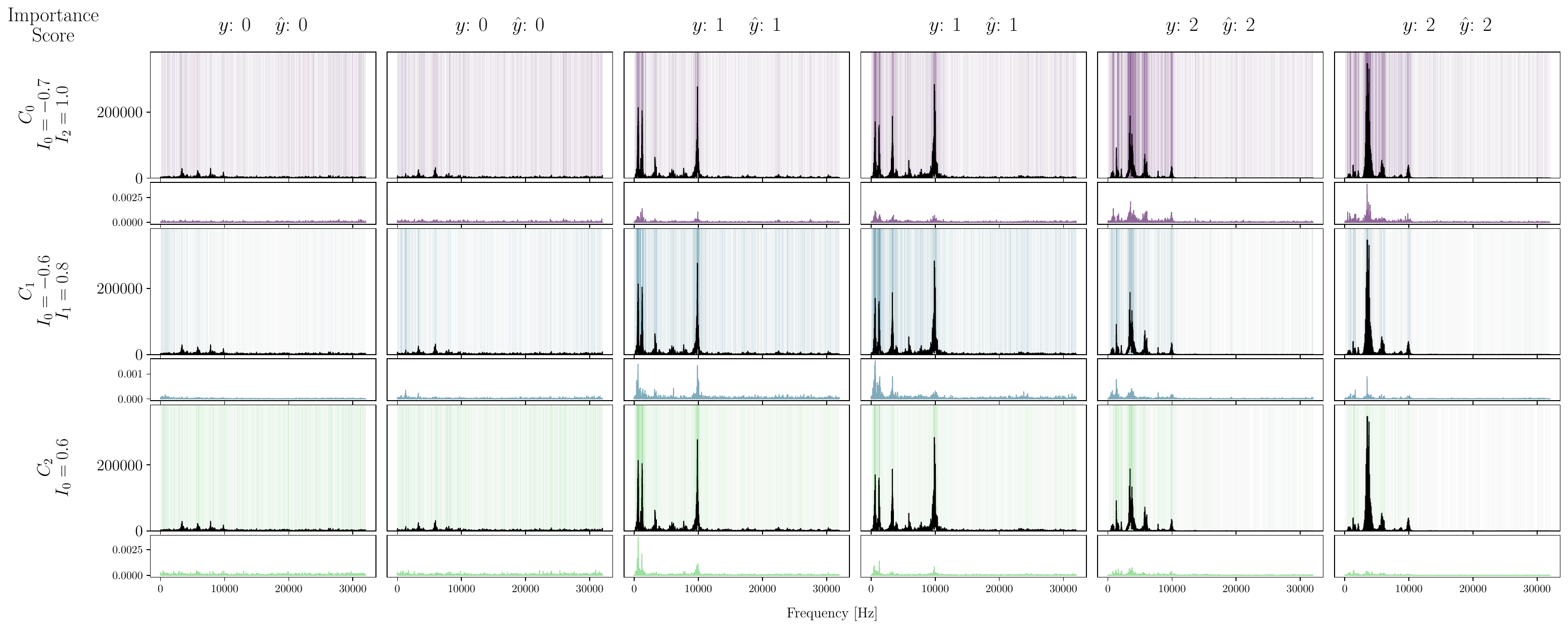}
            {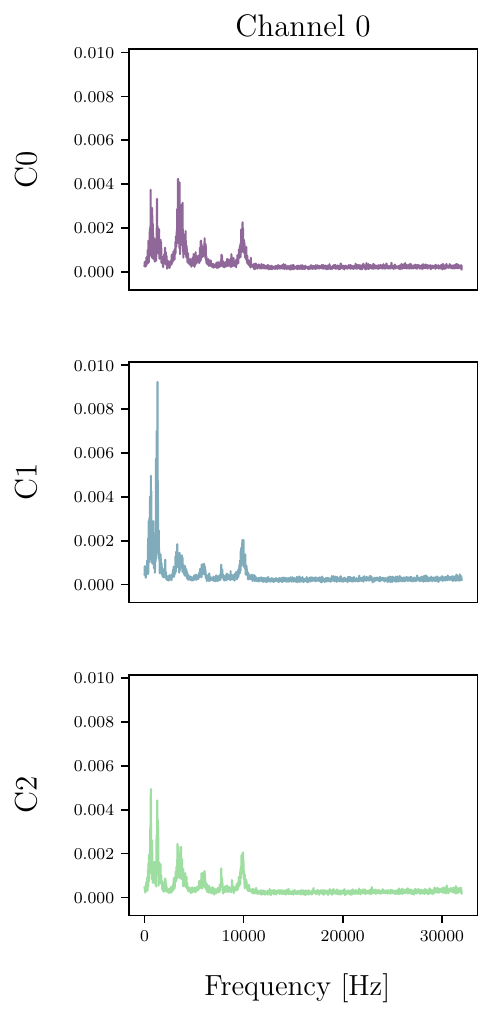}
        \caption{DenseNet1D-121, where $C_0$ peaks at $3.4$\,kHz, $C_1$ is dominated by a sharp $1.3$\,kHz peak, and $C_2$ carries the $0.6$--$1.3$\,kHz pair at comparable height.}
        \label{fig:bearingPDDensenet}
    \end{subfigure}
    \caption{
    Concept extraction on BearingPD (three classes: healthy, inner-race fault, outer-race fault) for two architectures, each trained with seed $0$ to $100\%$ validation accuracy.
    Top left of each panel: time-domain concept masks, which localize between high-amplitude transients and pick up patterns that align with the periodic impact structure of the signal.
    Bottom left: frequency-domain concept masks, whose mass concentrates on four bands at roughly $0.6$--$1.3$, $2$, $3.4$, and $10$\,kHz.
    Right: global concept-frequency correspondence per channel, where both architectures show a peak near $10$\,kHz.
    The two architectures recover the same four bands, all inside the structural-resonance region commonly used for envelope-based bearing diagnostics~\cite{randall2011rolling}, and differ only in how the bands distribute across concepts.
    }
    \label{fig:bearingPDboth}
\end{figure}

On BearingPD on the InceptionTime10 model (Figure~\ref{fig:bearingPD}), the extracted concepts partly separate by class, though some concepts activate in more than one class with differing activation strengths.
The frequency-domain masks and the global concept-correspondence concentrate concept mass on four distinguishable bands, with markedly different distributions across concepts:
$C_0$ spans all four, with its largest mass at $0.6$\,kHz and comparable mass in the $2$ and $3.4$\,kHz bands;
$C_1$ and $C_2$ concentrate on the low band, peaking at $1.3$\,kHz, and retain only weak activity at $2$ and $3.4$\,kHz;
all three carry a narrow peak near $10$\,kHz, and none shows a distinct peak above it.
All four bands fall inside the structural-resonance region typically used for envelope analysis on this kind of bearing rig~\cite{randall2011rolling}, and the per-concept differences within that region reflect that the model has partitioned the resonance content into class-specific sub-bands rather than treating it as a single feature.
These specific bands give domain experts the information needed to assess whether the model is reasoning over fault-relevant spectral content, and the agreement with the literature regions supports that CENDRe transfers from controlled synthetic settings to real signals.

On BearingPD on the DenseNet1D-121 model (Figure~\ref{fig:bearingPDDensenet}), CENDRe recovers the same four bands, distributed differently across the concepts.
The concepts again separate only partly by class, with some activating in more than one class at differing strengths.
$C_0$ places its largest mass at $3.4$\,kHz, with the low pair close behind and a clear peak near $10$\,kHz;
$C_1$ is dominated by a sharp peak at $1.3$\,kHz, roughly twice the height of any peak in the other two concepts;
$C_2$ carries the $0.6$ and $1.3$\,kHz peaks at comparable height, with weaker mass at $3.4$ and $10$\,kHz.
Both architectures resolve the low band into the same $0.6$ and $1.3$\,kHz pair, and in both, one concept is dominated by a single member of the pair while the others carry both at comparable height, though the concept indices playing each role differ.
Just as for InceptionTime10, these bands give domain experts the information needed to assess whether the model is reasoning over fault-relevant spectral content.
Recovering them under a second architecture supports that CENDRe transfers from controlled synthetic settings to real signals independently of the backbone.

\section{Hyperparameter sensitivity} 
\label{sec:app-sensitivity}
The CENDRe$_\mathrm{silhouette/HDBSCAN}$ variants replace the number of concepts $K$ with the number of micro-clusters $J$.
We investigate the sensitivity of CENDRe to $J$ by sweeping $J$ with every other hyperparameter fixed at its main-experiment value.
The grid is bounded by the method itself: the upper bound is the number of LADs in one mini-batch ($J_{\max}=3200$ in our configuration), since $k$-means cannot form more clusters than it sees in one partial fit, and the lower bound $J_{\min}=11$ follows from the agglomerative candidate search over up to $10$ concepts.
Within $[J_{\min}, J_{\max}]$ we sample $10$ log-spaced values, snapped to include the main-experiment value $J=50$, giving $J \in \{11, 21, 50, 73, 137, 257, 483, 907, 1704, 3200\}$.
The sweep covers two datasets from each of the main synthetic families and the $11$ seeds of the main experiments, reused at every level of $J$ so that seed acts as a block rather than an independent replicate.
Runs in which the extractor returned no concepts are excluded, which affects CENDRe$_{\mathrm{HDBSCAN}}$ at small $J$ only.

\begin{figure}
    \includegraphics[width=\linewidth]{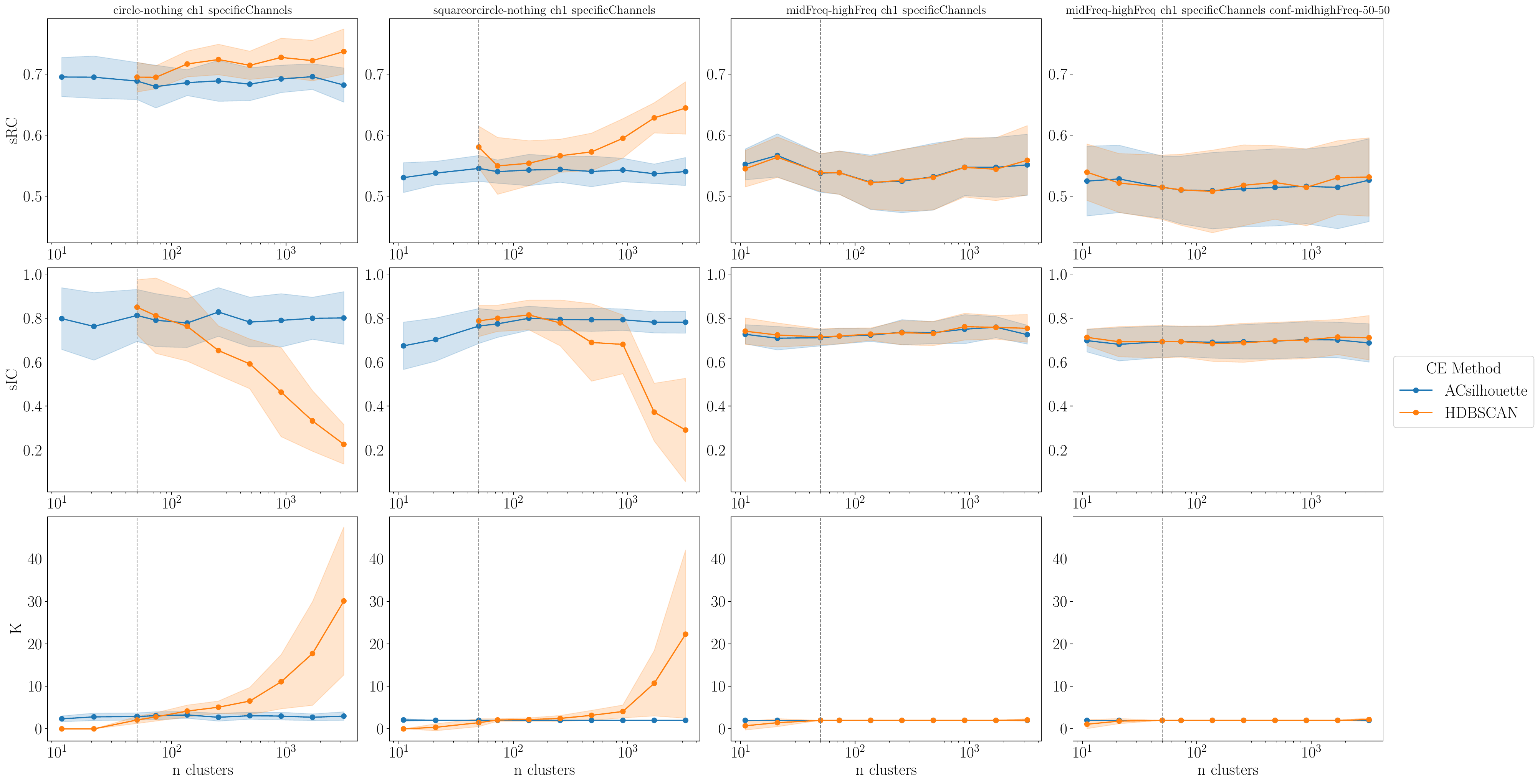}
    \caption{Response of sRC (top row), sIC (middle row) and $K$ (bottom row) to the number of micro-clusters $J$, per dataset, for CENDRe$_{\mathrm{silhouette}}$ and CENDRe$_{\mathrm{HDBSCAN}}$.
    Markers are the mean over $11$ seeds and shading is one standard deviation, with all panels on a common $y$-range.
    The dashed vertical line is the value $J=50$ used in the main experiments.}
    \label{fig:sensitivity0}
\end{figure}

Figure~\ref{fig:sensitivity0} shows the seed mean and standard deviation of sRC, sIC and $K$ per dataset.
CENDRe$_{\mathrm{silhouette}}$ is flat over three orders of magnitude of $J$ on all four datasets, except at the two smallest values, where too few micro-clusters remain for the candidate search.
CENDRe$_{\mathrm{HDBSCAN}}$ matches it on \textsc{syntheticFrequency}, but on \textsc{syntheticLocal} its sIC collapses above a few hundred micro-clusters, with a widening seed spread.

We quantify the sensitivity with two standard measures, both taken under a log-uniform distribution of $J$ on $[J_{\min}, J_{\max}]$, which the log-spaced grid samples uniformly~\cite{bergstra2012random}.
Writing the score as $y(J, \varepsilon)$, where $\varepsilon$ collects everything the seed controls, the first-order Sobol index~\cite{sobol2001global,saltelli2008global}
\begin{equation}
    S_1^J = \frac{\mathrm{Var}_J\!\left(\mathbb{E}_\varepsilon\!\left[y \mid J\right]\right)}{\mathrm{Var}(y)}
    \label{eq:sobol}
\end{equation}
is the share of the total variance explained by the choice of $J$, and $S_1^{\mathrm{seed}}$ exchanges the roles of $J$ and $\varepsilon$.
We estimate Equation~\ref{eq:sobol} from the paired design as a two-way variance decomposition over levels of $J$ and seeds, in the bias-corrected $\omega^2$ form~\cite{olejnik2003generalized}, weighting the levels by their log spacing.
A variance share carries no direction, so we add the elementary effects of Morris~\cite{morris1992factorial}.
Taking the factor as $\log_2 J$, which the geometric grid samples at near-uniform steps, the effect of the step from $J_j$ to $J_{j+1}$ under seed $s$ is
\begin{equation}
    EE_{s,j} = \frac{y_{s,j+1} - y_{s,j}}{\log_2 J_{j+1} - \log_2 J_j}.
    \label{eq:ee}
\end{equation}
Their mean $\mu_{EE}$ reads as the average change in the score per doubling of $J$, in the units of that score.
We report the signed mean rather than the absolute-value variant $\mu^*$~\cite{campolongo2007effective}, since with a single swept factor the sign carries the direction of the drift.

Tables~\ref{tab:sensitivitysRCrma}, \ref{tab:sensitivitysICrma}, and \ref{tab:sensitivityK} report the three measures.
For CENDRe$_{\mathrm{silhouette}}$, $J$ explains on average $0.022$ of the variance in sRC, $0.072$ in sIC, and $0.010$ in the selected $K$ itself, against $0.702$, $0.577$, and $0.159$ for the seed.
The mean elementary effect on $K$ is a negligible $+0.02$ concepts per doubling of $J$, consistent with the near-zero effect already observed on sRC and sIC.
Choosing $J$ is therefore a smaller decision than the run-to-run variation the method already tolerates, for the score and for $K$ alike.
CENDRe$_{\mathrm{HDBSCAN}}$ is more sensitive on average in sRC and sIC ($S_1^J = 0.144$ and $0.320$).
Its selected $K$ tracks $J$ far more directly, with $S_1^J = 0.416$ on average and a mean elementary effect reaching $+3.67$ concepts per doubling of $J$ on \textsc{circle-nothing}.
This uncontrolled growth in $K$ plausibly explains the sIC collapse noted above.
Sensitivity to $J$, in the score and in the selected $K$, is therefore a property of the density-based cluster mode rather than of CENDRe itself.
The one exception is sIC on \texttt{squareorcircle-nothing}, where the silhouette variant also reaches $S_1^J = 0.23$.

\begin{table*}[!hbtp]
\centering\scriptsize
\setlength{\tabcolsep}{3pt}
\caption{Sensitivity to the number of micro-clusters $J$, for sRC, sIC, and $K$. The columns $S_1^J$ and $S_1^{\mathrm{seed}}$ are the first-order Sobol indices of Equation~\ref{eq:sobol} for $J$ and for the random seed, and $\mu_{EE}$ is the mean elementary effect of Equation~\ref{eq:ee}, in units of the reported score per doubling of $J$. The final row of each panel averages each index over datasets.}
\label{tab:sensitivitygeneral}

\begin{subtable}{\textwidth}
\centering
\caption{Sensitivity of sRC to $J$.}
\label{tab:sensitivitysRCrma}
\resizebox{\textwidth}{!}{\input{tables/table_sensitivity_nclusters_InceptionTime10_sRC_rma}}
\end{subtable}

\begin{subtable}{\textwidth}
\centering
\caption{Sensitivity of sIC to $J$.}
\label{tab:sensitivitysICrma}
\resizebox{\textwidth}{!}{\input{tables/table_sensitivity_nclusters_InceptionTime10_sIC_rma}}
\end{subtable}

\begin{subtable}{\textwidth}
\centering
\caption{Sensitivity of K to $J$.}
\label{tab:sensitivityK}
\resizebox{\textwidth}{!}{\input{tables/table_sensitivity_nclusters_InceptionTime10_K}}
\end{subtable}
\end{table*}

%% file: tables/table_distributional_summary_rma_sRC.tex
\scriptsize
\setlength{\tabcolsep}{3pt}
\centering
\resizebox{\textwidth}{!}{%
\begin{tabular}{@{}p{3.2cm}|p{2.0cm}p{2.0cm}p{2.0cm}p{2.0cm}p{2.0cm}p{2.0cm}p{2.0cm}p{2.0cm}p{2.0cm}p{2.0cm}@{}}
\toprule
 & \multicolumn{2}{c}{CENDRe$_{\text{silhouette}}$} & \multicolumn{2}{c}{CENDRe$_{\text{HDBSCAN}}$} & \multicolumn{2}{c}{CENDRe$_{k\text{Means}}$} & \multicolumn{2}{c}{ECLAD-ts} & \multicolumn{2}{c}{MultiVISION} \\
\cmidrule(lr){2-3}\cmidrule(lr){4-5}\cmidrule(lr){6-7}\cmidrule(lr){8-9}\cmidrule(lr){10-11}
 & mean $\pm$ std & 95\% CI & mean $\pm$ std & 95\% CI & mean $\pm$ std & 95\% CI & mean $\pm$ std & 95\% CI & mean $\pm$ std & 95\% CI \\
\midrule
\multicolumn{11}{@{}l}{\textbf{\textsc{syntheticLocal}}} \\
\texttt{circle-nothing} & 0.731 $\pm$ 0.035 & [0.718, 0.743] & 0.716 $\pm$ 0.030 & [0.705, 0.726] & 0.774 $\pm$ 0.049 & [0.756, 0.791] & \textbf{0.824 $\pm$ 0.060} & \textbf{[0.803, 0.846]} & 0.584 $\pm$ 0.070 & [0.559, 0.609] \\
\texttt{square-circle} & 0.599 $\pm$ 0.053 & [0.581, 0.618] & 0.561 $\pm$ 0.035 & [0.549, 0.574] & 0.662 $\pm$ 0.066 & [0.639, 0.686] & \textbf{0.723 $\pm$ 0.094} & \textbf{[0.690, 0.756]} & 0.450 $\pm$ 0.052 & [0.431, 0.468] \\
\texttt{triangle-circle} & 0.622 $\pm$ 0.073 & [0.596, 0.648] & 0.575 $\pm$ 0.061 & [0.553, 0.597] & 0.672 $\pm$ 0.087 & [0.642, 0.703] & \textbf{0.689 $\pm$ 0.099} & \textbf{[0.654, 0.724]} & 0.443 $\pm$ 0.057 & [0.423, 0.463] \\
\texttt{square-nothing} & 0.710 $\pm$ 0.028 & [0.701, 0.720] & 0.711 $\pm$ 0.034 & [0.699, 0.724] & 0.757 $\pm$ 0.046 & [0.741, 0.774] & \textbf{0.813 $\pm$ 0.057} & \textbf{[0.792, 0.833]} & 0.600 $\pm$ 0.045 & [0.584, 0.616] \\
\texttt{square-triangle} & 0.597 $\pm$ 0.066 & [0.573, 0.620] & 0.563 $\pm$ 0.049 & [0.545, 0.580] & 0.639 $\pm$ 0.062 & [0.617, 0.661] & \textbf{0.674 $\pm$ 0.087} & \textbf{[0.643, 0.704]} & 0.453 $\pm$ 0.039 & [0.439, 0.467] \\
\texttt{triangle-nothing} & 0.742 $\pm$ 0.054 & [0.723, 0.762] & 0.729 $\pm$ 0.053 & [0.710, 0.747] & 0.804 $\pm$ 0.059 & [0.783, 0.825] & \textbf{0.838 $\pm$ 0.061} & \textbf{[0.817, 0.860]} & 0.609 $\pm$ 0.031 & [0.598, 0.620] \\
\texttt{squareorcircle-nothing} & 0.617 $\pm$ 0.054 & [0.598, 0.637] & 0.620 $\pm$ 0.035 & [0.607, 0.633] & 0.669 $\pm$ 0.064 & [0.646, 0.691] & \textbf{0.705 $\pm$ 0.046} & \textbf{[0.688, 0.721]} & 0.526 $\pm$ 0.042 & [0.511, 0.541] \\
\texttt{triangleorcircle-nothing} & 0.623 $\pm$ 0.063 & [0.601, 0.646] & 0.618 $\pm$ 0.069 & [0.592, 0.643] & 0.677 $\pm$ 0.067 & [0.653, 0.700] & \textbf{0.701 $\pm$ 0.040} & \textbf{[0.687, 0.715]} & 0.525 $\pm$ 0.040 & [0.511, 0.540] \\
\texttt{squareortriangle-nothing} & 0.618 $\pm$ 0.060 & [0.597, 0.640] & 0.600 $\pm$ 0.073 & [0.573, 0.627] & 0.684 $\pm$ 0.073 & [0.658, 0.710] & \textbf{0.703 $\pm$ 0.040} & \textbf{[0.689, 0.717]} & 0.533 $\pm$ 0.035 & [0.521, 0.546] \\
\midrule
\textit{Mean} & 0.651 $\pm$ 0.078 & [0.642, 0.660] & 0.633 $\pm$ 0.082 & [0.623, 0.642] & 0.704 $\pm$ 0.084 & [0.695, 0.714] & \textbf{0.741 $\pm$ 0.091} & \textbf{[0.731, 0.751]} & 0.525 $\pm$ 0.077 & [0.516, 0.534] \\
\midrule
\multicolumn{11}{@{}l}{\textbf{\textsc{syntheticFrequency}}} \\
\texttt{midFreq-highFreq} & 0.503 $\pm$ 0.042 & [0.488, 0.518] & 0.504 $\pm$ 0.042 & [0.489, 0.519] & \textbf{0.593 $\pm$ 0.058} & \textbf{[0.573, 0.614]} & -- & -- & -- & -- \\
\texttt{midFreq-highFreq\newline \_conf-midhighFreq-50-50} & 0.490 $\pm$ 0.050 & [0.472, 0.508] & 0.490 $\pm$ 0.049 & [0.473, 0.508] & \textbf{0.554 $\pm$ 0.054} & \textbf{[0.535, 0.573]} & -- & -- & -- & -- \\
\texttt{midFreq-midhighFreq} & 0.532 $\pm$ 0.034 & [0.520, 0.544] & 0.532 $\pm$ 0.034 & [0.520, 0.544] & \textbf{0.585 $\pm$ 0.052} & \textbf{[0.567, 0.603]} & -- & -- & -- & -- \\
\texttt{midFreq-midhighFreq\newline \_conf-highFreq-50-50} & 0.503 $\pm$ 0.036 & [0.490, 0.516] & 0.502 $\pm$ 0.038 & [0.489, 0.516] & \textbf{0.570 $\pm$ 0.047} & \textbf{[0.553, 0.586]} & -- & -- & -- & -- \\
\texttt{midhighFreq-highFreq} & 0.546 $\pm$ 0.037 & [0.533, 0.559] & 0.546 $\pm$ 0.036 & [0.533, 0.559] & \textbf{0.583 $\pm$ 0.047} & \textbf{[0.566, 0.599]} & -- & -- & -- & -- \\
\texttt{midhighFreq-highFreq\newline \_conf-midFreq-50-50} & 0.488 $\pm$ 0.031 & [0.477, 0.499] & 0.489 $\pm$ 0.030 & [0.478, 0.500] & \textbf{0.540 $\pm$ 0.049} & \textbf{[0.523, 0.558]} & -- & -- & -- & -- \\
\midrule
\textit{Mean} & 0.510 $\pm$ 0.044 & [0.504, 0.517] & 0.511 $\pm$ 0.044 & [0.504, 0.517] & \textbf{0.571 $\pm$ 0.054} & \textbf{[0.563, 0.578]} & -- & -- & -- & -- \\
\bottomrule
\end{tabular}
}

%% file: tables/table_distributional_summary_rma_sIC.tex
\scriptsize
\setlength{\tabcolsep}{3pt}
\centering
\resizebox{\textwidth}{!}{%
\begin{tabular}{@{}p{3.2cm}|p{2.0cm}p{2.0cm}p{2.0cm}p{2.0cm}p{2.0cm}p{2.0cm}p{2.0cm}p{2.0cm}p{2.0cm}p{2.0cm}@{}}
\toprule
 & \multicolumn{2}{c}{CENDRe$_{\text{silhouette}}$} & \multicolumn{2}{c}{CENDRe$_{\text{HDBSCAN}}$} & \multicolumn{2}{c}{CENDRe$_{k\text{Means}}$} & \multicolumn{2}{c}{ECLAD-ts} & \multicolumn{2}{c}{MultiVISION} \\
\cmidrule(lr){2-3}\cmidrule(lr){4-5}\cmidrule(lr){6-7}\cmidrule(lr){8-9}\cmidrule(lr){10-11}
 & mean $\pm$ std & 95\% CI & mean $\pm$ std & 95\% CI & mean $\pm$ std & 95\% CI & mean $\pm$ std & 95\% CI & mean $\pm$ std & 95\% CI \\
\midrule
\multicolumn{11}{@{}l}{\textbf{\textsc{syntheticLocal}}} \\
\texttt{circle-nothing} & 0.806 $\pm$ 0.091 & [0.774, 0.839] & \textbf{0.850 $\pm$ 0.100} & \textbf{[0.814, 0.886]} & 0.666 $\pm$ 0.140 & [0.616, 0.716] & 0.684 $\pm$ 0.283 & [0.584, 0.784] & 0.686 $\pm$ 0.137 & [0.637, 0.734] \\
\texttt{square-circle} & 0.721 $\pm$ 0.127 & [0.676, 0.766] & \textbf{0.816 $\pm$ 0.140} & \textbf{[0.767, 0.866]} & 0.605 $\pm$ 0.098 & [0.570, 0.639] & 0.550 $\pm$ 0.250 & [0.461, 0.638] & 0.735 $\pm$ 0.120 & [0.692, 0.777] \\
\texttt{triangle-circle} & 0.794 $\pm$ 0.123 & [0.751, 0.838] & \textbf{0.849 $\pm$ 0.121} & \textbf{[0.806, 0.893]} & 0.653 $\pm$ 0.145 & [0.602, 0.705] & 0.474 $\pm$ 0.298 & [0.369, 0.580] & 0.743 $\pm$ 0.135 & [0.695, 0.791] \\
\texttt{square-nothing} & 0.799 $\pm$ 0.103 & [0.763, 0.836] & \textbf{0.873 $\pm$ 0.090} & \textbf{[0.840, 0.905]} & 0.668 $\pm$ 0.135 & [0.620, 0.716] & 0.466 $\pm$ 0.234 & [0.383, 0.549] & 0.711 $\pm$ 0.078 & [0.683, 0.739] \\
\texttt{square-triangle} & 0.766 $\pm$ 0.097 & [0.731, 0.800] & \textbf{0.842 $\pm$ 0.111} & \textbf{[0.803, 0.881]} & 0.626 $\pm$ 0.110 & [0.587, 0.665] & 0.596 $\pm$ 0.228 & [0.515, 0.677] & 0.731 $\pm$ 0.104 & [0.694, 0.768] \\
\texttt{triangle-nothing} & 0.784 $\pm$ 0.125 & [0.740, 0.828] & \textbf{0.873 $\pm$ 0.115} & \textbf{[0.832, 0.914]} & 0.661 $\pm$ 0.174 & [0.599, 0.723] & 0.599 $\pm$ 0.287 & [0.497, 0.701] & 0.741 $\pm$ 0.083 & [0.712, 0.771] \\
\texttt{squareorcircle-nothing} & 0.819 $\pm$ 0.098 & [0.784, 0.854] & \textbf{0.870 $\pm$ 0.094} & \textbf{[0.835, 0.905]} & 0.712 $\pm$ 0.128 & [0.667, 0.758] & 0.551 $\pm$ 0.193 & [0.483, 0.619] & 0.707 $\pm$ 0.100 & [0.672, 0.743] \\
\texttt{triangleorcircle-nothing} & 0.812 $\pm$ 0.098 & [0.778, 0.847] & \textbf{0.884 $\pm$ 0.084} & \textbf{[0.853, 0.915]} & 0.676 $\pm$ 0.123 & [0.633, 0.720] & 0.558 $\pm$ 0.305 & [0.450, 0.666] & 0.739 $\pm$ 0.065 & [0.717, 0.762] \\
\texttt{squareortriangle-nothing} & 0.827 $\pm$ 0.097 & [0.792, 0.861] & \textbf{0.869 $\pm$ 0.095} & \textbf{[0.834, 0.904]} & 0.651 $\pm$ 0.128 & [0.606, 0.697] & 0.523 $\pm$ 0.217 & [0.446, 0.600] & 0.706 $\pm$ 0.092 & [0.673, 0.738] \\
\midrule
\textit{Mean} & 0.792 $\pm$ 0.110 & [0.779, 0.805] & \textbf{0.858 $\pm$ 0.107} & \textbf{[0.846, 0.871]} & 0.658 $\pm$ 0.134 & [0.642, 0.673] & 0.556 $\pm$ 0.262 & [0.526, 0.586] & 0.722 $\pm$ 0.105 & [0.710, 0.734] \\
\midrule
\multicolumn{11}{@{}l}{\textbf{\textsc{syntheticFrequency}}} \\
\texttt{midFreq-highFreq} & 0.802 $\pm$ 0.096 & [0.768, 0.836] & \textbf{0.804 $\pm$ 0.096} & \textbf{[0.770, 0.837]} & 0.694 $\pm$ 0.127 & [0.649, 0.739] & -- & -- & -- & -- \\
\texttt{midFreq-highFreq\newline \_conf-midhighFreq-50-50} & \textbf{0.776 $\pm$ 0.121} & \textbf{[0.733, 0.819]} & 0.775 $\pm$ 0.123 & [0.731, 0.818] & 0.690 $\pm$ 0.114 & [0.650, 0.731] & -- & -- & -- & -- \\
\texttt{midFreq-midhighFreq} & \textbf{0.820 $\pm$ 0.111} & \textbf{[0.781, 0.860]} & 0.819 $\pm$ 0.111 & [0.780, 0.859] & 0.744 $\pm$ 0.161 & [0.687, 0.801] & -- & -- & -- & -- \\
\texttt{midFreq-midhighFreq\newline \_conf-highFreq-50-50} & \textbf{0.792 $\pm$ 0.109} & \textbf{[0.754, 0.831]} & 0.790 $\pm$ 0.111 & [0.751, 0.830] & 0.690 $\pm$ 0.108 & [0.651, 0.728] & -- & -- & -- & -- \\
\texttt{midhighFreq-highFreq} & \textbf{0.791 $\pm$ 0.103} & \textbf{[0.754, 0.828]} & 0.789 $\pm$ 0.103 & [0.752, 0.825] & 0.709 $\pm$ 0.144 & [0.658, 0.761] & -- & -- & -- & -- \\
\texttt{midhighFreq-highFreq\newline \_conf-midFreq-50-50} & 0.770 $\pm$ 0.128 & [0.724, 0.815] & \textbf{0.772 $\pm$ 0.127} & \textbf{[0.726, 0.817]} & 0.679 $\pm$ 0.116 & [0.638, 0.720] & -- & -- & -- & -- \\
\midrule
\textit{Mean} & \textbf{0.792 $\pm$ 0.112} & \textbf{[0.776, 0.807]} & 0.791 $\pm$ 0.112 & [0.776, 0.807] & 0.701 $\pm$ 0.130 & [0.683, 0.719] & -- & -- & -- & -- \\
\bottomrule
\end{tabular}
}

%% file: tables/table_sensitivity_nclusters_InceptionTime10_sRC_rma.tex
\scriptsize
\setlength{\tabcolsep}{3pt}
\centering
\resizebox{\linewidth}{!}{%
\begin{tabular}{@{}p{4.4cm}|p{1.55cm}p{1.55cm}p{1.55cm}p{1.55cm}p{1.55cm}p{1.55cm}@{}}
\toprule
 & \multicolumn{3}{c}{CENDRe$_{\mathrm{silhouette}}$} & \multicolumn{3}{c}{CENDRe$_{\mathrm{HDBSCAN}}$} \\
\cmidrule(lr){2-4} \cmidrule(lr){5-7}
Dataset & $S_1^J$ & $S_1^{\mathrm{seed}}$ & $\mu_{EE}$ & $S_1^J$ & $S_1^{\mathrm{seed}}$ & $\mu_{EE}$ \\
\midrule
\texttt{circle-nothing} & 0.009 & 0.666 & -0.0021 & 0.081 & 0.383 & +0.0078 \\
\texttt{squareorcircle-nothing} & 0.008 & 0.696 & +0.0005 & 0.448 & 0.045 & +0.0174 \\
\texttt{midFreq-highFreq} & 0.069 & 0.591 & +0.0009 & 0.036 & 0.746 & +0.0032 \\
\texttt{midFreq-highFreq\newline \_conf-midhighFreq-50-50} & 0.000 & 0.854 & +0.0003 & 0.012 & 0.910 & +0.0022 \\
\midrule
Mean & 0.022 & 0.702 & -0.0001 & 0.144 & 0.521 & +0.0076 \\
\bottomrule
\end{tabular}
}

%% file: tables/table_sensitivity_nclusters_InceptionTime10_sIC_rma.tex
\scriptsize
\setlength{\tabcolsep}{3pt}
\centering
\resizebox{\linewidth}{!}{%
\begin{tabular}{@{}p{4.4cm}|p{1.55cm}p{1.55cm}p{1.55cm}p{1.55cm}p{1.55cm}p{1.55cm}@{}}
\toprule
 & \multicolumn{3}{c}{CENDRe$_{\mathrm{silhouette}}$} & \multicolumn{3}{c}{CENDRe$_{\mathrm{HDBSCAN}}$} \\
\cmidrule(lr){2-4} \cmidrule(lr){5-7}
Dataset & $S_1^J$ & $S_1^{\mathrm{seed}}$ & $\mu_{EE}$ & $S_1^J$ & $S_1^{\mathrm{seed}}$ & $\mu_{EE}$ \\
\midrule
\texttt{circle-nothing} & 0.000 & 0.633 & -0.0029 & 0.599 & 0.047 & -0.1073 \\
\texttt{squareorcircle-nothing} & 0.233 & 0.479 & +0.0118 & 0.640 & 0.094 & -0.0932 \\
\texttt{midFreq-highFreq} & 0.056 & 0.290 & +0.0003 & 0.041 & 0.207 & +0.0066 \\
\texttt{midFreq-highFreq\newline \_conf-midhighFreq-50-50} & 0.000 & 0.907 & -0.0016 & 0.000 & 0.729 & +0.0029 \\
\midrule
Mean & 0.072 & 0.577 & +0.0019 & 0.320 & 0.269 & -0.0478 \\
\bottomrule
\end{tabular}
}

%% file: tables/table_sensitivity_nclusters_InceptionTime10_K.tex
\scriptsize
\setlength{\tabcolsep}{3pt}
\centering
\resizebox{\linewidth}{!}{%
\begin{tabular}{@{}p{4.4cm}|p{1.55cm}p{1.55cm}p{1.55cm}p{1.55cm}p{1.55cm}p{1.55cm}@{}}
\toprule
 & \multicolumn{3}{c}{CENDRe$_{\mathrm{silhouette}}$} & \multicolumn{3}{c}{CENDRe$_{\mathrm{HDBSCAN}}$} \\
\cmidrule(lr){2-4} \cmidrule(lr){5-7}
Dataset & $S_1^J$ & $S_1^{\mathrm{seed}}$ & $\mu_{EE}$ & $S_1^J$ & $S_1^{\mathrm{seed}}$ & $\mu_{EE}$ \\
\midrule
\texttt{circle-nothing} & 0.041 & 0.635 & +0.0880 & 0.589 & 0.041 & +3.6664 \\
\texttt{squareorcircle-nothing} & 0.000 & 0.000 & -0.0108 & 0.470 & 0.001 & +2.7356 \\
\texttt{midFreq-highFreq} & 0.000 & 0.000 & +0.0000 & 0.356 & 0.026 & +0.1462 \\
\texttt{midFreq-highFreq\newline \_conf-midhighFreq-50-50} & 0.000 & 0.000 & +0.0000 & 0.248 & 0.010 & +0.1250 \\
\midrule
Mean & 0.010 & 0.159 & +0.0193 & 0.416 & 0.020 & +1.6683 \\
\bottomrule
\end{tabular}
}

%% file: text/94_appendix_compute.tex
\section{Computational resources}
\label{app:compute}
The experiments span two systems, an institutional HPC cluster and a local workstation.
Each HPC GPU node provides two Intel Xeon CPUs ($24$ physical cores each, $192$\,GB RAM), four NVIDIA A100 (40\,GB) GPUs, and an InfiniBand interconnect.
The workstation has dual Intel Xeon Gold 6330 CPUs (2$\times$28 cores, $112$ logical threads at $2.00$\,GHz, $1$\,TiB RAM) and two NVIDIA A100-PCIE-40GB GPUs.
Model training ran entirely on the cluster.
The CE and synthetic-metric experiments ran on both systems, so their runtimes are reported as wall-clock measurements taken on the workstation under controlled conditions and aggregated over the full sweep.

\subsection{Model training}
\label{app:compute-training}
Model training was performed on $3$ architectures over $27$ dataset configurations, with $11$ seeds on the $15$ \textsc{syntheticLocal} and \textsc{syntheticFrequency} datasets and $3$ seeds on the $12$ \textsc{syntheticLmc} and \textsc{syntheticLconf} datasets, totaling $603$ trained models.
Training consumed a total of $40.64$ GPU hours. 

\subsection{Concept extraction and synthetic metrics}
\label{app:compute-ce}
Runtimes below are wall-clock measurements collected on the workstation from controlled batches on a single partition with $25$ allocated CPUs and no concurrent GPU users.
Both CE and synthetic-metric runtimes are reported per (architecture, method) as the mean over $3$ seeds on the \textsc{syntheticLocal} square-triangle variant, summed over the $K$-sweep used in the comparison.
For the AC-clustered CENDRe variants (silhouette, HDBSCAN), $K$ is determined in a single pass, so a single run replaces the sweep, and the tables reflect this (they are $6$--$9\times$ faster than the fixed-$K$ baselines).
The experiments on ECLAD-ts and MultiVISION were done on the \textsc{syntheticLocal} family ($9$ datasets, $11$ seeds) and the \textsc{syntheticLmc} and \textsc{syntheticLconf} families ($12$ datasets, $3$ seeds), which correspond to $9 \times 11 + 12 \times 3 = 135$ runs per architecture.
The CENDRe variants additionally run on the \textsc{syntheticFrequency} family ($6$ datasets, $11$ seeds), making for a total of $135 + 66 = 201$ runs per architecture.

\paragraph{Concept extraction.}
\begin{table}[h]
\centering
\caption{Mean CE runtime per (architecture, method), in seconds, over $3$ seeds on a representative \textsc{syntheticLocal} variant. Fixed-$K$ methods (ECLAD-ts, MultiVISION, CENDRe$_{k\text{Means}}$) sum over $K\in\{2,4,6,8,10\}$; AC variants run once.}
\label{tab:app_ce_time}
\small
\begin{tabular}{@{}lrrrrr@{}}
\toprule
Architecture & ECLAD-ts & MultiVISION & CENDRe$_{k\text{Means}}$ & CENDRe$_{\text{silhouette}}$ & CENDRe$_{\text{HDBSCAN}}$ \\
\midrule
InceptionTime10  & $102$  & $120$ & $118$ & $18$ & $15$ \\
ResNet1D-18      & $\phantom{0}88$  & $103$ & $\phantom{0}91$  & $11$ & $11$ \\
DenseNet1D-121   & $251$  & $321$ & $227$ & $40$ & $35$ \\
\bottomrule
\end{tabular}
\end{table}
Aggregated over the full sweep, CE consumed $68.54$ GPU hours. 
A further $29.97$ GPU hours were spent on exploratory and repeated runs during development, an amount that does not grow with the seed count.

\paragraph{Synthetic metrics.}
The representation-correctness (RC) and importance-correctness (IC) metrics are computed jointly in a single pass, so the table reports one runtime per (architecture, method) covering both metrics together.
\begin{table}[h]
\centering
\caption{Mean synthetic-metric runtime (RC and IC computed jointly) per (architecture, method), in seconds, over $3$ seeds on a representative \textsc{syntheticLocal} variant. Fixed-$K$ methods sum over $K\in\{2,4,6,8,10\}$; AC variants run once.}
\label{tab:app_synthmetric_time}
\small
\begin{tabular}{@{}lrrrrr@{}}
\toprule
Architecture & ECLAD-ts & MultiVISION & CENDRe$_{k\text{Means}}$ & CENDRe$_{\text{silhouette}}$ & CENDRe$_{\text{HDBSCAN}}$ \\
\midrule
InceptionTime10  & $108$ & $74$ & $333$ & $42$ & $29$ \\
ResNet1D-18      & $107$ & $55$ & $308$ & $23$ & $23$ \\
DenseNet1D-121   & $111$ & $72$ & $464$ & $77$ & $50$ \\
\bottomrule
\end{tabular}
\end{table}
Aggregated over the full sweep, synthetic-metric computation consumed $95.08$ GPU hours. 
As for CE, an additional $39.58$ GPU hours were spent during development, independent of the seed count.

In total, across model training, CE and synthetic-metric experiments, the computational resources used to produce this paper amount to approximately $274$ GPU hours.

%% file: literature.bib
@inproceedings{holzapfel2025concept,
  title={Concept Extraction for Time Series with ECLAD-ts},
  author={Holzapfel, Antonia and Posada Moreno, Andres Felipe and Trimpe, Sebastian},
  booktitle={World Conference on Explainable Artificial Intelligence},
  pages={90--112},
  year={2025},
  publisher={Springer}
}

@article{posada-moreno2023eclad,
  title = {{{ECLAD}}: {{Extracting Concepts}} with {{Local Aggregated Descriptors}}},
  author = {Posada-Moreno, Andrés Felipe and Surya, Nikita and Trimpe, Sebastian},
  year = 2023,
  journal = {Pattern Recognition},
  volume = {147},
  pages = {110146},
}

@article{posada-moreno2023scalepreserving,
  title = {Scale-Preserving Automatic Concept Extraction ({{SPACE}})},
  author = {Posada-Moreno, Andrés Felipe and Kreisköther, Lukas and Glander, Tassilo and Trimpe, Sebastian},
  year = 2023,
  journal = {Machine Learning},
  volume = {112},
  number = {11},
  pages = {4495--4525},
  doi = {10.1007/s10994-023-06373-2},
}

@inproceedings{ghorbani2019automatic,
  title = {Towards {{Automatic Concept-based Explanations}}},
  booktitle = {NeurIPS},
  author = {Ghorbani, Amirata and Wexler, James and Zou, James Y and Kim, Been},
  year = 2019,
  volume = {32},
  publisher = {Curran Associates, Inc.},
}

@inproceedings{kamakshi2021pace,
  title = {{{PACE}}: {{Posthoc Architecture-Agnostic Concept Extractor}} for {{Explaining CNNs}}},
  booktitle = {IJCNN},
  author = {Kamakshi, Vidhya and Gupta, Uday and Krishnan, Narayanan C},
  year = 2021,
  pages = {1--8},
  doi = {10.1109/IJCNN52387.2021.9534369},
}

@misc{UCRArchive2018,
        title = {The UCR Time Series Classification Archive},
        author = {Dau, Hoang Anh and Keogh, Eamonn and Kamgar, Kaveh and Yeh, Chin-Chia Michael and Zhu, Yan 
                  and Gharghabi, Shaghayegh and Ratanamahatana, Chotirat Ann and Yanping and Hu, Bing 
                  and Begum, Nurjahan and Bagnall, Anthony and Mueen, Abdullah and Batista, Gustavo, Hexagon-ML},
        year = {2018},
        month = {October}
}

@inproceedings{decker2023explaining,
  title = {Explaining {{Deep Neural Networks}} for {{Bearing Fault Detection}} with {{Vibration Concepts}}},
  booktitle = {INDIN},
  author = {Decker, Thomas and Lebacher, Michael and Tresp, Volker},
  year = 2023,
  pages = {1--6},
  doi = {10.1109/INDIN51400.2023.10218170},
}

@inproceedings{madsen2023conceptbased,
  title={Concept-Based Explainability for an EEG Transformer Model},
  author={Madsen, Anders Gj{\o}lbye and Lehn-Schi{\o}ler, William Theodor and J{\'o}nsd{\'o}ttir, {\'A}shildur and Arnard{\'o}ttir, Bergd{\'\i}s and Hansen, Lars Kai},
  booktitle={MLSP},
  pages={1--6},
  year={2023},
  publisher={IEEE}
}

@inproceedings{kim2018interpretability,
  title = {Interpretability {{Beyond Feature Attribution}}: {{Quantitative Testing}} with {{Concept Activation Vectors}} ({{TCAV}})},
  booktitle = {ICML},
  author = {Kim, Been and Wattenberg, Martin and Gilmer, Justin and Cai, Carrie and Wexler, James and Viegas, Fernanda and Sayres, Rory},
  year = 2018,
  pages = {2668--2677},
  publisher = {PMLR},
}

@inproceedings{yeh2020completenessaware,
  title = {On Completeness-Aware Concept-Based Explanations in Deep Neural Networks},
  booktitle = {NeurIPS},
  author = {Yeh, Chih-Kuan and Kim, Been and Arik, Sercan and Li, Chun-Liang and Pfister, Tomas and Ravikumar, Pradeep},
  year = 2020,
  volume = {33},
  pages = {20554--20565},
}

@article{hills2014classification,
  title={Classification of time series by shapelet transformation},
  author={Hills, Jon and Lines, Jason and Baranauskas, Edgaras and Mapp, James and Bagnall, Anthony},
  journal={Data mining and knowledge discovery},
  year={2014},
  publisher={Springer}
}

@article{ismail2020inceptiontime,
  title={Inceptiontime: Finding alexnet for time series classification},
  author={Ismail Fawaz, Hassan and Lucas, Benjamin and Forestier, Germain and Pelletier, Charlotte and Schmidt, Daniel F and Weber, Jonathan and Webb, Geoffrey I and Idoumghar, Lhassane and Muller, Pierre-Alain and Petitjean, Fran{\c{c}}ois},
  journal={Data Mining and Knowledge Discovery},
  volume={34},
  number={6},
  pages={1936--1962},
  year={2020},
  publisher={Springer}
}

@inproceedings{pytorch,
author = {Ansel, Jason and Yang, Edward and He, Horace and Gimelshein, Natalia and Jain, Animesh and Voznesensky, Michael and Bao, Bin and Bell, Peter and Berard, David and Burovski, Evgeni and Chauhan, Geeta and Chourdia, Anjali and Constable, Will and Desmaison, Alban and DeVito, Zachary and Ellison, Elias and Feng, Will and Gong, Jiong and Gschwind, Michael and Hirsh, Brian and Huang, Sherlock and Kalambarkar, Kshiteej and Kirsch, Laurent and Lazos, Michael and Lezcano, Mario and Liang, Yanbo and Liang, Jason and Lu, Yinghai and Luk, C. K. and Maher, Bert and Pan, Yunjie and Puhrsch, Christian and Reso, Matthias and Saroufim, Mark and Siraichi, Marcos Yukio and Suk, Helen and Zhang, Shunting and Suo, Michael and Tillet, Phil and Zhao, Xu and Wang, Eikan and Zhou, Keren and Zou, Richard and Wang, Xiaodong and Mathews, Ajit and Wen, William and Chanan, Gregory and Wu, Peng and Chintala, Soumith},
title = {PyTorch 2: Faster Machine Learning Through Dynamic Python Bytecode Transformation and Graph Compilation},
year = {2024},
publisher = {Association for Computing Machinery},
doi = {10.1145/3620665.3640366},
booktitle = {ACM International Conference on Architectural Support for Programming Languages and Operating Systems},
pages = {929–947},
series = {ASPLOS '24}
}

@inproceedings{he2016deep,
  title={Deep residual learning for image recognition},
  author={He, Kaiming and Zhang, Xiangyu and Ren, Shaoqing and Sun, Jian},
  booktitle={CVPR},
  pages={770--778},
  year={2016}
}

@inproceedings{younis2022multivariate,
  title={Multivariate time series analysis: An interpretable cnn-based model},
  author={Younis, Raneen and Zerr, Sergej and Ahmadi, Zahra},
  booktitle={International Conference on Data Science and Advanced Analytics (DSAA)},
  pages={1--10},
  year={2022},
  publisher={IEEE}
}

@inproceedings{chung2024time,
  title={Time is Not Enough: Time-Frequency based Explanation for Time-Series Black-Box Models},
  author={Chung, Hyunseung and Jo, Sumin and Kwon, Yeonsu and Choi, Edward},
  booktitle={Proceedings of the 33rd ACM International Conference on Information and Knowledge Management},
  pages={394--403},
  year={2024},
doi = {10.1145/3627673.3679844}
}

@article{brusch2024explaining,
  title={Explaining time series models using frequency masking},
  author={Br{\"u}sch, Thea and Wickstr{\o}m, Kristoffer K and Schmidt, Mikkel N and Alstr{\o}m, Tommy S and Jenssen, Robert},
  journal={arXiv preprint arXiv:2406.13584},
  year={2024},
doi = {10.48550/arXiv.2406.13584}
}

@article{smilkov2017smoothgrad,
  title={Smoothgrad: removing noise by adding noise},
  author={Smilkov, Daniel and Thorat, Nikhil and Kim, Been and Vi{\'e}gas, Fernanda and Wattenberg, Martin},
  journal={arXiv preprint arXiv:1706.03825},
  year={2017}
}

@article{vielhaben2024explainable,
  title={Explainable AI for time series via virtual inspection layers},
  author={Vielhaben, Johanna and Lapuschkin, Sebastian and Montavon, Gr{\'e}goire and Samek, Wojciech},
  journal={Pattern Recognition},
  volume={150},
  pages={110309},
  year={2024},
  publisher={Elsevier}
}

@inproceedings{huang2017densely,
  title={Densely connected convolutional networks},
  author={Huang, Gao and Liu, Zhuang and Van Der Maaten, Laurens and Weinberger, Kilian Q},
  booktitle={CVPR},
  pages={4700--4708},
  year={2017}
}

@article{smith2015rolling,
  title={Rolling element bearing diagnostics using the {Case Western Reserve University} data: A benchmark study},
  author={Smith, Wade A. and Randall, Robert B.},
  journal={Mechanical Systems and Signal Processing},
  volume={64--65},
  pages={100--131},
  year={2015},
  publisher={Elsevier}
}

@article{randall2011rolling,
  title={Rolling element bearing diagnostics—A tutorial},
  author={Randall, Robert B and Antoni, Jerome},
  journal={Mechanical systems and signal processing},
  volume={25},
  number={2},
  pages={485--520},
  year={2011},
  publisher={Elsevier}
}

@inproceedings{lessmeier2016condition,
  title={Condition monitoring of bearing damage in electromechanical drive systems by using motor current signals of electric motors: A benchmark data set for data-driven classification},
  author={Lessmeier, Christian and Kimotho, James Kuria and Zimmer, Detmar and Sextro, Walter},
  booktitle={PHM society European conference},
  volume={3},
  year={2016}
}

@article{loshchilov2017decoupled,
  title={Decoupled weight decay regularization},
  author={Loshchilov, Ilya and Hutter, Frank},
  journal={arXiv preprint arXiv:1711.05101},
  year={2017}
}

@article{arras2022clevr,
  title={CLEVR-XAI: A benchmark dataset for the ground truth evaluation of neural network explanations},
  author={Arras, Leila and Osman, Ahmed and Samek, Wojciech},
  journal={Information Fusion},
  volume={81},
  pages={14--40},
  year={2022},
  publisher={Elsevier}
}

@inproceedings{campello2013density,
  title={Density-based clustering based on hierarchical density estimates},
  author={Campello, Ricardo JGB and Moulavi, Davoud and Sander, J{\"o}rg},
  booktitle={Pacific-Asia conference on knowledge discovery and data mining},
  pages={160--172},
  year={2013},
  publisher={Springer}
}

@article{rousseeuw1987silhouettes,
  title={Silhouettes: a graphical aid to the interpretation and validation of cluster analysis},
  author={Rousseeuw, Peter J},
  journal={Journal of computational and applied mathematics},
  volume={20},
  pages={53--65},
  year={1987},
  publisher={Elsevier}
}

@article{poeta2023concept,
  title={Concept-based explainable artificial intelligence: A survey},
  author={Poeta, Eleonora and Ciravegna, Gabriele and Pastor, Eliana and Cerquitelli, Tania and Baralis, Elena},
  journal={ACM Computing Surveys},
  year={2023},
  publisher={ACM New York, NY}
}

@article{kraft2025atrial,
  title={Atrial fibrillation and atrial flutter detection using deep learning},
  author={Kraft, Dimitri and Rumm, Peter},
  journal={Sensors},
  volume={25},
  number={13},
  pages={4109},
  year={2025},
  publisher={MDPI}
}

@article{ravi2024hybrid,
  title={A hybrid 1D CNN-BiLSTM model for epileptic seizure detection using multichannel EEG feature fusion},
  author={Ravi, Swathy and Radhakrishnan, Ashalatha},
  journal={Biomedical physics \& engineering express},
  volume={10},
  number={3},
  pages={035040},
  year={2024},
  publisher={IOP Publishing}
}

@article{fratti2024multi,
  title={A multi-scale CNN for transfer learning in sEMG-based hand gesture recognition for prosthetic devices},
  author={Fratti, Riccardo and Marini, Niccol{\`o} and Atzori, Manfredo and M{\"u}ller, Henning and Tiengo, Cesare and Bassetto, Franco},
  journal={Sensors},
  volume={24},
  number={22},
  pages={7147},
  year={2024},
  publisher={MDPI}
}

@article{kim2023time,
  title={Time-frequency multi-domain 1D convolutional neural network with channel-spatial attention for noise-robust bearing fault diagnosis},
  author={Kim, Yejin and Kim, Young-Keun},
  journal={Sensors},
  volume={23},
  number={23},
  pages={9311},
  year={2023},
  publisher={MDPI}
}

@article{stathatos2024convolutional,
  title={Convolutional neural networks for raw signal classification in CNC turning process monitoring},
  author={Stathatos, Emmanuel and Tzimas, Evangelos and Benardos, Panorios and Vosniakos, George-Christopher},
  journal={Sensors},
  volume={24},
  number={5},
  pages={1390},
  year={2024},
  publisher={MDPI}
}

@book{oppenheim1999discrete,
  title={Discrete-time signal processing},
  author={Oppenheim, Alan V},
  year={1999},
  publisher={Pearson Education India}
}

@book{aastrom2021feedback,
  title={Feedback systems: an introduction for scientists and engineers},
  author={{\AA}str{\"o}m, Karl Johan and Murray, Richard},
  year={2021},
  publisher={Princeton university press}
}

@article{bergstra2012random,
  title={Random search for hyper-parameter optimization.},
  author={Bergstra, James and Bengio, Yoshua},
  journal={Journal of machine learning research},
  volume={13},
  number={2},
  year={2012}
}

@article{morris1992factorial,
  title={Factorial sampling plans for preliminary computational experiments},
  author={Morris, Max D},
  journal={Quality control and applied statistics},
  volume={37},
  number={6},
  pages={307--310},
  year={1992},
  publisher={Executive Sciences Institute}
}

@article{campolongo2007effective,
  title={An effective screening design for sensitivity analysis of large models},
  author={Campolongo, Francesca and Cariboni, Jessica and Saltelli, Andrea},
  journal={Environmental modelling \& software},
  volume={22},
  number={10},
  pages={1509--1518},
  year={2007},
  publisher={Elsevier}
}

@article{sobol2001global,
  title={Global sensitivity indices for nonlinear mathematical models and their Monte Carlo estimates},
  author={Sobol, Ilya M},
  journal={Mathematics and computers in simulation},
  volume={55},
  number={1-3},
  pages={271--280},
  year={2001},
  publisher={Elsevier}
}

@book{saltelli2008global,
  title={Global sensitivity analysis: the primer},
  author={Saltelli, Andrea and Ratto, Marco and Andres, Terry and Campolongo, Francesca and Cariboni, Jessica and Gatelli, Debora and Saisana, Michaela and Tarantola, Stefano},
  year={2008},
  publisher={John Wiley \& Sons}
}

@article{olejnik2003generalized,
  title={Generalized eta and omega squared statistics: measures of effect size for some common research designs.},
  author={Olejnik, Stephen and Algina, James},
  journal={Psychological methods},
  volume={8},
  number={4},
  pages={434},
  year={2003},
  publisher={American Psychological Association}
}

@article{holm1979simple,
  title={A simple sequentially rejective multiple test procedure},
  author={Holm, Sture},
  journal={Scandinavian journal of statistics},
  pages={65--70},
  year={1979},
  publisher={JSTOR}
}

@article{wolter2024ptwt,
  title={Ptwt-the pytorch wavelet toolbox},
  author={Wolter, Moritz and Blanke, Felix and Garcke, Jochen and Hoyt, Charles Tapley},
  journal={Journal of Machine Learning Research},
  volume={25},
  number={80},
  pages={1--7},
  year={2024}
}

@article{lee2019pywavelets,
  title={PyWavelets: A Python package for wavelet analysis},
  author={Lee, Gregory and Gommers, Ralf and Waselewski, Filip and Wohlfahrt, Kai and O'Leary, Aaron},
  journal={Journal of Open Source Software},
  volume={4},
  number={36},
  pages={1237},
  year={2019},
  publisher={The Open Journal}
}

@incollection{wilcoxon1992individual,
  title={Individual comparisons by ranking methods},
  author={Wilcoxon, Frank},
  booktitle={Breakthroughs in statistics: Methodology and distribution},
  pages={196--202},
  year={1992},
  publisher={Springer}
}

@article{kerby2014simple,
  title={The simple difference formula: An approach to teaching nonparametric correlation},
  author={Kerby, Dave S},
  journal={Comprehensive Psychology},
  volume={3},
  pages={11--IT},
  year={2014},
  publisher={SAGE Publications Sage CA: Los Angeles, CA}
}

@inproceedings{sculley2010web,
  title={Web-scale k-means clustering},
  author={Sculley, David},
  booktitle={Proceedings of the 19th international conference on World wide web},
  pages={1177--1178},
  year={2010}
}

@article{mullner2011modern,
  title={Modern hierarchical, agglomerative clustering algorithms},
  author={M{\"u}llner, Daniel},
  journal={arXiv preprint arXiv:1109.2378},
  year={2011}
}
